\documentclass{clv3}

\usepackage{hyperref}
\usepackage{xcolor}
\definecolor{darkblue}{rgb}{0, 0, 0.5}
\hypersetup{colorlinks=true,citecolor=darkblue, linkcolor=darkblue, urlcolor=darkblue}

\usepackage{todonotes}
\usepackage{times}
\usepackage{latexsym}
\usepackage{graphicx}
\usepackage{amsmath,amssymb}
\usepackage{rotating}
\usepackage{subcaption}
\usepackage{enumitem}
\usepackage{multirow}
\newcolumntype{M}[1]{>{\centering\arraybackslash}m{#1}}

\usepackage{microtype}
\usepackage{lscape}
\usepackage{array}
\usepackage{booktabs}
\usepackage{longtable}
\usepackage{amsmath}

\usepackage[normalem]{ulem}

\definecolor{bleudefrance}{rgb}{0.19, 0.55, 0.91}
\newcommand{\marco}[1]{\textcolor{black}{#1}}

\begin{document}
\issue{1}{1}{2024}


\runningtitle{Investigating Idiomaticity in Word Representations}

\runningauthor{Wei He}

\pageonefooter{Action editor: Tal Linzen. Submission received: 22 April 2024; revised version received: 29 August 2024; accepted for publication: 22 October 2024.}

\title{Investigating Idiomaticity in Word Representations}







\author{Wei He\thanks{211 Portobello, Sheffield, S1 4DP, UK.}
}
\affil{University of Sheffield\\
School of Computer Science
}

\author{Tiago Kramer Vieira\thanks{Av. Bento Gon\c{c}alves, 9500, Porto Alegre, 90000 Brazil.}
}
\affil{Federal University of Rio Grande do Sul\\Institute of Informatics}

\author{Marcos Garcia\thanks{Rua de Jenaro de la Fuente Dominguez s/n, 15782 Santiago de Compostela, Galiza, Spain.}
}
\affil{University of Santiago de Compostela\\CiTIUS Research Center}

\author{Carolina Scarton$^{*}$
}
\affil{University of Sheffield\\
School of Computer Science
}

\author{Marco Idiart$^{**}$\thanks{Av. Bento Gon\c{c}alves, 9500, Porto Alegre, 90000, Brazil.}
}
\affil{Federal University of Rio Grande do Sul\\ Institute of Physics}

\author{Aline Villavicencio$^{*}$\thanks{Innovation Centre, Rennes Drive, Exeter, EX4 4RN, UK. E-mail: a.villavicencio@exeter.ac.uk.}}
\affil{University of Exeter\\Institute for Data Science and Artificial Intelligence\\
The Alan Turing Institute\\University of Sheffield\\
School of Computer Science}

\maketitle

\begin{abstract}
Idiomatic expressions are an integral part of human languages, often used to express complex ideas in compressed or conventional ways (e.g. \textbf{eager beaver} as a keen and enthusiastic person). However, their interpretations may not be straightforwardly linked to the meanings of their individual components in isolation and this may have an impact for compositional approaches. 
In this paper, we investigate to what extent word representation models are able to go beyond compositional word combinations and capture multiword expression idiomaticity and some of the expected properties related to idiomatic meanings. 
We focus on noun compounds of varying levels of idiomaticity in two languages (English and Portuguese), 
presenting a dataset of minimal pairs containing human idiomaticity judgments for each noun compound at both type and token levels, their paraphrases and their occurrences in naturalistic and sense-neutral contexts, totalling 32,200 sentences.
We propose this set of minimal pairs 
for evaluating how well a model captures idiomatic meanings,
and define a set of fine-grained metrics of Affinity and Scaled Similarity, to determine how sensitive the models are to perturbations that may lead to changes in idiomaticity. Affinity is a comparative measure of the similarity between an experimental item, 
a target and a potential distractor, and Scaled Similarity incorporates a rescaling factor to magnify the meaningful similarities within the spaces defined by each specific model. The results obtained with a variety of representative and widely used models indicate that, despite superficial indications to the contrary in the form of high similarities, idiomaticity is not yet accurately represented in current models. Moreover, the performance of models with different levels of contextualisation suggests that their ability to capture context  is not yet able to go beyond more superficial lexical clues provided by the words and to actually incorporate the relevant semantic clues needed for idiomaticity.      
By proposing model-agnostic measures for assessing the ability of models to capture idiomaticity, this paper contributes to determining  limitations in the handling of non-compositional structures, which is one of the directions that needs to be considered for more natural, accurate and robust language understanding. The source code and additional materials related to this paper are available at our GitHub repository\footnote{\url{https://github.com/risehnhew/Finding-Idiomaticity-in-Word-Representations}}.

\end{abstract}

\section{Introduction}

\label{sec:Intro}

The evolution of word representation models has resulted in models with seemingly remarkable language abilities.  
Not surprisingly these models have been found to store  a wealth of linguistic information \cite{henderson-2020-unstoppable,Manning:2020,vulic-etal-2020-probing,Lenci:2022}, displaying high levels of performance on various tasks ranging from the abilities of even the 
static models of detecting semantic similarities between different words \cite{lin-1999-automatic,mikolov-word2vec,baroni-etal-2014-dont} to those of contextualised models of grouping representations in clusters which seem to be related to the various senses of the word \cite{schuster-etal-2019-cross} and can be matched to specific sense definitions \cite{chang-chen-2019-word}. While substantial evaluation efforts  
have concentrated on word and subword units and on larger compositional combinations derived from them, there is less understanding about their ability for handling less compositional structures, such as those found on multiword expressions (MWEs), like noun compounds (NCs) \cite{garcia-etal-2021-assessing}, verb-noun combinations \cite{King:2018,hashempour:2020} and idioms \cite{yu-ettinger-2020-assessing,dankers-etal-2022-transformer}. 
Indeed, MWEs include a variety of distinct phenomena and have been described as interpretations that cross word boundaries \cite{sag2002multiword}, whose meanings are not always straightforwardly derivable from the meanings of their individual components. Moreover, although they include, on the one hand, more transparent and compositional expressions (like \emph{salt and pepper}) or expressions with implicit relations (like \emph{olive oil}  as oil \emph{made from} olives), on the other hand they also include more idiomatic expressions (like \emph{eager beaver} as a person who is willing to work very hard\footnote{Definition from the Cambridge dictionary.}), falling into a continuum of idiomaticity\footnote{We understand idiomaticity as \emph{semantic opacity}, and its continuum as different \emph{degrees of opacity}. } \cite{sag2002multiword,fazly-etal-2009-unsupervised}. This leads to potential problems for models if they follow the Principle of Compositionality \cite{frege1956thought, montague1973proper}, building the meaning of a larger unit (like a sentence or an expression) from a combination of the individual meanings of the words that are contained in it, as this would result in potentially incomplete or incorrect interpretation for more idiomatic cases (e.g. the idiomatic \emph{eager beaver} interpreted literally as \emph{impatient rodent}). Although understanding the meaning of an MWE may require knowledge that goes beyond that of the meanings of these individual words in isolation \cite{Nunberg:1994}, failure to take idiomaticity into account can affect the quality of downstream tasks \cite{,sag2002multiword,constant-etal-2017-survey,cordeiro-etal-2019-unsupervised} such as reasoning and inference \cite{chakrabarty-etal-2022-rocket,chakrabarty-etal-2022-flute,saakyan-etal-2022-report}, information retrieval \cite{acosta:2011}  and machine translation \cite{dankers-etal-2022-transformer}. For machine translation, for example, the degree of idiomaticity and ambiguity of MWEs (literal vs. idiomatic usages) were found to have an impact on the quality of the results obtained \cite{dankers-etal-2022-transformer}. Due to their non-compositional nature, idiomatic expressions result in lower quality translations than literal expressions, as evidenced by lower BLEU scores for translations that are paraphrased rather than translated word-for-word. In this paper, we investigate to what extent widely used word representation models are able to capture idiomaticity in MWEs. We focus, in particular, on their initial abilities for representing idiomaticity, looking at noun compounds of varying degrees of idiomaticity.\footnote{
We use the off-the-shelf publicly available pre-trained versions of widely adopted word representation models, standard operations and common similarity measures. Even in scenarios in which adopting 
additional optimisations, more complex operations or fine-tuning could lead to improvements in performance, this may depend on 
 the availability of comprehensive training data for the target model, domain and language. Measuring the initial idiomaptic abilities of models can help understand the potential loss of idiomatic meaning that could be propagated to the downstream tasks that use them off-the-shelf. 
}  
In addition to the complex interactions between MWEs, their component words and their contexts \cite{sag2002multiword},  characteristics of languages and of word representation models may affect how accurately MWEs can be represented and processed, and we investigate the impact of some of these factors for compounds  
in two different languages (English and Portuguese).

One of the challenges is that 
uncovering how word representation models  capture a specific type of knowledge is a non-trivial problem \cite{vulic-etal-2020-probing},  and  may depend on factors like the particular model and the way it encodes different types of linguistic information \cite{yu-ettinger-2020-assessing}. For instance, while in Transformer-based models, the initial layers seem to represent more lexical level knowledge and the final layers seem to capture more semantic and pragmatic information \cite{rogers-etal-2020-primer}, determining where phenomena which sit at the interface of various levels are encoded, like multiword expressions   \cite{sag2002multiword}, is challenging since they could potentially involve information distributed across different layers. Moreover, the possible findings from an investigation about where in the architecture of a given model   idiomaticity is encoded, or about the role of particular components in representing it may not generalise to other models and architectures. 
In this paper we propose instead a set of model-agnostic idiomatic probes for assessing the representation of  idiomaticity. These probes contain NCs of different levels of idiomaticity, ranging from idiomatic to compositional cases, which form the basis for minimal pairs. In these pairs one of them contains an NC and the other contains a semantically related item (such as a synonym   ) or a distractor. 
The hypothesis is that if a model is able to accurately represent an NC, higher similarities will be observed for minimal pairs involving NCs and their synonyms (e.g. for the idiomatic \emph{eager beaver} and  \emph{hardworking person}). Conversely, for minimal pairs with variants that may incorporate changes in meaning, such as those containing NCs and synonyms of their individual component words (e.g. the idiomatic \emph{eager beaver} and \emph{impatient rodent}) or other distractors, lower similarities should be observed.

As word representation models may form spaces that are anisotropic \cite{ethayarajh-jurafsky-2021-attention} with representations concentrating on parts of the space, or may have rogue dimensions that dominate similarity measures  \cite{timkey2021all}, these could lead to high similarities overall \cite{liu-etal-2020-multilingual}, affecting the ability to distinguishing  meaningful similarities from spurious ones arising from specific characteristics of a given space. 
In this paper, we propose two new measures to assess idiomaticity within a model while taking into account its potential for high similarities. The first, \textit{Assessment of Feature Familiarity and Idiomatic Nuance by Interpreting Target Yielding} (Affinity), takes two representations of different levels of relatedness to a given target, and can be used to determine if a model accurately reflects their degree of similarity 
to the target. 
Focusing on idiomaticity, we use Affinity to assess if greater similarities are observed for NCs and related words (in this case their synonyms), than for NCs and other potentially less related alternatives including  distractors. The second measure, Scaled Similarity, determines a new lowerbound for a given space in terms of similarities for unrelated representations, rescaling the space  to help distinguish them from the  meaningful similarities for related representations. For idiomaticity, we analyse the similarities between the NCs and their synonyms adopting the similarities between the NCs and 
random items as a new lowerbound. These measures of Affinity and Scaled Similarity do not directly address the problem of rogue dimensions, and we discuss this further in the Conclusions section.

Using these metrics and minimal pairs for evaluation, this paper presents a fine-grained analysis of the ability of a model to capture idiomaticity, looking at the following questions:
\begin{itemize}
    \item[Q1] \textbf{To what extent is idiomaticity captured by word representation models?} We assess this by comparing the predictions of models for NCs and their synonyms against human judgements about idiomaticity in the same sentences, analysing how sensitive these models are to potential changes in meaning resulting from the lexical variations in the minimal pairs.   

    \item[Q2] \textbf{Is this ability affected by the degree of idiomaticity of the NCs, the informativeness of the contexts, or the languages involved?} To determine if more idiomatic expressions are more challenging for models, we present an analysis of the impact of the level of idiomaticity of the NCs. We also analyse more informative contexts provided by naturalistic sentences against uninformative neutral contexts 
    to determine their impact on idiomaticity representation.   
    These evaluations include two languages, to measure the potential language dependence of these results.  
    \item[Q3] \textbf{Do contextualised models (from transformer-based models) perform better compared to static models in idiomaticity representation?}
     In addressing this question, we conduct a comparative analysis across different static and contextualised  models, focusing on their ability to capture idiomatic expressions. This involves examining how each model represents idiomatic NCs of varying levels of idiomaticity, in sentences that contain more (or less) informative contexts, and the accuracy with which they reflect the nuanced meanings that idiomaticity often entails. The analyses consider various linguistic scenarios that can change the idiomatic meaning to comprehensively assess the accuracy of contextualised models over their static counterparts.

\end{itemize}

The main contributions of this work include:
\begin{itemize}

\item The Noun Compound Idiomaticity Minimal Pairs (NCIMP) Dataset, a dataset of minimal pair sentences containing NCs of varying levels of idiomaticity, along with human judgments about the degree of NC idiomaticity and gold standard paraphrases, at both type and token level. In total, the dataset contains 32,200 sentences for two languages 
(19,600 in English and 12,600 in Portuguese).\footnote{This work extends the idiomatic probes proposed by \citet{garcia-etal-2021-probing} and the type and token annotations by \citet{garcia-etal-2021-assessing}, also introducing new measures, additional tests and substantially expanding the analyses with new baselines and results from a larger set of models.}

\item A comparative measure of Affinity to help determine how accurately idiomaticity is incorporated in these representations contrasting similarities for semantically related and unrelated representations. 

\item A novel model-agnostic measure of Scaled Similarity, which rescales a space in relation to a new lowerbound taking into account expected similarities among random items to magnify meaningful similarities among semantically related representations. 

\item In-depth analyses of the representation of idiomaticity in widely used word representation models, examining their ability to display sensitivity to changes in idiomaticity.

\end{itemize}

The remainder of this paper is organised as follows: Section {2} presents related work, and Section 3 the NCIMP dataset (Section {3.1}), the models (Section {3.2}) and the proposed idiomatic probes and measures (Section {3.3}). 
Finally, in Section {4} we discuss the results of our experiments and draw conclusions in Section {5}.

\section{Representing Multiword Expressions and Idiomaticity}
\label{related1}

\subsection{Static and contextualised models for representing MWEs}
A variety of vector models have been used to investigate the representation of multiword expressions (MWEs), ranging from static to contextualised representations, each with its own set of challenges \cite{contreras2022models,garcia-etal-2021-assessing,liu-neubig-2022-representations}. The former include models like Word2Vec \cite{mikolov-word2vec}, GloVe \cite{pennington-etal-2014-glove} and fastText \cite{bojanowski:2017}, which represent words at type-level, producing a single vector for each word that conflates all its senses. At this level, MWEs are often represented based on their overall syntactic and semantic properties as they are generally understood, without taking into account the variability of contexts. For example, both the literal and the idiomatic meaning of \emph{gold mine}\footnote{``Opportunity for making a lot of money'' (definition from the Cambridge dictionary).} would be represented jointly in a single vector regardless of its use in any specific sentence.
At the other end of the scale are the contextualised models, from ELMo \cite{peters-etal-2018-deep}, BERT \cite{devlin-etal-2019-bert} and GPT-3 \cite{brown:2020} to LLaMA \cite{llama} and other large language models, which produce token-level dynamic representations dedicated to capturing specific usages of a word in a particular context, resulting in several vectors for each word \cite{Lenci:2022,Apidianaki:2023}.
Token-level representations focus on the specific occurrences of words or subwords within contexts, and how their meaning or function may vary or be influenced by the surrounding text. Therefore, they have the potential for accurately representing MWEs, capturing the interdependence of the idiomatic meaning on a particular configuration of words, while also anchoring the MWEs in relation to their immediate linguistic environment. The primary challenge at token-level is accurately determining the presence, meaning and role of MWEs in specific contexts, especially when they have possibly multiple literal and idiomatic readings or when they are part of complex syntactic structures \cite{zeng-bhat-2021-idiomatic}.

Evaluation of successive generations of word representation models, ranging from static 
\cite{landauer1997solution,lin-1999-automatic,baroni-lenci-2010-distributional,mikolov-word2vec,bojanowski:2017} to contextualised models \cite{peters-etal-2018-deep,devlin-etal-2019-bert,brown:2020,llama}, has devoted considerable attention to their linguistic abilities \cite{Mandera:2017,wang-etal-2018-glue,henderson-2020-unstoppable,rogers-etal-2020-primer,Lenci:2022}. On lexical semantics, the representations extracted from contextualised models seem to be able to reflect word senses in clusters of vectors (e.g., \citet{wiedemann2019does} for BERT) including in cross-lingual alignments involving polysemous words (e.g., \citet{schuster-etal-2019-cross} for ELMo).
However, controlled uniform evaluations of different generations of word representation models settings have also reported strong performances from static models, which were able to outperform contextualised models in most tasks \cite{Lenci:2022}.

\subsection{Vector models evaluation on idiomaticity} 
Regarding idiomaticity, uniform assessment of the performance of different models on the processing of MWEs are particularly important, as independent evaluations have reported mixed results \cite{King:2018,nandakumar:2019,cordeiro-etal-2019-unsupervised,hashempour:2020,garcia-etal-2021-probing,klubivcka2023idioms}.
For instance, for the task of identifying the degree of idiomaticity of MWEs at type level (i.e. the potential of an MWE to be idiomatic in general), good performances have been obtained with static word embeddings \cite{MitchellLapata2010,reddy-etal-2011-empirical,cordeiro-etal-2019-unsupervised}, and they have even been reported as obtaining better performance than contextualised models for capturing idiomaticity in MWEs in some evaluations \cite{King:2018,nandakumar:2019}. Likewise, BERT-based models obtained similar results to those of static vector representations for predicting the degree of compositionality of a given NC \cite{miletic-schulte-im-walde-2023-systematic}.\footnote{See \citet{miletic-walde-2024-semantics} for a recent survey on the representation of MWEs in Transformer-based models.}

However, a potential limitation of static models is that in representing different word senses in the same vector, the literal usage of an expression may differ considerably from its idiomatic usage (e.g. a \emph{brass ring} as an idiomatic prize or as a literal ring made of brass), and complex operations may be required to deal with semantic phenomena like polysemy \cite{erk2012vector}. In this sense, contextualised models may provide the means for distinguishing literal from idiomatic usages, along with fine-grained sense distinctions. In this respect, \citet{garcia-etal-2021-assessing} proposed probing metrics to investigate and understand the linguistic information encoded in the models' representations. Similarly, using a method of probing with noise and a repurposed idiomatic usage probing task revealed better performance by BERT in encoding idiomaticity compared to GloVe \cite{klubivcka2023idioms}.
These types of intrinsic evaluations have also been framed as shared tasks, like SemEval-2022 task 2B \cite{tayyar-madabushi-etal-2022-semeval} which proposed the assessment of idiomaticity representation in multilingual texts (English, Portuguese and Galician) while also requiring models to predict the semantic text similarity (STS) scores between sentence pairs, regardless of whether or not either sentence contains an idiomatic expression.

Extrinsic evaluations have measured how well the representation of idiomaticity in a model impacts downstream tasks, e.g., sentence generation \cite{zhou2021pie}, or conversational systems \cite{adewumi2022vector}. For instance, evaluations of different classifiers initialised with static and contextualised embeddings in five tasks related to lexical composition (including the literality of NCs) found that contextualised models led to better performance across all tasks \cite{shwartz-dagan-2019-still}, and supervised methods that used contextualised models also outperformed alternatives on the classification of potentially idiomatic expressions in both monolingual and cross-lingual (English and Russian) scenarios \cite{kurfali-ostling-2020-disambiguation,fakharian-cook-2021-contextualized}. Alternatively, both types of representations can be combined, as for example, in a supervised neural architecture to identify and classify potentially idiomatic expressions combining contextualised and static embeddings in an attention flow \cite{zeng-bhat-2021-idiomatic}. Regarding machine translation, a recent evaluation of compositional generalisation in transformer models found that they tend to perform too compositional translations even for idiomatic expressions \cite{dankers-etal-2022-transformer}. Furthermore, an analysis of GPT-3 \cite{brown:2020} reported $50.7\%$ accuracy in idiom comprehension \cite{zeng2022getting}, suggesting that the models' ability to deal with idiomaticity is not yet adequate.

\subsection{Vector operations and idiomatic knowledge induction}
In addition to the level of contextualisation, the performance of vector space models may also be affected by the way the target words of an expression are composed, with functions like sum, concatenation and multiplication used for combining the words of static models \cite{cordeiro-etal-2019-unsupervised,MitchellLapata2010,reddy-etal-2011-empirical} or the subwords of contextualised models \cite{garcia-etal-2021-probing}. For the embeddings extracted from language models, other potential sources of variation include which input is given to the model (e.g., one vs. several sentences including the target MWE in evaluations at the type level), or the number of layers that will be taken into account to obtain the vector representation \cite{miletic-walde-2024-semantics}. In this regard, the intermediate and last layers seem to encode more semantic information at the token level \cite{tenney-etal-2019-bert,garcia-2021-exploring}, while other evaluations at the type level found that the averaging the initial layers of the target expressions achieved the best results (e.g., \citet{miletic-schulte-im-walde-2023-systematic} for NCs and \citet{vulic-etal-2020-probing} for single word semantic tasks). With respect to semantic composition, \citet{yu-ettinger-2020-assessing} explored the type level representation of two word phrases (which in many cases correspond to NCs as the ones used in our study) in various contextualised models, showing that phrase representations miss compositionality effects as they heavily rely on word content. Similar conclusions, for neural machine translation, can be inferred from \citet{dankers-etal-2022-transformer}. While some of these evaluations rely on substitutivity and the changes to a larger phrase representation caused by substitutions to its constituents \cite{garcia-etal-2021-probing,yu-ettinger-2020-assessing}, alternatively, the notion of localism has also been analysed \cite{liu-neubig-2022-representations} focusing on whether the operations of a model are local \cite{Hupkes:2021}, that is, the extent to which the representation of a phrase is derivable from its local structure.

Crucially, a substantial amount of the discussed studies evaluate idiomaticity at the type-level, i.e., they obtain the embedding of a given MWE by averaging its representation in several sentences that have been previously extracted in an automatic way. A more detailed controlled comparison of type-level and token-level idiomaticity reported compatible results for both levels, with type-level being a close approximation for token-level \cite{garcia-etal-2021-assessing} in sentences where the NC occurs with the same sense. Further analysis of the occurrences of these NCs in fine-grained sense annotations of literal and idiomatic usages \cite{tayyar-madabushi-etal-2021-astitchinlanguagemodels-dataset} provided additional confirmation that the ability of contextualised models to capture idiomaticity during pre-training was limited, with approaches for building single token representations \cite{phelps-etal-2022-sample} and for fine-tuning leading to more accurate representations \cite{tayyar-madabushi-etal-2022-semeval}. Recent alternatives for representing idiomatic expressions also include adding a new adapter module which has been developed and trained to recognise idioms \cite{zeng2022getting}. This module functions as a language expert for idioms, augmenting the learning process of BART \cite{lewis2019bart} with additional information, and this approach effectively improves the representation of idiomatic expressions in off-the-shelf pre-trained language models, equipping them with greater ability to navigate the intricacies of natural language. \citet{zeng2023unified} also proposed PIER+, a language model improvement for handling both literal and figurative language. This is achieved by combining a base model with an additional curriculum learning framework that gradually introduces more complex potentially idiomatic expressions. Compared to other models, PIER+ demonstrates better performance at identifying, understanding, and maintaining proficiency in both types of expressions. Finally, \citet{zeng2023iekg} introduce a knowledge graph designed to enhance the understanding of idiomatic expressions, which integrates commonsense knowledge to aid in deciphering the non-literal meanings of idioms. This work demonstrates how to inject MWE-related knowledge into pre-trained language models effectively. However, it is still unclear to what extent the context and its representation in contextualised models are contributing to a more accurate representation of MWEs according to their idiomaticity level \cite{nedumpozhimana-kelleher-2021-finding,miletic-schulte-im-walde-2023-systematic}.

\subsection{Towards a more controlled assessment of idiomaticity in vector space models}
Shedding some light on these questions requires a more controlled evaluation setup and measures that can abstract away from the particularities of these word representation spaces. In this effort, we take inspiration in psycholinguistic methodologies, which have been traditionally used to examine how humans process language in controlled experimental setups, to allow the removal of obvious biases and potentially confounding factors from evaluations \cite{linzen-etal-2016-assessing,gulordava-etal-2018-colorless}. They also enable comparative analyses of performance in artificially constructed but controlled sentences and in naturally occurring sentences.

Setups like these have been used, for instance, to investigate how models represent syntax, if they understand negation \cite{van-schijndel-linzen-2018-neural,prasad-etal-2019-using,ettinger-2020-bert,kassner-schutze-2020-negated}, 
and if they are aware of which properties are relevant for which concepts \cite{misra:2023}. Adopting evaluation protocols that use minimal pair sentences (e.g.,\citet{warstadt-etal-2020-blimp-benchmark,misra:2023}) allows for a controlled comparison of the target item against carefully selected distractors that may share linguistic properties with them. For instance, a dataset of Conceptual Minimal Pair Sentences (COMPS) was used to compare the performance of 22 large language models including both masked language models (like BERT) and autoregressive language models (like GPT-2), where the models have to validate which of two concepts a given property belongs to (e.g. \emph{stripes} for \emph{zebras} vs. \emph{oaks}). Although the models seem to obtain relatively high accuracies for attributing properties to concepts, when semantically related concepts are involved or distractors are included, performances drop substantially, and go below chance even for models like GPT-3 \cite{misra:2023}. Similarly, in targeted syntactic evaluation \cite{marvin-linzen-2018-targeted}, models are assessed using minimal pairs datasets focused on specific syntactic phenomena, such as those included in the BLiMP dataset for English \citep{warstadt-etal-2020-blimp-benchmark}. Analyses like these highlight the importance of adding controls to the experimental setup to distinguish seemingly sophisticated behaviour with high performances that give the illusion of knowledge from robust understanding with access to meaning \cite{misra:2023,de-dios-flores-etal-2023-dependency}. With this is mind, we follow \citet{garcia-etal-2021-probing} and use minimal pairs to propose a set of intrinsic evaluations including probes and affinity measures aimed at gaining a better understanding of how vector space models represent MWEs with different degrees of semantic compositionality in context.

\subsection{Datasets for exploring idiomaticity in computational models}
Concerning experimental data, the first datasets to evaluate computational models were composed of different types of multiword expressions annotated at the type-level \cite{mccarthy-etal-2003-detecting,venkatapathy-joshi-2005-measuring}.
Further studies released annotations of MWEs in context, such as the VNC-tokens dataset \cite{cook2008vnc}, which includes 60 English verb-noun combinations occurring in almost 3,000 sentences annotated as idiomatic or literal, or the IDIX corpus \cite{sporleder-etal-2010-idioms}, with almost 6,000 labeled sentences of 78 expressions extracted from the BNC. Using a crowdsourcing platform, \citet{reddy-etal-2011-empirical} released a dataset with numerical ratings of the compositionality degree of 90 noun compounds in English, which also includes the contribution of each component to the meaning of the MWEs. Similar efforts were carried out for other languages, such as the GhoSt-NN dataset for German \cite{schulte-im-walde-etal-2016-ghost}, or the NC Compositionality (NCC) dataset \cite{cordeiro-etal-2019-unsupervised}, which expanded the resource provided by \citet{reddy-etal-2011-empirical} with additional NCs for English, and new data for Portuguese and French. Semi-automatic techniques combined with crowdsourced annotations were used to compile MAGPIE \cite{haagsma-etal-2020-magpie}, a large resource of more than 50,000 sentences with binary annotations at the token level of potentially idiomatic expressions. Similarly, the AStitchInLanguageModels dataset \cite{tayyar-madabushi-etal-2021-astitchinlanguagemodels-dataset}, used in the ``SemEval-2022 Task 2: Multilingual Idiomaticity Detection and Sentence Embedding'' \cite{tayyar-madabushi-etal-2022-semeval}, also contains potentially idiomatic expressions annotated in naturalistic sentences.

Recently, \citet{garcia-etal-2021-assessing} and \citet{garcia-etal-2021-probing} enriched the English and Portuguese data of the NCC dataset with crowdsourced annotations of the compositionality degree of noun compounds and their components at the token level, paraphrases of the NCs in context, and different types of controlled replacements. These variants compose a large set of minimal pairs which allow for the systematic exploration of the representation of idiomaticity in vector space models.\footnote{We refer to \citet{ramisch2023multiword} for a recent review on MWEs processing, including datasets, and to \citet{im2023collecting} for a comprehensive overview on compositionality ratings for MWEs.}

In this paper, we adopt the minimal pairs paradigm as one of the bases for the evaluation and present the Noun Compound Idiomaticity Minimal Pairs dataset, which contains a set of idiomatic probes to explore to what extent idiomaticity is captured in word representation models. To do so, we rely on the datasets for English and Portuguese by \citet{garcia-etal-2021-assessing} and \citet{garcia-etal-2021-probing} and extend them with new semantically related variants and distractors and sets of  minimal pairs as discussed in the next section to conduct in-depth intrinsic evaluations.


\section{Materials and Methods}\label{sec:materials}

\subsection{Noun Compound Idiomaticity Minimal Pairs Dataset}
\label{sec:NCI}

The Noun Compound Idiomaticity Minimal Pairs (NCIMP) dataset contains 32,200 sentences targeting two-word NCs in two languages, 280 in English (EN) and 180 in Portuguese (PT), with idiomatic (e.g. \textit{gravy train\footnote{Referring to an easy way of making money without doing much work (Cambridge Dictionary).}}), partly compositional (e.g., \textit{grandfather clock}\footnote{A type of tall free-standing clock.}), and compositional (e.g., \textit{research project}) NCs.\footnote{The NCIMP dataset is based on the Noun Compound Senses \cite{garcia-etal-2021-probing}, the Noun Compound Type and Token Idiomaticity \cite{garcia-etal-2021-assessing} and the NC Compositionality \cite{cordeiro-etal-2019-unsupervised} datasets, significantly extending them with new  data.} For each NC, the dataset contains minimal pairs formed by a first sentence with the target NC and a second sentence where the NC was replaced by an experimental item. These experimental items were selected on the basis of MWE properties, like more limited substitutability (or greater lexical fixedness), and can be used to determine if models are sensitive to perturbations to these properties, and if this is affected by how idiomatic the NCs are. For example, depending on the degree of lexical fixedness of an NC, the variants generated may not fully retain its original meaning (e.g. \textit{panda car}\footnote{Referring to a police car.} and ?\textit{bear automobile}). In particular, we analyse the following: 

\begin{itemize}
    \item $NC_{Syn}$: the minimal pairs are formed by the NC being replaced by one of the \textbf{gold standard synonyms} provided holistically for the NC by the annotators (e.g. \emph{brain} for \emph{grey matter}). In this case, we adopted the synonyms provided by the Noun Compound Senses (NCS) dataset \cite{garcia-etal-2021-probing}, which were selected on the basis of the most frequent paraphrases given by native speaker annotators. These pairs are used to assess if the models provide similar representations for NCs and their synonyms, even if they involve lexically diverse surface forms. 

    \item $NC_{WordsSyn}$: minimal pairs where each component word of the NC is replaced individually by a synonym generating new \textbf{two-word compositional replacements} (e.g. forming \emph{alligator sobs} for the NC \textit{crocodile tears} by replacing \textit{alligator} for \textit{crocodile} and \textit{sobs} for \textit{tears}). The synonyms were manually selected from WordNet \cite{miller1995wordnet} for English, and OpenWordNet \cite{rademaker-etal-2014-openwordnet} for Portuguese, and from online dictionaries of synonyms where additional coverage was required. In case of ambiguity (due to polysemy or homonymy), the most common meaning of each component was selected. For each NC, 5 compositional replacements were generated. These pairs are used to evaluate how sensitive a model is to the conventionality and lexical fixedness of these NCs, especially the more idiomatic ones, and if it can detect when the (idiomatic) meaning changes with the replacements.

    \item $NC_{Comp}$: the minimal pairs are formed by replacing the NC by only one of its \textbf{component words} i.e., replacing the NC by its head in one minimal pair, and by the modifier in the other pair (e.g. \textit{crocodile} for  \textit{crocodile tears} and \textit{tears} for \textit{crocodile tears}). These pairs are used to explore if the models can detect when the meaning of an NC is related to the meaning of a component (in more compositional cases) from when it is not (in more idiomatic cases).

    \item $NC_{Rand}$: the \textbf{random replacement controlled by frequency} is a two word expression in which the words are chosen to match the frequencies of the components of the target NC. The frequency values were extracted from corpora (in this case ukWaC and brWaC) as follows: we averaged the frequency of each NC and of its components ($f_{avg}=(f_{NC} + f_{w1} + f_{w2})/3$), and extracted the compound with the closest average value (e.g. \emph{police car} and \emph{supermarket city}). For each NC, 5 random replacements were used for each sentence. These pairs are used as controls to determine the lowerbound similarities for the target NCs, avoiding the potential impact of any differences in frequency.

\end{itemize}

The NCs were pre-selected by experts trying to maintain a balance between the 3 classes (idiomatic, partial, and compositional)\footnote{The two-word compounds were selected to be representative cases of compositional NCs (meaning related to the two words), partly idiomatic (meaning related to one of the words) and idiomatic (meaning unrelated to either of the two words), as our aim is to investigate to what extent the degree of idiomaticity affects the ability of models to generate an accurate representation. 
For English, the dataset contains 103, 88, and 89 idiomatic, partial, and compositional expressions respectively, while for Portuguese it has 60 NCs per class.}, and they appear in the context of three \textbf{naturalistic sentences} ($Nat$) from corpora that exemplify the same compound sense \cite{garcia-etal-2021-assessing}. Using Amazon Mechanical Turk (for English) and a dedicated custom built online platform (for Portuguese) compositionality scores for each NC and its components were obtained following the procedure of \citet{reddy-etal-2011-empirical} and \citet{cordeiro-etal-2019-unsupervised}. A \textit{Likert} scale from 0 (idiomatic) to 5 (compositional) was used for the human judgements, and the resulting scores were aggregated from the average of the different annotators \cite{garcia-etal-2021-assessing}.\footnote{On average, the compositionality scores were of 0.95/2.34/4.13 for English, and of 1.52/2.46/3.61 for Portuguese (idiomatic/partial/compositional).} The annotators also provided synonyms or paraphrases for the NCs in these sentences, which were used by language experts to manually generate the $NC_{Syn}$ variants \cite{garcia-etal-2021-assessing}. These annotations, including the synonyms, were collected at two levels of granularity: a more fine-grained token level, where annotations for each sentence are collected individually, and a more rough-grained type level, where a single annotation for each NC is collected considering all three sentences at once \cite{garcia-etal-2021-probing}. This allows for analyses of the impact of each individual context in the interpretation of the NC. A total of 8,725 annotations was obtained for English (421 annotators, each labelling an average of 21 sentences, resulting in a 10.4 annotations per sentence). In Portuguese, 5,091 annotations were provided by 33 annotators (with an average of 154 annotated sentences per annotator, and 9.4 annotations per sentence).

\begin{table*}[!ht]
    \centering
    \begin{tabular}{|l|l|m{10cm}|}
    \hline
    {\bf \#} & {\bf NC} & {\bf Sentence}\\
    \hline
    1 & Original  & John Paul II was an effective \textit{front man} for the catholic church. \\
    2 & $NC_{Syn}$  & John Paul II was an effective \textit{representative} for the catholic church. \\
    3 & $NC_{WordsSyn}$ & John Paul II was an effective \textit{forepart woman} for the catholic church. \\
    4 & $NC_{Comp}$ & John Paul II was an effective \textit{man} for the catholic church. \\
    &  & John Paul II was an effective \textit{front} for the catholic church. \\
    \hline
    
    5 & $NC_{Rand}$ & John Paul II was an effective \textit{battlefront serviceman} for the catholic church. \\
    \hline
    \end{tabular}
    \caption{Naturalistic sentence containing the NC \textit{front man} (in row 1) forming minimal pairs with sentences in rows 2-4, and with control random baselines in row 5.\label{tab:sent_examples}}
\end{table*}

In addition, NCIMP also contains \textbf{sense-neutral sentences} ($Neut$) in which the NCs appear in uninformative contexts
containing only 5 words and following the pattern \emph{This is a/an $<$NC$>$} for English (e.g. ``This is an \emph{eager beaver}'') and the Portuguese equivalent \emph{Este/a \'{e} um(a) $<$NC$>$}.\footnote{NCIMP also contains a second longer pattern of uninformative neutral sentences (10 words in English and 9 in Portuguese) following the patterns \emph{This is what a/an $<$NC$>$ is supposed to be} and the Portuguese equivalent \emph{Isto \'e o que um/uma $<$NC$>$ deveria ser}, to measure the potential impact of the length of the neutral context and of the position of the NC in the sentence. As the two types of neutral sentences elicit similar results, in the paper we only present the results for the short neutral sentences.} These neutral contexts can be used to examine how much contextual information is added to a representation in the more informative naturalistic contexts. Moreover, as some NCs may have more than one meaning (e.g. \emph{fish story} as either the literal \emph{aquatic tale} or the idiomatic \emph{big lie}), they can also be used to determine the default usage elicited for the NC in the absence of any informative contextual clues, in particular, whether it leans towards an idiomatic or a literal sense, potentially serving as an indication of the predominant sense sampled during training.

Experts (native or near-native speakers with background in Linguistics) reviewed both the naturalistic and the sense-neutral sentences in the minimal pairs, editing them if needed for preserving grammaticality after substitution (e.g. revising gender, number and definiteness agreement with determiners and adjectives). However, some of the variants generated may be semantically nonsensical, especially those involving random replacements. Table~\ref{tab:sent_examples} displays an example with the original sentence in the first row and the relevant sentences for each of the minimal pairs in the other rows.

Finally, each NC was also annotated with frequency, Pointwise Mutual Information \cite{church-hanks-1989-word} and Positive Pointwise Mutual Information  values, calculated from the ukWaC (2.25B tokens, \citet{baroni2009wacky}) and brWaC corpora (2.7B tokens, \citet{wagner-filho-etal-2018-brwac}), which can serve as approximations for their familiarity and conventionality. 


\subsection{Word Representation Models}
\label{sec:models}

We evaluate representative static and contextualised models. For the former, we compare GloVe and Word2Vec, 
using the official models for English, and the 300 dimensions vectors for Portuguese \cite{hartmann-etal-2017-portuguese}.

For the latter, we evaluate a large set of models, including the Bi-LSTM-based ELMo \cite{peters-etal-2018-deep}, and several Transformer-based language models: BERT \cite{devlin-etal-2019-bert} and some of its variants, such as multilingual BERT (mBERT\footnote{\url{https://huggingface.co/google-bert/bert-base-multilingual-cased}}) \cite{pires2019multilingual}, multilingual DistilBERT (mDistilB\footnote{\url{https://huggingface.co/distilbert/distilbert-base-multilingual-cased}}) \cite{sanh2019distilbert} and multilingual Sentence-BERT (mSBERT\footnote{\url{https://huggingface.co/sentence-transformers/distiluse-base-multilingual-cased}}) \cite{reimers-gurevych-2019-sentence}.  The recent flagship model LLaMA2 \cite{llama} is also included in our experiments. OpenAI text embeddings \cite{neelakantan2022text} are included in the  evaluations at sentence-level as they can only be accessed by the API \footnote{\url{https://platform.openai.com/docs/guides/embeddings/what-are-embeddings}} rather than by direct inspection of the whole model, which would be required for analyses at NC-level. Therefore, the latter are not conducted for OpenAI text embeddings.
For ELMo, we use the small model provided by \namecite{peters-etal-2018-deep}, and for Portuguese we adopt the weights released by \namecite{quintaDeCastro-elmoPTweights}. For LLaMA2 and OpenAI's embeddings, we use the \emph{13B} version and \emph{text-embedding-ada-002} version, respectively. For all other contextualised models, we use the pre-trained weights publicly available through  Flair\footnote{\url{https://github.com/flairNLP/flair}} \cite{akbik-etal-2019-flair} and HuggingFace\footnote{\url{https://github.com/huggingface/transformers}} \cite{wolf-etal-2020-transformers}. For BERT-based models (and for DistilB in English), we report the results obtained both by the multilingual uncased (ML) and by monolingual models for English (large, uncased) and Portuguese (large, cased), all available through HuggingFace. 

\subsubsection{Sentence and NC Embeddings}
Embeddings for the whole sentence as well as for the NCs are generated by averaging the (sub)word embeddings\footnote{In our preliminary experiments, we tested various pooling strategies, including max pooling, min pooling, the CLS token from BERT, concatenation, and mean pooling. The performance was similar across these methods, but to maintain simplicity and avoid complications from variable vector lengths, we chose mean pooling for the reported experiments.} of the relevant tokens involved, according to the model: 
\begin{itemize}
\item for  static models, the word embeddings are derived directly from the vocabulary, with missing out-of-vocabulary words being ignored; 

\item for ELMo the output word embeddings are averaged, and the concatenation of its three layers is adopted;

\item for Transformer-based models, the word embeddings are generated by averaging the representations of the sub-tokens and we report results using the last four layers.\footnote{Extensive evaluation of the individual layers and their combination were performed, but as the results follow the trend of those reported here, they are not included in the paper.}

\end{itemize}

In general we adopt standard widely used configurations to determine what the landscape of results is before any task optimisation, even if alternative tokenisation approaches \cite{gow-smith-etal-2022-improving}, dedicated representations for MWEs as single-tokens \cite{cordeiro-etal-2019-unsupervised,phelps-etal-2022-sample} and different combinations of layers and weighting schemes \cite{DBLP:journals/corr/abs-1904-02954,vulic-etal-2020-improving,rogers-etal-2020-primer} may generate better results in downstream tasks. Additional configurations were also extensively analysed and as they produced qualitatively similar results, they are not included in the paper.

\subsection{Measuring idiomatic meaning}
The general premises of this work, shared by many similar investigations, are that:
\begin{enumerate}
\item {\bf Vector embeddings approximate meaning.} We assume that the vector embeddings produced by the models are representations of usages in a semantic space that can approximate meaning. Since there is no absolute reference frame for meaning in that space, the meaning of a word/sentence is always relative and it is evaluated in terms of its similarity to other relevant words/sentences in the same semantic space. 
\item {\bf Word/multiword/sentence representations are the combinations of the (sub)word representations}. We adopt as the meaning of a word, multiword expression, or of a sentence, the compositional combination of its components. In this paper we focus on the additive combination, summing/averaging the vector embeddings of each token in the word, expression or sentence, and summing/averaging the vector embeddings of the relevant layers, when more than one layer is used.
\item {\bf Similarity of meanings can be approximated by similarity of vectors.} Similarity is a measure of the proximity between two vector embeddings. Throughout this paper we adopt cosine similarity as the similarity metric.\footnote{Other compositional operations and measures of distance were also used during these analyses, but with qualitatively similar results, and  have been omitted from the paper.} As contextualised models provide different vector representations for the same linguistic expression in different contexts, its vector representations would be found among different clusters of meaning as it transitions between its meanings in different sentences, and this would be reflected by the similarity measures.
\end{enumerate}

\subsubsection{The probing strategies}
\label{sec:probes}

To evaluate how word representation models deal with idiomaticity, we propose  a probing strategy where a target item in a sentence,  in this case an NC, is  systematically replaced by a set of different paraphrases or probes (P), forming the minimal pairs discussed in section \ref{sec:NCI}. We then use similarity measures to compare the representation for the sentence before and after replacing NC by P. 
Given the focus on idiomaticity we select a set of probes specifically for the expected changes in meaning they would induce in a sentence, and we refer to these potential changes in meaning as Linguistic Predictions (LPs). If the representations generated by a model reflect these predictions, passing the probing tests, then we consider that particular model as capturing to some extent the idiomatic meaning in NCs. The idiomatic probes are defined as follows, where {\it Comp} is the average human annotation compositionality score:

\begin{itemize}
    \item {P$_{Syn}$} - The true synonym. The replacement is a single word or a two word compositional noun compound that represents closely the meaning of the target NC, forming the minimal pair $NC_{Syn}$. Linguistic Prediction: after the replacement, the resulting sentence should be a near perfect paraphrase of the original sentence. Therefore high similarities are expected for all minimal pairs independently of the degree of compositionality of the target NC, from the more idiomatic \emph{grey matter} (and \emph{brain}) to the more literal \emph{economic aid} (and \emph{financial assistance}), with no correlation expected with \emph{Comp}.

    \item {P$_{Comp}$} - The partial expression. The replacement is one of the component words of the target compound, and in particular we consider the one that preserves most of the meaning, forming the minimal pair $NC_{Comp}$. Linguistic Prediction: the resulting sentence may preserve some of the original meaning for more compositional cases, but not for idiomatic cases. Therefore, high similarities are only expected between minimal pairs involving compositional and partly compositional cases (e.g. \emph{economic aid} and \emph{aid}, \emph{crocodile tears} and \emph{tears}, but not for \emph{wet blanket} and \emph{blanket} or \emph{wet}), with some correlation expected with \emph{Comp}.

    \item {P$_{WordsSyn}$} - The literal synonyms of the individual NC components. The replacement is a two-word expression formed from frequent out-of-context synonyms for each of the component words of an NC when considered independently, forming the minimal pair $NC_{WordsSyn}$. Linguistic Prediction: after replacement, the resulting sentence may not preserve the meaning of the original sentence, especially for more idiomatic cases. Therefore,  higher similarities are only expected for minimal pairs involving more compositional NCs (e.g. \emph{wedding day} and \emph{marriage date} but not \emph{eager beaver} and \emph{restless rodent}), with a high correlation expected with  
    {\it Comp}. 
    
    \item {P$_{Rand}$} - The random replacement controlled by frequency. The replacement is a two word expression where the words are chosen to match the frequencies of the components of the target NC, forming the $NC_{Rand}$ minimal pair. Linguistic Prediction: after replacement, the resulting sentence should not preserve the meaning of the original sentence, independently of the level of idiomaticity of the original NC 
    (e.g. for \emph{police car} and \emph{supermarket city}), with no correlation expected with \emph{Comp}. 
\end{itemize}

For a more in-depth analysis of expected changes in meaning, 
we follow \citet{garcia-etal-2021-probing} comparing representations both at a macro sentence level and also at a micro NC level, analysing the representations of NC (and its variants P) extracted from the context of the sentence. 
Although any differences in meaning should be reflected both at sentence and at NC representation levels (only magnified in the latter), this comparison aims to highlight the impact of the level of granularity used when analysing idiomaticity.\footnote{Our prior work reveals that only looking at similarities at sentence level when comparing the representations of the original and the resulting sentences may not accurately reflect their differences 
 \cite{garcia-etal-2021-probing}.}

\subsection{Metrics}
\label{metrics}

\subsubsection{The Human  Compositionality score (Comp)}

Assuming a list of $N$ NCs, chosen to provide balanced test scenarios of different levels of idiomaticity,  
we denote NC$_\alpha$ with 
$\alpha=1,..., N$  the different NCs to be evaluated.  The meaning of theses NCs is exemplified by a set of $N \times M$ sentences 
$\mbox{Sent}_{ \alpha \beta} $ with $\alpha=1,..., N$ and  $\beta=1,...,M$ the sentence index. The dataset contains $M$=3 naturalistic sentences to exemplify the use of each NC (see section ~\ref{sec:NCI}), with each sentence annotated by human judges according to the compositionality of the target NC in the sentence.  The resulting scores are denoted Comp$_{\alpha \beta j}$, with  $\alpha=1,...,  \mbox{$N$}$,  $\beta=1,...,\mbox{$M$}$, and $j = 1,..., A_{\alpha \beta}$  where $A_{\alpha \beta}$ is the number of annotators for sentence Sent$_{\alpha \beta}$.  Comp$_{\alpha \beta j}$  are integer values derived from a Likert scale and range from 0 (totally idiomatic) to 5 (totally compositional).
We define the compositionality score for  a specific NC$_\alpha$ as the average of the annotations for sentences Sent$_{\alpha \beta}$, 
\begin{equation}
 \mbox{Comp}(\mbox{NC}_\alpha) =  \left\langle \left\langle \; \mbox{Comp}_{\alpha \beta j} \right\rangle_{Annot} \right\rangle_{Sent} 
\label{goldstandard} 
\end{equation} where $ \left\langle \cdots \right\rangle_{Sent} $ are averages on sentences and $ \left\langle \cdots \right\rangle_{Annot} $ averages on annotations.
These average values are the gold standard in this work.

\subsubsection{The Similarity score (Sim)}

Probing the meaning of a compound NC$_\alpha$ in a sentence Sent$_{\alpha \beta}$  requires the generation of a new set of modified sentences  Sent$_{Pi \;\beta  \gamma }$ where NC$_\alpha$ is replaced by a probe P$i$ (discussed in section \ref{sec:probes}). 
We measure
the effect of the probe substitution  directly from the similarity between the  representation of the original expression, X, and the representation of the new expression after substitution, Y, adopting, throughout this paper, cosine similarity as a measure of the similarity of meaning between two vector embeddings. 
\begin{equation}
   \mbox{cossim}(X,Y) = \frac{ \epsilon_X \cdot \epsilon_Y }{ || \epsilon_X || \; || \epsilon_Y  ||}
\end{equation}
where $\epsilon_X$ and $\epsilon_Y$ are vector embeddings of D components, $\epsilon_X \cdot \epsilon_Y$  their inner products,
and $|| \epsilon_X || $, $|| \epsilon_Y || $ are their L2 norms. Therefore the average similarity between the original expression and the probe-modified expression for a given NC can be defined as
\begin{equation}
 \mbox{Sim}(\mbox{P}i, Target)  = \left\langle  \; \mbox{cossim} ( expr(\mbox{P}i), expr(\mbox{NC})) \;\right\rangle_{Pi}
\end{equation}
where $expr(\mbox{NC})$ is the target NC expression, and $expr(\mbox{P}i)$ is the expression where NC is replaced by a probe of the type P$i$, and $\left\langle \cdots \right\rangle_{Pi}$ means the average over possible substitutions of this type. We use more than one substitution only for random probes ({P$_{Rand}$}),
for all other probes a single substitution is reported.

\subsubsection{The Affinity score (Aff)}\label{sec:affinity}
Cosine similarity measures are not sensitive enough to capture subtle meaning differences, especially in anisotropic representation spaces \cite{ethayarajh-jurafsky-2021-attention}. 
Additionally, 
there may be a `horizon of interest,' beyond which word connections lose meaningful inference  \cite{karlgren2021semantics}, which may be a challenge for representing idiomatic expressions, as the necessary context may lie within this critical boundary. Investigating measures that account for anisotropic spaces and for a horizon of interest are interesting avenues for future research for improving idiomaticity detection.
In this paper, we propose a comparative measure that we refer to as Affinity (Assessment of Feature Familiarity and Idiomatic Nuance by Interpreting Target Yielding), that identifies which between two representations is the closest to a given target representation.

Given a target representation $Target$ and two possible probes P$i$ and P$j$, the affinity is defined as: 

\begin{equation}
\mbox{Aff}
(\mbox{P}i,\mbox{P}j\;|\;Target) = \mbox{Sim}\left( \mbox{P}i , Target \right) - \mbox{Sim}\left(   \mbox{P}j  , Target \right)\;. 
\end{equation}  
Affinities closer to 1 or larger indicate a greater similarity between the target and the first probe P$i$,  values closer to -1 or lower indicate the opposite situation where the target is more similar to the second probe P$j$, and values near zero indicate no preference. 
Given the focus of this paper on detecting idiomaticity in representations, we measure the affinities involving the minimal pairs defined in section \ref{sec:probes}, 
analysing if, as expected, the target NCs have higher similarities with probes with substitutions that maintain the original meaning as P${i}$ than with probes that involve potential changes in meaning as P${j}$. In particular: 

\begin{itemize}
    \item Affinity $ A_{Syn|WordsSyn} =\mbox{Aff}(\mbox{P}_{Syn}, \mbox{P}_{WordsSyn} | 
 \mbox{NC} )$ measures if the target NCs have greater similarities with their gold synonyms than with synonyms of the individual components (e.g. \emph{eager beaver} with \emph{hardworking person} than with \emph{restless rodent}).  
    
    \item Affinity $A_{Syn|Rand}=\mbox{Aff}(\mbox{P}_{Syn}, \mbox{P}_{Rand} | 
 \mbox{NC} )$ 
 compare if the target NCs display greater similarities to their gold synonyms than to random substitutions.

\end{itemize}

Our Affinity measure extends traditional forced-choice evaluations \cite{warstadt-etal-2020-blimp-benchmark} by quantifying the degree of similarity preference between two options. Unlike binary choices, Affinity provides a continuous measure of relative similarity, offering a more detailed assessment of how well models capture idiomatic meanings. This nuanced analysis reveals subtle differences in model performance, providing deeper insights into the representation of idiomatic expressions.

\subsubsection{The Scaled Similarity score (Sim$_R$)}

Even though Affinity is an advance over the simple similarity measure,  
additional measures may still need to be adopted for models if the average similarity between two random embeddings is larger than zero, as affinities will tend to have small values even for very dissimilar probes (see discussion). To address this issue, we propose a scaled version of the similarity:
\begin{equation}
\mbox{Sim}_R(\mbox{P}i | Target) = \left\langle \  \frac{ \mbox{Sim}(\mbox{P}i, Target) -  \; \mbox{Sim}(\mbox{P}_{Rand}, Target)  }{1- \;\mbox{Sim}(\mbox{P}_{Rand}, Target)  } \;\right\rangle_{Sent}
 \label{simR} 
\end{equation} 
where $\langle \cdots \rangle_{Sent}$ denotes the average over the $M$ sentences that illustrate the meaning of a particular NC and P$_{Rand}$ is a random substitution.
The scaled similarity is defined such that if replacing the target with a probe P$i$  results in cosine similarities close to one ($\text{Sim}(\text{P}i, Target) \approx 1$), the scaled similarity is also close to one, $ \text{Sim}_R \approx 1 $. Conversely, if the replacement is similar to a random replacement ($\text{Sim}(\text{P}i, \text{Target}) \approx \text{Sim}(\text{P}_{Rand}, Target) $), then \( \text{Sim}_R \approx 0 \). This approach is equivalent to a max-min normalisation\footnote{Given a value $ x $ in a dataset, the max-min normalisation of $ x $ is calculated as follows:
$$x' = \frac{x - \min(x)}{\max(x) - \min(x)} .$$} in the anisotropic space of a model.

In particular, given the focus on idiomaticity, we focus as before on two similarities:
\begin{itemize}
\item $\mbox{Sim}_{R|Syn} = \mbox{Sim}_R(\mbox{P}_{Syn}|\mbox{NC})$, where the NCs are replaced by gold synonyms and no changes in meaning are expected, therefore {$\mbox{Sim}_{R|Syn}$} should be close to 1.
\item $\mbox{Sim}_{R|WordsSyn} = \mbox{Sim}_R(\mbox{P}_{WordsSyn}|\mbox{NC})$, where the NCs are replaced by synonyms of the individual components and greater changes in meaning, and therefore small values ($\sim 0$) of $\mbox{Sim}_R$, are expected for more idiomatic cases. 
\end{itemize}

\subsubsection{The Correlation measure ($\rho$)}

Finally, to assess the impact of idiomaticity for the probe substitutions we use Spearman Correlation between the different measurements and the gold standard human annotations of compositionality (Comp) 
given by Eq. \ref{goldstandard}.

\section{Probing for Idiomaticity}
\begin{figure}[!ht]
\begin{subfigure}{0.5\textwidth}
  \centering

{P$_{Syn}$ \hspace{2cm} \vspace{5pt}}
\includegraphics[width=0.8\textwidth,height=0.8\textwidth]{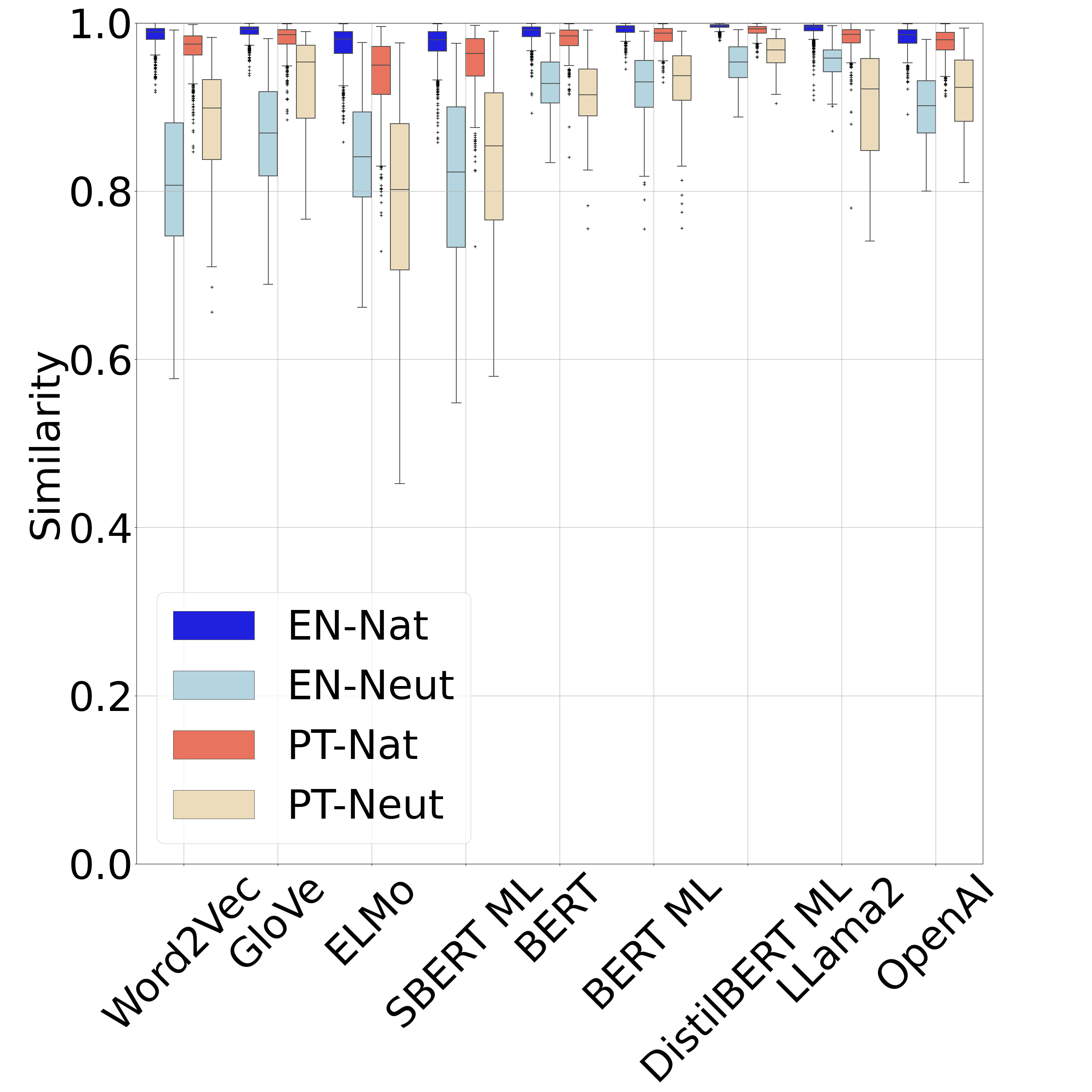}  
\end{subfigure}
\begin{subfigure}{0.5\textwidth}
  \centering

{P$_{Comp}$ \hspace{2cm}} \vspace{5pt}
\includegraphics[width=0.8\textwidth,height=0.8\textwidth]{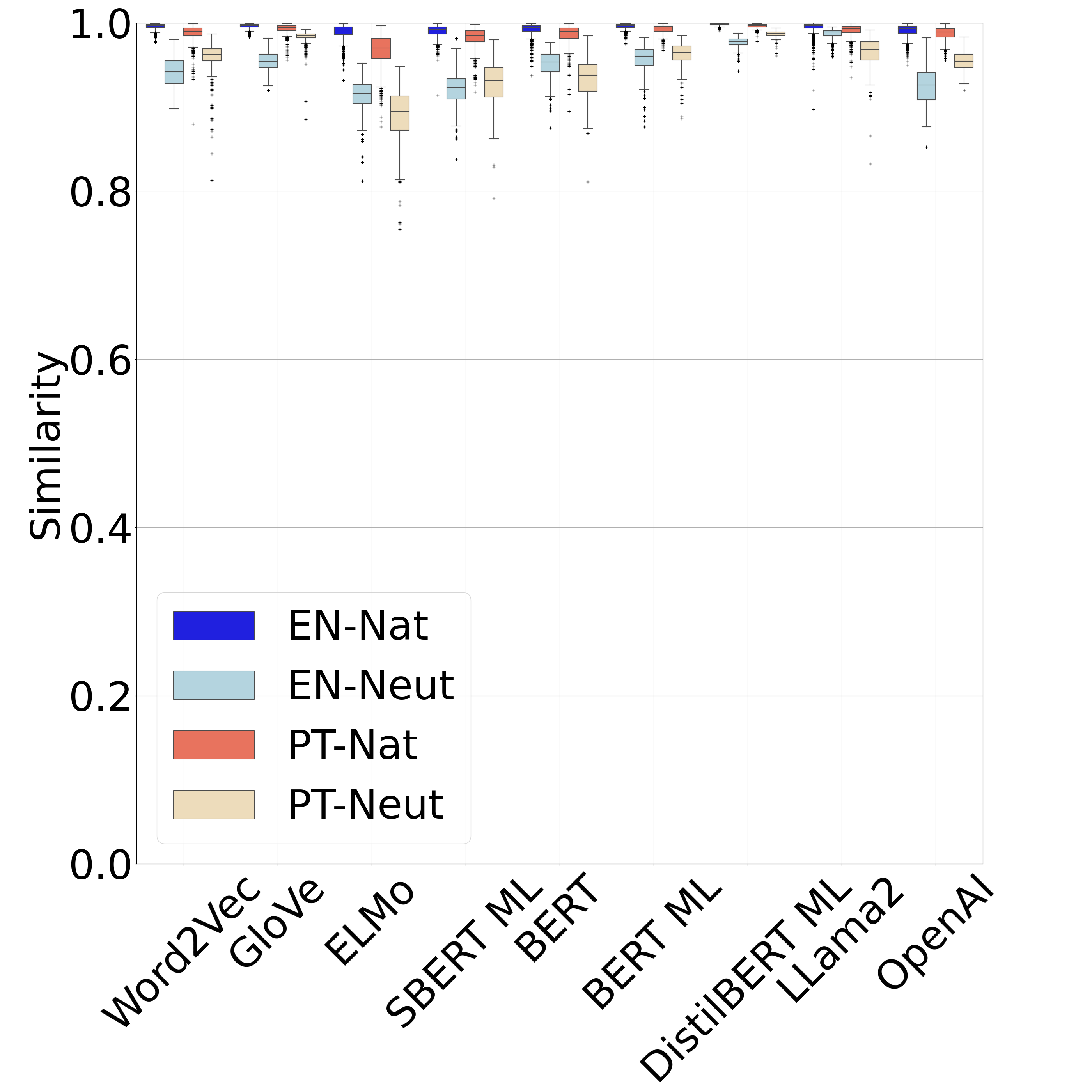}  
\end{subfigure}
\begin{subfigure}{0.5\textwidth}
  \centering

{P$_{WordsSyn}$ \hspace{2cm}} \vspace{5pt}
\includegraphics[width=0.8\textwidth,height=0.8\textwidth]{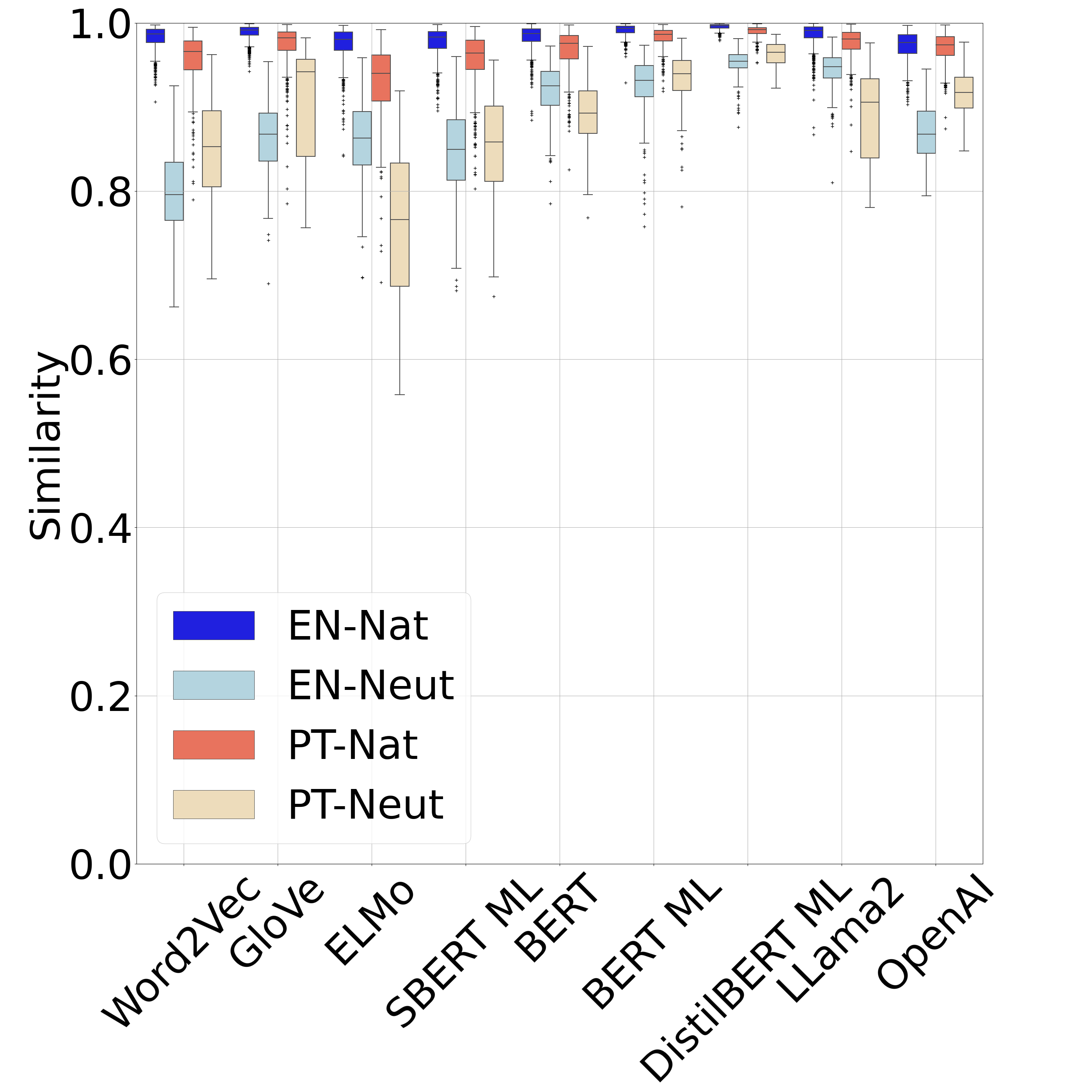}  
\end{subfigure}
\begin{subfigure}{0.5\textwidth}
  \centering

{P$_{Rand}$ \hspace{2cm}} \vspace{5pt}   \includegraphics[width=0.8\textwidth,height=0.8\textwidth]{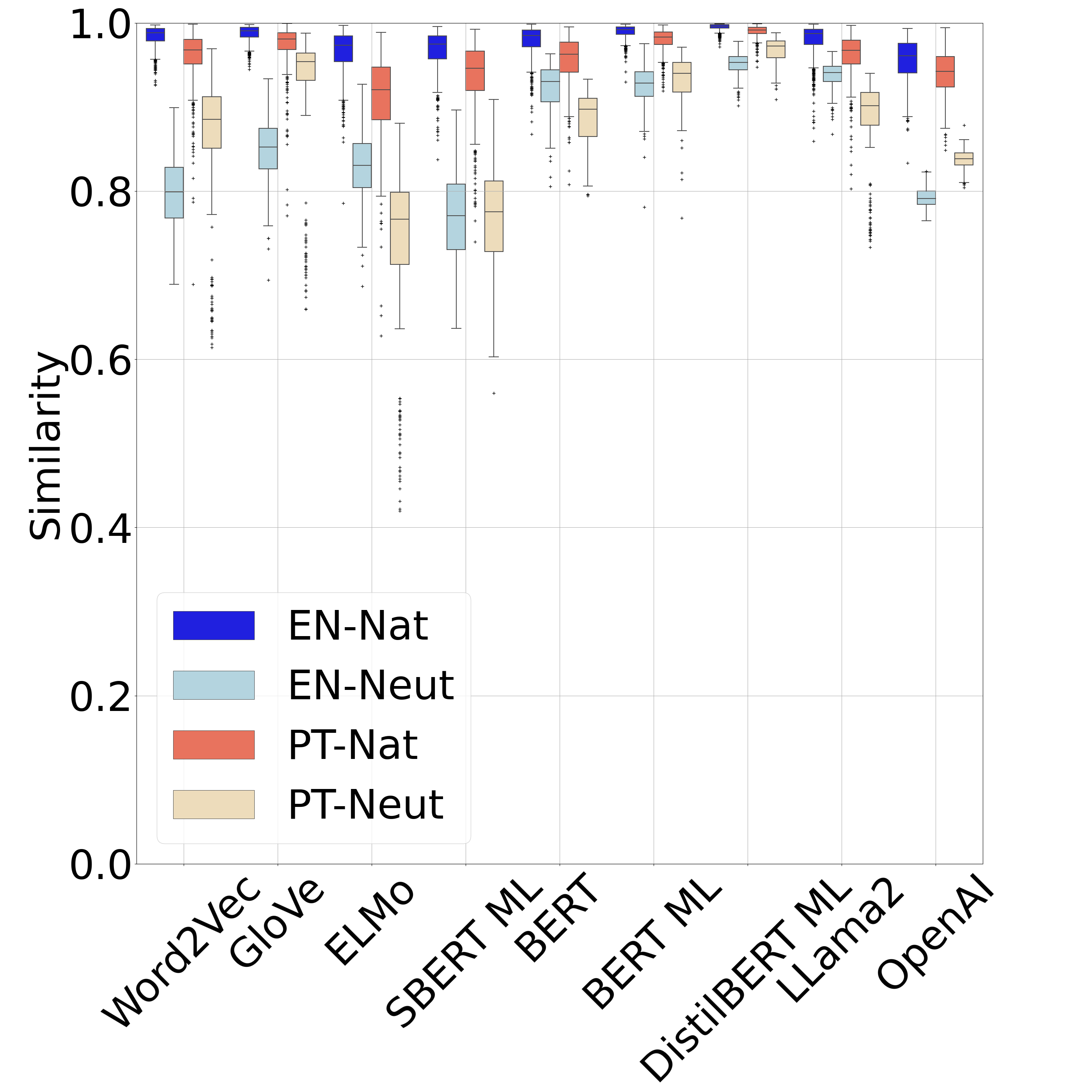}  
\end{subfigure}

\begin{subfigure}{0.5\textwidth}
       \centering
        Ideal Values \vspace{5pt}  \hspace{-5.5pt}    \includegraphics[width=0.77\textwidth,height=0.7\textwidth]
        {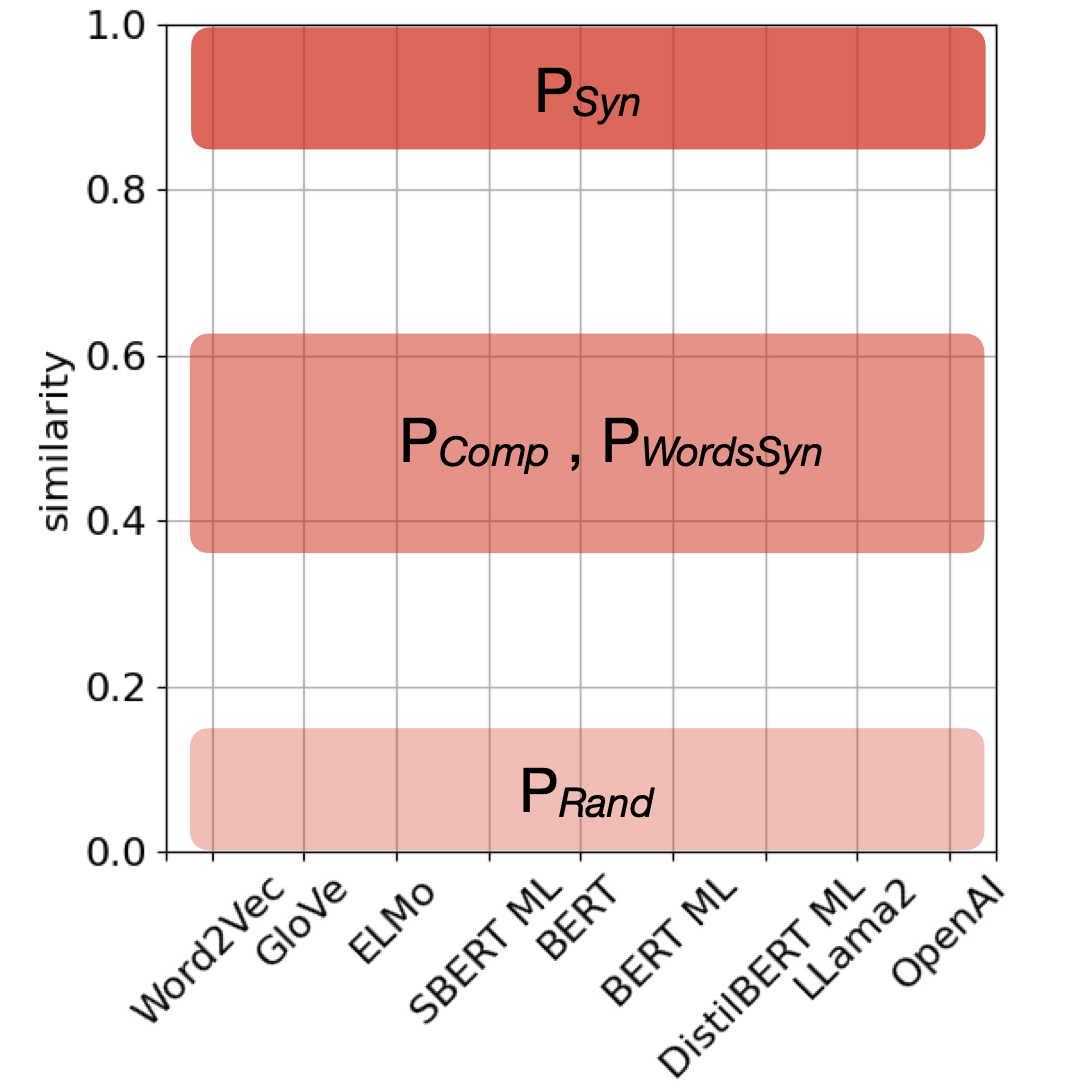} 
\end{subfigure}
\caption{Distribution of cosine similarities between the minimal pairs at sentence level, with the original NC and the probe-modified substitution for English (EN, in blue) and Portuguese (PT, in orange), with naturalistic (Nat) sentences in darker shade and neutral (Neut) in lighter. The lower panel (Ideal Values) is an illustration of similarity values ideally expected for the different probes. The means and standard deviations are in Table \ref{tab:num_sent} in the Appendix.}
\label{fig:p1-3-sent} 
\end{figure}

\subsection{Are the representations of the NCs and their synonyms similar?}
\label{sec:results_aff}

A first indication of the successful modeling of idiomaticity is if a model assigns similar representations for the target NCs and for their synonyms, regardless of their level of compositionality. We measure this using the minimal pairs of probe P$_{Syn}$ and compare it with less appropriate substitutions represented by the other probes P$j$.
The distribution of similarities obtained for each of the probes is shown in Figure \ref{fig:p1-3-sent}, along with the correlations of these similarities with the human compositionality scores for the NCs at sentence ($\rho_{Sent}$) and NC ($\rho_{NC}$) levels, in Tables \ref{tab:results-1} and \ref{tab:results-2}. Considered in isolation, 
the high similarity scores for {P$_{Syn}$} at sentence level (close to 1 for naturalistic sentences, and mostly above 0.75 for neutral sentences, Figure \ref{fig:p1-3-sent}({P$_{Syn}$})) 
seem to suggest that these models are able to capture idiomaticity. However, when compared against the scores for the minimal pairs of the other probes a different story emerges. 

When the components of a target NC are replaced
with one of their component words (Figure \ref{fig:p1-3-sent}({P$_{Comp}$})) or
with the synonyms of their component words  (Figure \ref{fig:p1-3-sent}({P$_{WordsSyn}$})), lower similarities should be observed between the minimal pairs since, although these substitutions could preserve some of the meaning of the more compositional cases, they would not do so for the more idiomatic cases. Moreover, random substitutions should lead to even lower similarities for all NCs (Figure \ref{fig:p1-3-sent}({P$_{Rand}$})),  
since they could result in nonsensical sentences.
This expected staggered pattern of similarities, highest for {P$_{Syn}$}, moderate for {P$_{Comp}$} and {P$_{WordsSyn}$}, and lower for {P$_{Rand}$}, illustrated in Figure \ref{fig:p1-3-sent}(Ideal Values) does not seem to be reflected by a visible reduction of the similarities at sentence level, in Figures \ref{fig:p1-3-sent}. In fact, even random substitutions seem to result in high sentence similarities, even if they are not as high the other substitutions.  

Another important point relates to the correlation of these similarities for the different NCs with human judgments for compositionality. It is expected that there would be almost no correlation for the similarities derived from {P$_{Syn}$} and {P$_{Rand}$}, and a moderate correlation for {P$_{Comp}$} and {P$_{WordsSyn}$}, as they may be more acceptable for compositional NCs than for idiomatic ones. However, this expected pattern is not observed in the results presented in Table \ref{tab:results-1}.
For most models, $\rho_{Sent}$ ({P$_{Syn}$}) shows moderate correlation, while $\rho_{Sent}$ ({P$_{Comp}$}) and $\rho_{Sent}$ ({P$_{WordsSyn}$}) are either weak or non-significant.

\begin{table}[!ht]
\centering
\begin{footnotesize}
\begin{tabular}{|l|ccccccccc|}
\hline
$\rho_{Sent}$ 
   & Word2Vec  & GloVe & ELMo  & SBERT& BERT  & BERT  & DistilB& LLaMA2& OpenAI  \\ 
&   &  &   & ML &   & ML  & ML & &   \\
   \hline
{P$_{Syn}$}   
   &      &       &       &       &       &        &         &       &        \\
EN-Nat 
& 0.30 & 0.31  & 0.43  & 0.47  & 0.39  & 0.51   & 0.38    & 0.15  & 0.41   \\
EN-Neut & 0.60 & 0.58  & 0.55  & 0.60  & 0.51  & 0.53   & 0.56    & 0.37  & 0.54   \\
PT-Nat & 0.18 &  0.13 & 0.33  & 0.31  & 0.32  & 0.29   & 0.20    & 0.27  & 0.46   \\
PT-Neut & 0.31 & 0.22  & 0.37  & 0.46  & 0.35  & 0.30   & 0.31    & 0.31  & 0.51   \\ \hline
{P$_{Comp}$} 
   &      &       &       &       &       &        &         &       &        \\
EN-Nat   & -    & -     & -     & -     & 0.17  & -      & -       & -     & 0.37   \\
EN-Neut & 0.19 & 0.29  & -     & -     & -     & -      & -0.12   & -     & 0.51   \\
PT-Nat &  -   & -0.12 & 0.12  & -     & 0.16  & -      & -0.15   & -     & 0.21   \\
PT-Neut & 0.13 & -     & 0.17  & -     &  -    & -0.14  & -       & -     & 0.27   \\ \hline
{P$_{WordsSyn}$} 
   &      &       &       &       &       &        &         &       &        \\
EN-Nat   & -    & -     & -     & -     & -     & -      & -       & -     & 0.21   \\ 
EN-Neut & 0.19 & -     & -     & -0.13 & -0.15 & -      & -       & 0.20  & 0.13   \\ 
PT-Nat & -0.12& -0.19 & -     & -     & -     & -      & -0.14   & -     & 0.11   \\ 
PT-Neut & -    & -0.13 & -     & -     & -     & -      & -       & -     & 0.17  \\ 
 \hline
{P$_{Rand}$}   
   &      &       &       &       &       &        &         &       &        \\
EN-Nat   & -    &-0.11  & -0.13 & -0.16 & -0.27 & -0.11  & -0.18   & -0.11 & -      \\
EN-Neut & 0.11 & -     & -0.31 & -0.36 & -0.29 & -      & -0.13   & -     & -      \\   
PT-Nat & -0.17& -0.20 & -0.13 & -0.11 & -0.14 & -0.12  & -       &-0.18  & -      \\
PT-Neut & 0.13 & -0.17 & -0.14 & -0.11 & -0.22 & -0.11  & -       & -     & -     \\ \hline
\end{tabular}
\end{footnotesize}
\caption{Spearman $\rho$ correlation between cosine similarities and human compositionality judgments (Comp) at sentence level. Only significant results (p$\leq$0.05) are displayed, for P$_{Syn}$, P$_{Comp}$, P$_{WordsSyn}$ and P$_{Rand}$, for English (EN) and Portuguese (PT), naturalistic (Nat) and neutral (Neut) sentences. }
\label{tab:results-1}
\end{table}

\begin{table}[!ht]
\centering
\begin{footnotesize}
\begin{tabular}{|l|cccccccc|}
\hline
$\rho_{NC}$ 
   & Word2Vec  & GloVe & ELMo  & SBERT& BERT  & BERT  & DistilB& LLaMA2  \\
   &   &  &   & ML &   & ML  & ML &    \\
   \hline
{P$_{Syn}$}   
   &      &       &       &       &       &        &         &          \\
EN-Nat & 0.62 & 0.62  & 0.60  & 0.66  & 0.39  & 0.67   & 0.58    & 0.36     \\
EN-Neut & 0.62 & 0.61  & 0.60  & 0.65  & 0.34  & 0.58   & 0.54    & 0.37     \\
PT-Nat & 0.45 &  0.40 & 0.47  & 0.48  & 0.57  & 0.44   & 0.39    & 0.37     \\
PT-Neut & 0.43 & 0.41  & 0.47  & 0.48  & 0.48  & 0.35   & 0.37    & 0.31     \\ \hline
{P$_{Comp}$} 
   &      &       &       &       &       &        &         &          \\
EN-Nat & 0.20 & 0.45  & 0.17  & 0.34  & 0.17  & 0.35   & 0.26    & 0.15     \\
EN-Neut & 0.20 & 0.44  & 0.23  & 0.28  & -0.31 & -      & -       & 0.12     \\
PT-Nat & 0.30 &  0.20 & 0.29  & 0.11  & 0.43  & 0.16   & -       & 0.22     \\
PT-Neut & 0.27 & 0.18  & 0.21  & -     & 0.24  & -      & -       & 0.13     \\ \hline
{P$_{WordsSyn}$}
   &      &       &       &       &       &        &         &          \\
EN-Nat & -    & 0.18  & -     & -     & -0.40 & 0.21   & 0.15    & 0.29     \\ 
EN-Neut & 0.11 & 0.18  & -     & -     & -0.40 & -      & -       & 0.22     \\ 
PT-Nat & -    & -     & 0.13  & -     & 0.17  & 0.11   & -       & -        \\ 
PT-Neut & -    & -     & -     & -     & 0.14  & -      & -       & -        \\  \hline
{P$_{Rand}$}   
   &      &       &       &       &       &        &         &          \\
EN-Nat & 0.11 & 0.18  & -0.18 & -0.23 & -0.58 & -0.22  & -0.29   & -        \\
EN-Neut & 0.12 & 0.18  & -0.21 & -0.20 & -0.49 & -      & -0.24   & 0.13     \\ 
PT-Nat & -    & -     & -     & -     & -     & -      & -       &-         \\
PT-Neut & -    & -     & -     & -     & -     & -0.11  & -       & -        \\ \hline
\end{tabular}
\end{footnotesize}
\caption{Spearman $\rho$ correlation between cosine similarities and human compositionality judgments (Comp) at NC level. Only significant results (p$\leq$0.05) are displayed, for P$_{Syn}$, P$_{Comp}$, P$_{WordsSyn}$ and P$_{Rand}$, for English (EN) and Portuguese (PT), naturalistic (Nat) and neutral (Neut) sentences. }
\label{tab:results-2}
\end{table}

\begin{figure}[!ht]
\begin{subfigure}{0.5\textwidth}
    \centering
{P$_{Syn}$ \hspace{2cm}} \vspace{5pt} 
\includegraphics[width=0.8\textwidth,height=0.8\textwidth]{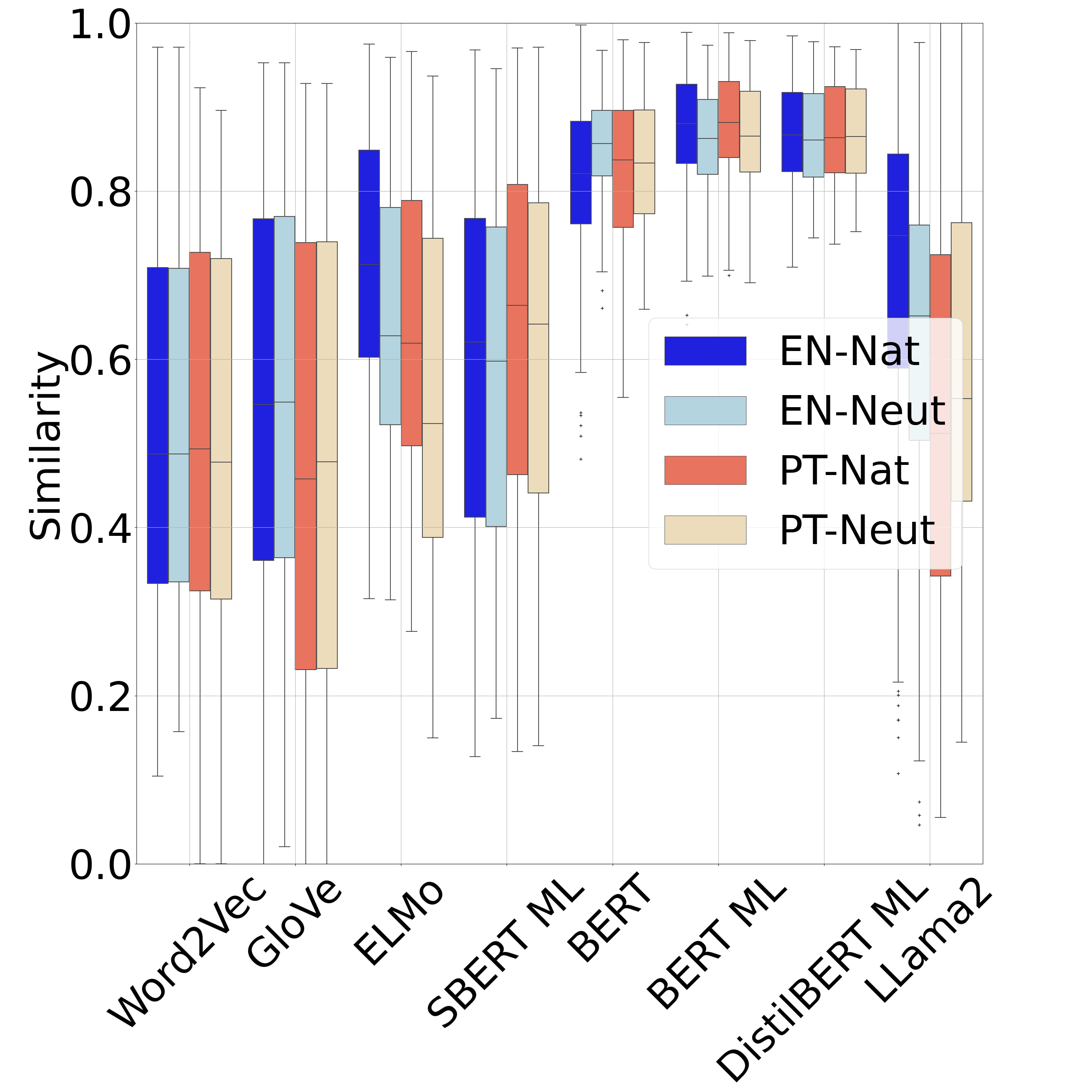}
 \end{subfigure}
\begin{subfigure}{0.5\textwidth}
    \centering
{P$_{Comp}$ \hspace{2cm}} \vspace{5pt}
\includegraphics[width=0.8\textwidth,height=0.8\textwidth]{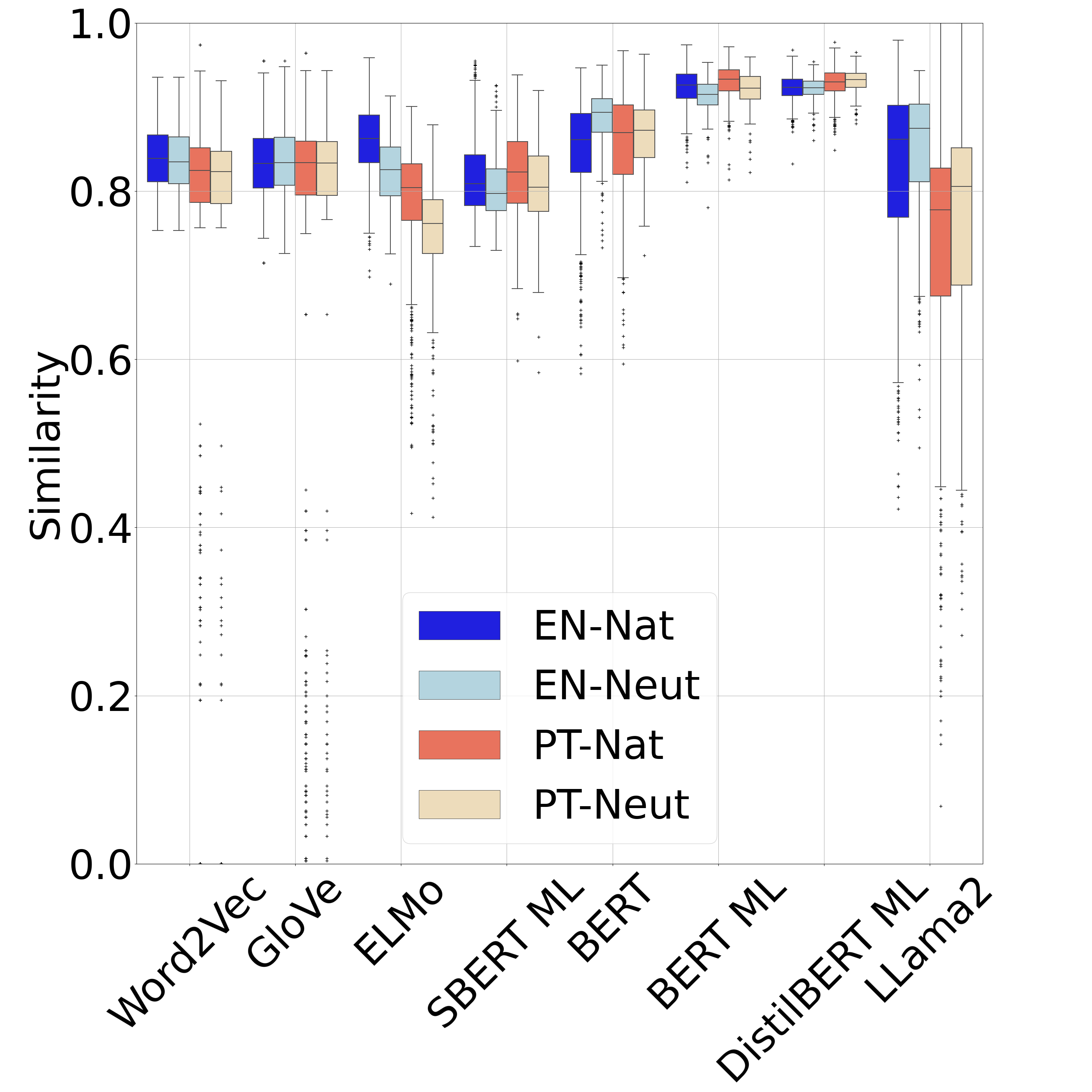}
 \end{subfigure}
\begin{subfigure}{0.5\textwidth}
    \centering

{P$_{WordsSyn}$ \hspace{2cm}} \vspace{5pt}
\includegraphics[width=0.8\textwidth,height=0.8\textwidth]{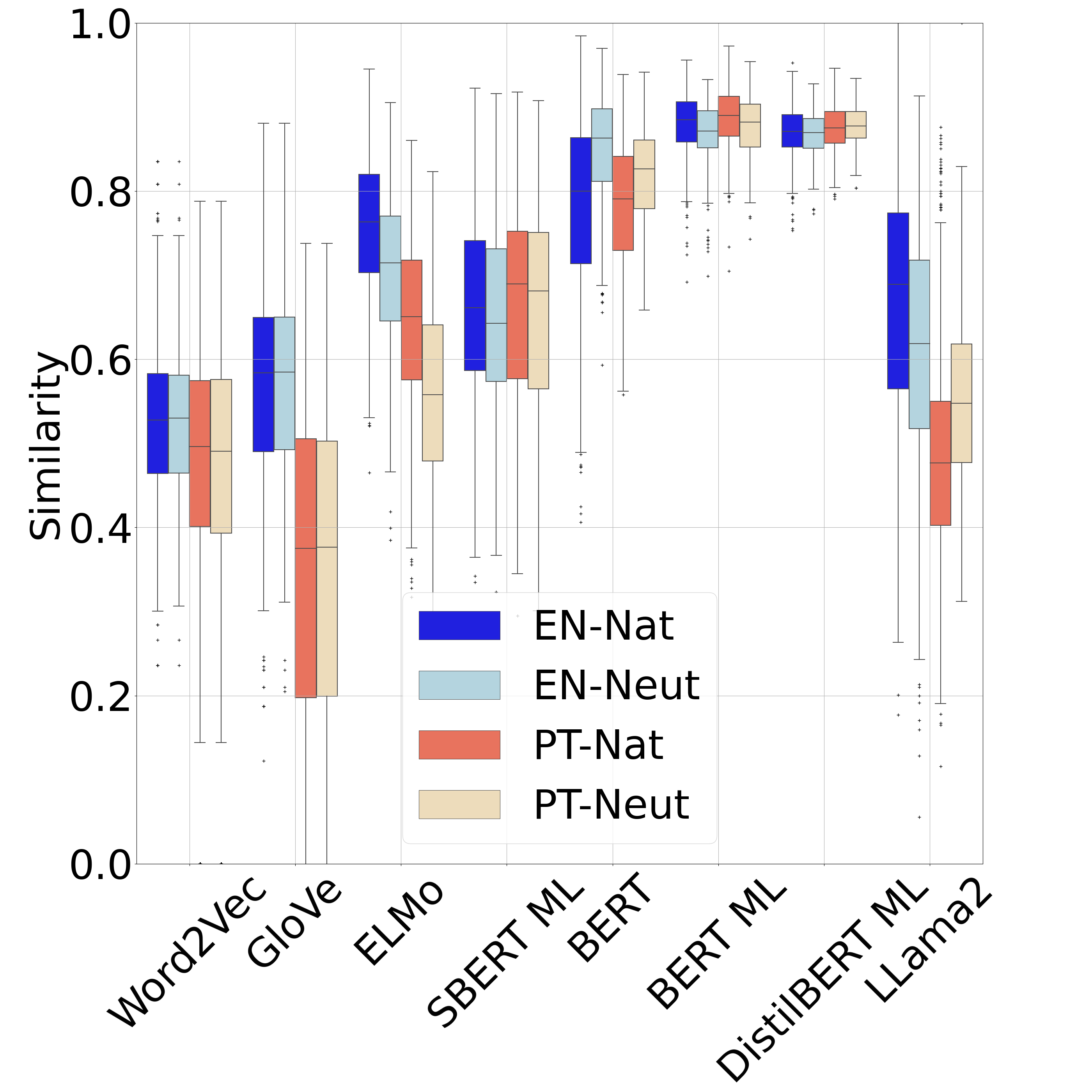}
 \end{subfigure}
\begin{subfigure}{0.5\textwidth}
    \centering
{P$_{Rand}$ \hspace{2cm}} \vspace{5pt}
\includegraphics[width=0.8\textwidth,height=0.8\textwidth]{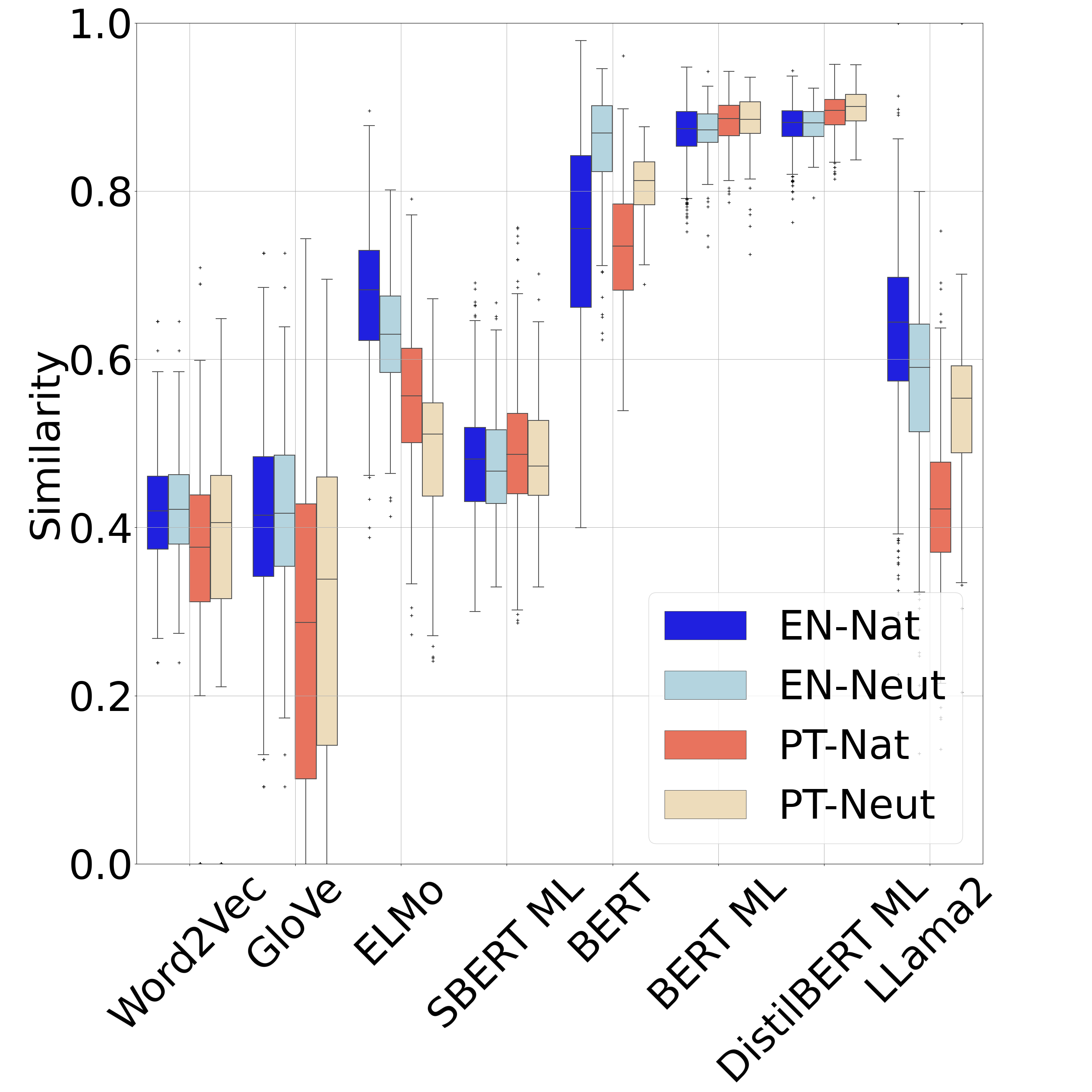}
 \end{subfigure}

\begin{subfigure}{0.5\textwidth}
       \centering
        Ideal Values\vspace{5pt}      \includegraphics[width=0.77\textwidth,height=0.7\textwidth]
        {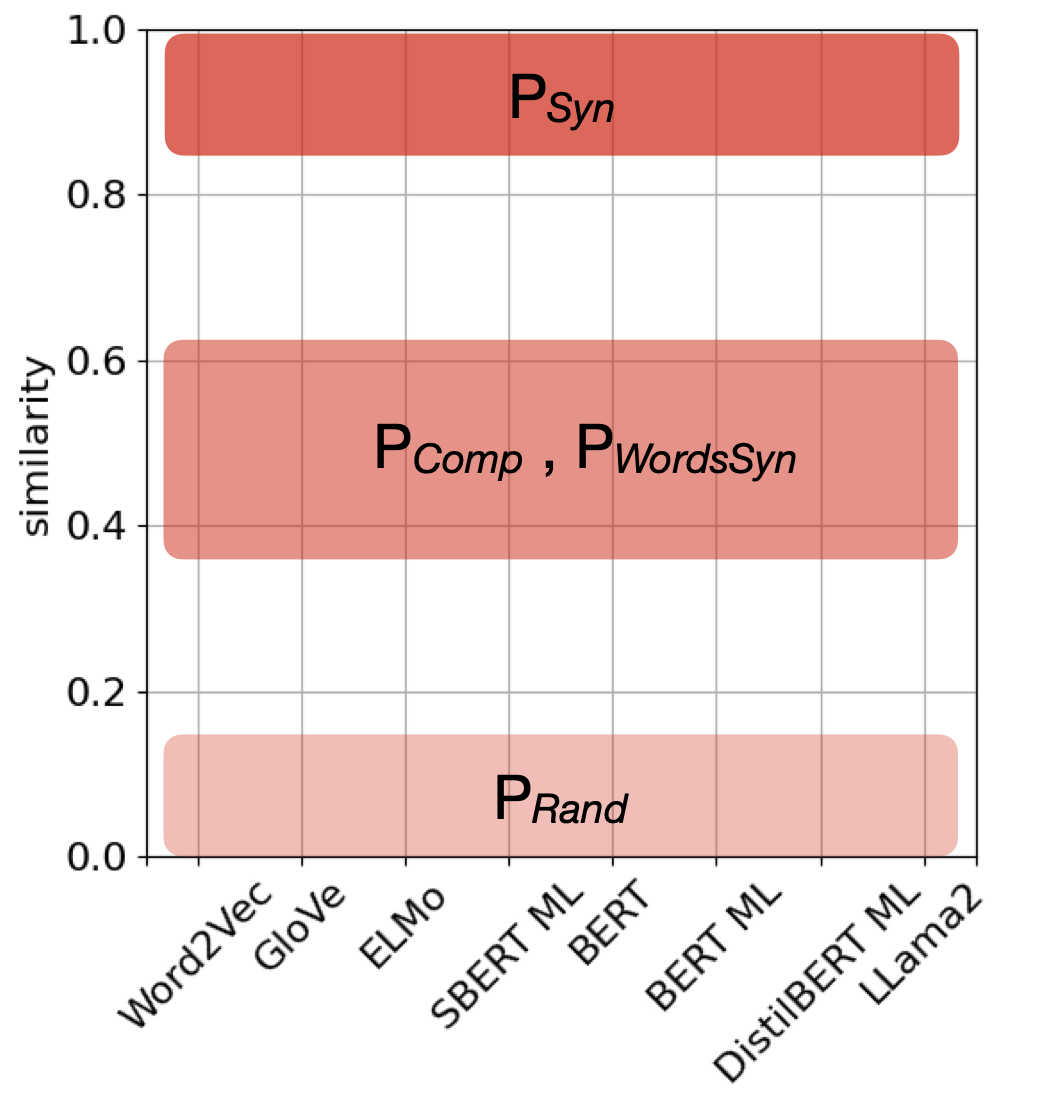} 
\end{subfigure}
\caption{Distribution of cosine similarities between the minimal pairs at NC level, with the original NC and the probe-modified substitution for English (blue) and Portuguese (orange), with naturalistic sentences in darker shade and neutral in lighter.  
The lower panel (Ideal Values) is an illustration of similarity values ideally expected for the different probes. The means and standard deviations are in Table \ref{tab:num_sent} in the Appendix.}
\label{fig:p1-3-NC} 
\end{figure}

Since in these minimal pairs only the target NCs and their substitutions change, the high similarities found may be an effect of the lexical overlap between the sentences of a minimal pair. Indeed, comparing the output of the models in relation to sentence lengths for  naturalistic sentences, there is a significant moderate to strong positive correlation between the lexical overlap and the cosine similarity of a pair, for both English and Portuguese  (Table \ref{tab:corr_sent_length}), where the greater the overlap between the sentences, the higher their similarity.  
This can also explain the higher similarities observed for naturalistic than for neutral sentences, since the former are longer than the latter with a higher lexical overlap proportional to the length of the sentence:  average sentence length for naturalistic sentences is 23.4 words for English (lexical overlap $>$ 91\%) and 13.0 words for Portuguese  (overlap $>$ 84\%), while for the neutral sentences it is five words (overlap $>$ 60\%) for both languages.\footnote{We also compared longer neutral contexts with 10 words for English ($>$ 80\%), and 9 words for Portuguese ($>$ 77\%), and found similar results.}  It could be argued that the influence of lexical overlap is expected, given that a compositional representation is used for sentences, where the embeddings for each token are added. However, while this holds true for static models, it may not necessarily apply to contextualised models. In contextualised models, it is expected that each token/word would interact with others via attention heads, and if the model accurately captures semantics, all tokens/words will adjust to the context of the sentence as a whole. Ideally, even with the simple compositional representation of the sentence, we would anticipate that a correct sentence would exhibit low similarity with the mostly nonsensical sentences produced by random probes.  Even though similarities coming from contextualised model seems to present lower correlations with sentence size, still lexical overlap appears to dominate across all types of models.

\begin{table}[!ht]
\centering
\begin{footnotesize}
\begin{tabular}{|l|ccccccccc|}
\hline
 & Word2Vec & GloVe  & ELMo & SBERT  & BERT & BERT  & DistilB  & LLaMA2 & OpenAI  \\
 &   &  &   & ML &   & ML  & ML &  &  \\
 \hline
{P$_{Syn}$} &   &   &   &   &   &   &   &   &        \\
EN-Nat    & 0.71 & 0.71& 0.49& 0.49& 0.58& 0.53& 0.67& 0.46& 0.44    \\
PT-Nat  & 0.66& 0.59& 0.42&0.46 & 0.48&0.57 & 0.65& 0.51& 0.26  \\ \hline
{P$_{Comp}$} &   &   &   &   &   &   &   &   &     \\
EN-Nat    & 0.82 & 0.86& 0.74& 0.80& 0.75& 0.78& 0.89& 0.55& 0.52      \\
PT-Nat & 0.72& 0.80& 0.67& 0.58&0.63 & 0.72&0.83 & 0.59& 0.50\\ \hline
{P$_{WordsSyn}$} &   &   &   &   &   &   &   &   &     \\
EN-Nat  &   0.86  & 0.87      & 0.70 & 0.74     & 0.75 & 0.81    & 0.87       & 0.54&0.60 \\ 
PT-Nat  &   0.74  & 0.70       & 0.58 & 0.62 & 0.69 & 0.67    & 0.78  &0.60 &0.46           \\ \hline
{P$_{Rand}$} &   &   &   &   &   &   &   &   &     \\
EN-Nat & 0.87& 0.88&0.77 &0.85 &0.82 & 0.85& 0.87& 0.62& 0.80\\ 
PT-Nat & 0.77&0.79 & 0.73& 0.74&  0.87& 0.76&0.81 &0.62 & 0.68\\ \hline
\end{tabular}
\end{footnotesize}
\caption{Spearman $ \rho $ correlation between naturalistic sentence length and cosine similarity, p $\leq$ 0.05, for {P$_{Syn}$}, {P$_{Comp}$}, {P$_{WordsSyn}$} and {P$_{Rand}$}. }
\label{tab:corr_sent_length}
\end{table}

To minimise the effect of the lexical overlap in the similarities, we now focus our analyses only on the similarities among the tokens representing the NCs and their substitutions in the context of the target sentences. In this case,  lower similarities were obtained for all probes and all models compared to those at sentence level (Figure \ref{fig:p1-3-NC} vs.  Figure \ref{fig:p1-3-sent}). 
This is even the case for similarities for the NCs and their synonyms ({P$_{Syn}$}), which are centred around the same values as those for the NCs and synonyms of the individual components ({P$_{WordsSyn}$}) and those for the random replacements ({P$_{Rand}$}) follow this trend, and they are all lower than those for the NCs and only one NC component ({P$_{Comp}$}). In fact, similarities for the gold standard synonyms are lower than for many of the other probes, regardless of the extent to which the original NC meaning is changed, as probes {P$_{Comp}$} to {P$_{Rand}$}  involve some change in meaning while {P$_{Syn}$} does not. 
Finally, there is more variation displayed among the models, as there are lower similarities for static than for most contextualised models.
Overall, the resulting similarities at NC level do not follow the expected patterns for representing  idiomaticity, illustrated in Figure \ref{fig:p1-3-NC} (Ideal Values).
  The same holds true for their correlations with human judgements. In line with  what occurs at the sentence level, the similarities at NC level exhibit correlations that contradict linguistic expectations. In particular, it is expected that true synonymous substitutions work well across the idiomatic-compositionality spectrum. Therefore, no correlation should be expected for {P$_{Syn}$}, while for {P$_{Comp}$} and {P$_{WordsSyn}$}, a moderate correlation is expected and no correlation for {P$_{Rand}$}. However, Table  \ref{tab:results-2} indicates that for most models, $\rho_{NC}$({P$_{Syn}$}) $ > \rho_{NC}$ ({P$_{Comp}$}  \; or \; {P$_{WordsSyn}$})  with the latter being either weak or not significant.

 In the next section we analyse if, at least at a detailed level, the similarities between NCs and their synonyms are mostly  higher than of other alternatives.    
 
\subsection{Are the representation of the NCs and their synonyms relatively more similar when compared to other alternatives?}

\begin{figure}[ht!]
\begin{subfigure}{0.52\textwidth}
    \centering
    {A$_{Syn|WordsSyn}$}
   \includegraphics[scale=0.08]{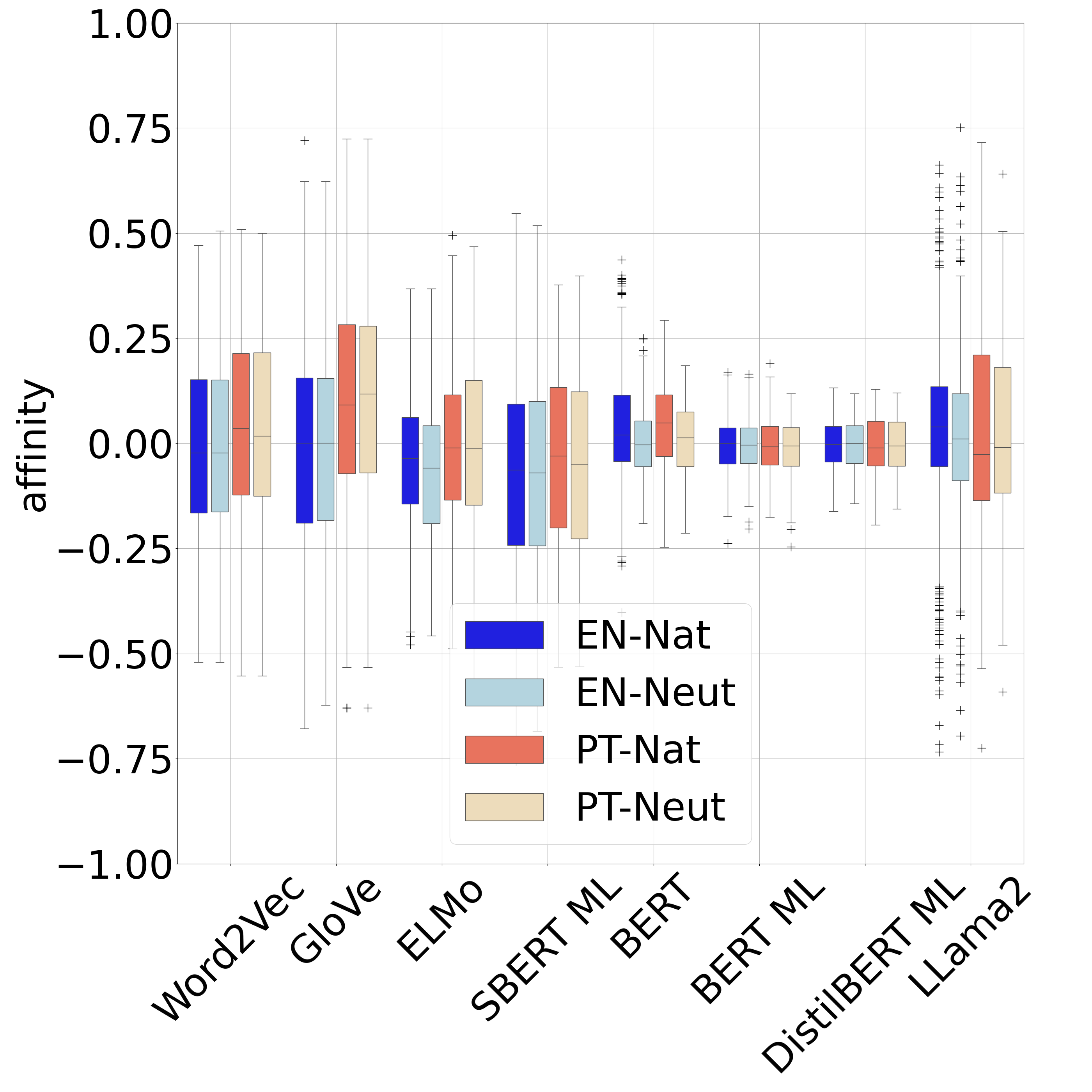}
\end{subfigure}
\begin{subfigure}{0.52\textwidth}
    \centering
{A$_{Syn|Rand}$}
\includegraphics[scale=0.08]{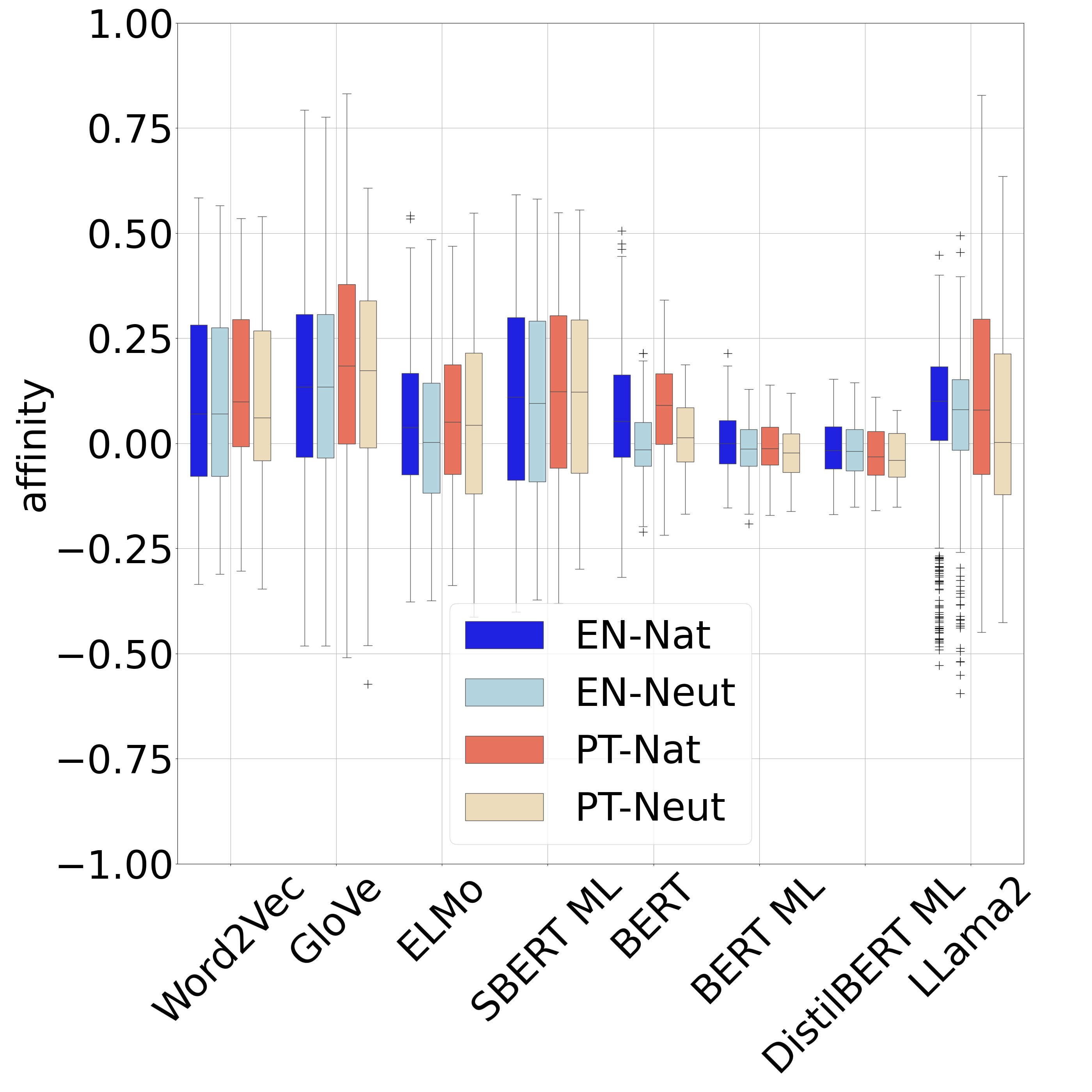}
\end{subfigure}

\begin{subfigure}{0.5\textwidth}
        \centering Ideal Values
\includegraphics[scale=0.155,trim=1mm 1mm 0mm 0mm, clip=true]
{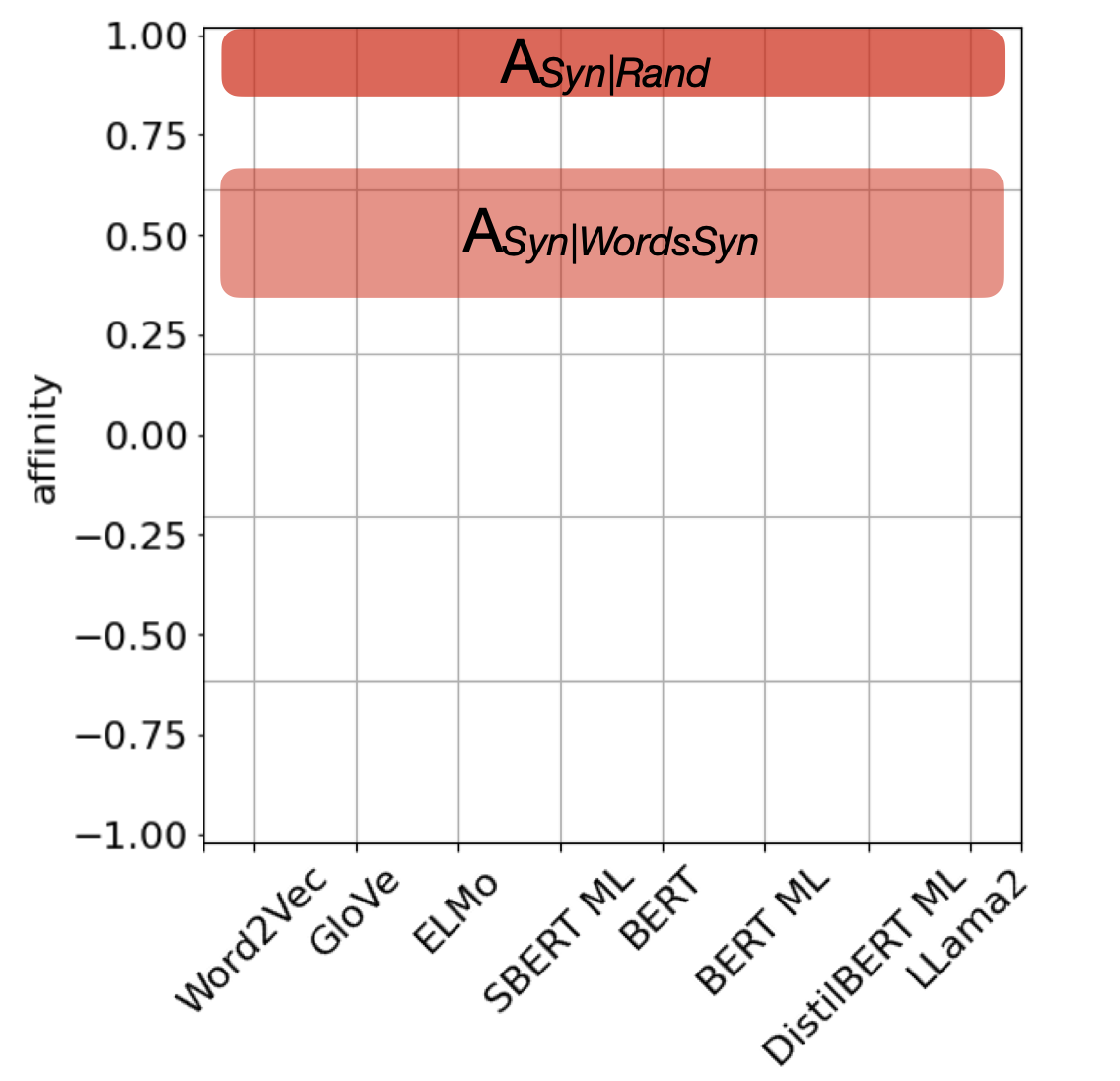}   
\end{subfigure}
\caption{Affinity at the NC level for English (blue) and Portuguese (orange), with naturalistic sentences in darker shade and  neutral in lighter. The lower panel (Ideal Values) is an illustration of values ideally expected for the different affinities. The means and standard deviations are in Table \ref{tab:num_Affinity} in the Appendix.
}
    \protect\label{fig:affinities}
\end{figure}

If a model accurately represents idiomaticity, the representation of a given NC should be more similar to its synonym than to other alternatives, including distractors and random representations. Using the proposed comparative measures of Affinity (introduced in section \ref{sec:affinity}), we now assess whether the models we are evaluating are able to reliably distinguish between a substitution that preserves meaning ({P$_{Syn}$}) from those that do not ({P$_{WordsSyn}$} for more idiomatic NCs, {P$_{Rand}$} for all NCs).  The results from the previous section demonstrated that, on average, the models do not seem to represent idiomaticity correctly. For instance, Figure \ref{fig:p1-3-NC} shows that probe {P$_{Comp}$} yields larger average similarities than probe {P$_{Syn}$}, and that {P$_{Syn}$} and {P$_{WordsSyn}$} have similar averages, but with {P$_{Syn}$} exhibiting more variance. Both results are incompatible with a good idiomatic representation. Affinity will allow us to verify this on a per-NC basis.
In particular, we compare the Affinities for NCs and their synonyms against the synonyms of their individual components ({A$_{Syn|WordsSyn}$}), and against random substitutions   ({A$_{Syn|Rand}$ }),
with the expected affinity ranges shown in Figure \ref{fig:affinities}(Ideal Values). 

First of all, comparing against synonyms of the individual components (Figure \ref{fig:affinities}(({A$_{Syn|WordsSyn}$})) on the whole the models display comparable abilities in term of averages,  around 0 for all models, but differ to some extent in their variances. As the Affinities obtained are mostly neutral (around 0) the models do not display the higher similarities between the NCs and their gold synonyms to the extent that would be expected. Moreover, this holds even for random replacements (Figure \ref{fig:affinities}({A$_{Syn|Rand}$})) where some models display small positive averages, but are far from the expected ideal (Figure \ref{fig:affinities}(Ideal Values)).

\begin{figure}[ht!]
\begin{subfigure}{0.5\textwidth}
    \centering  A$_{Syn|WordsSyn}$ \hspace{0.3cm} EN
\includegraphics[scale=0.08]{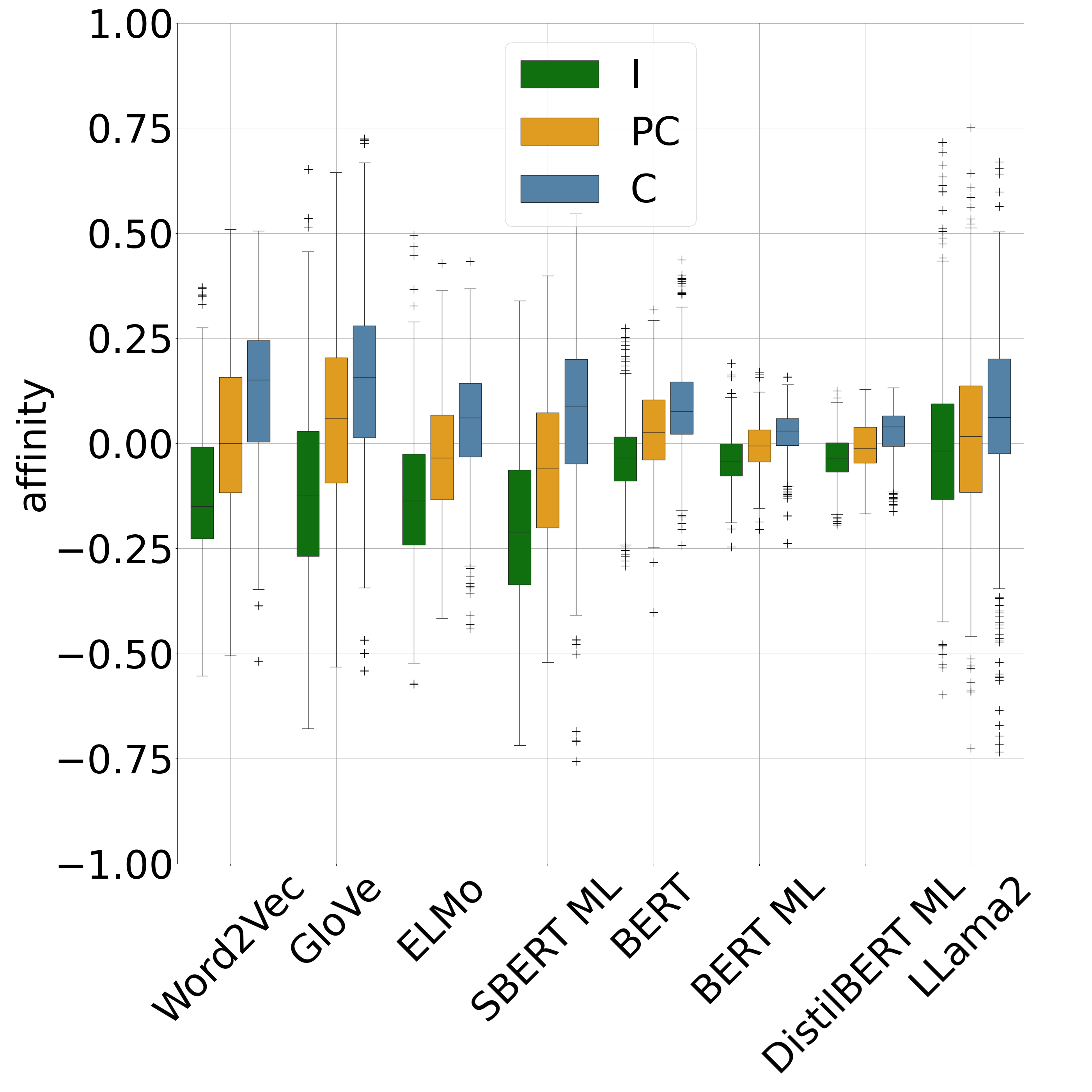}
\end{subfigure}
\begin{subfigure}{0.5\textwidth}
        \centering
          \hspace{7pt} A$_{Syn|Rand}$ \hspace{0.3cm} EN
\includegraphics[scale=0.08]{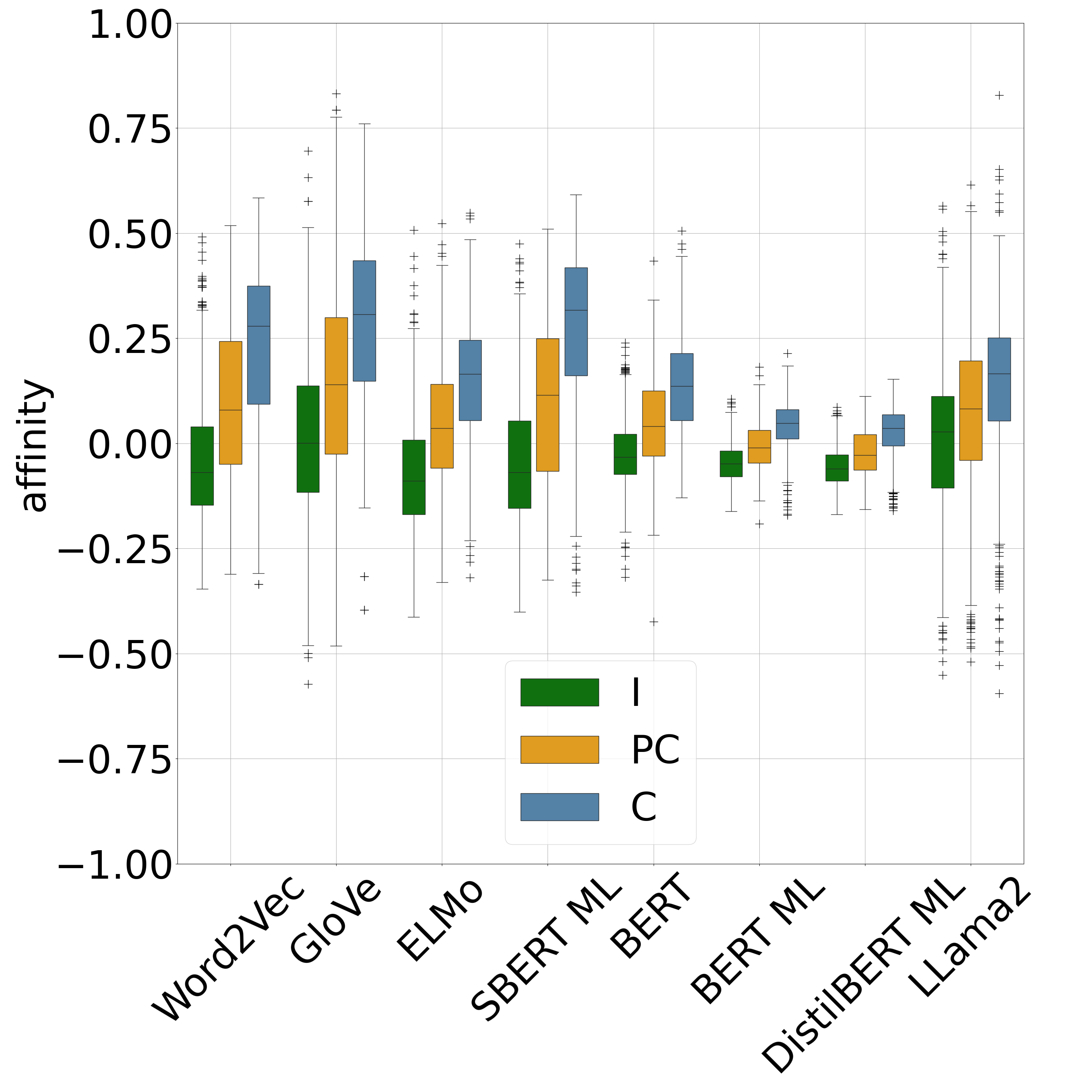}
\end{subfigure}
\begin{subfigure}{0.5\textwidth}
    \centering
A$_{Syn|WordsSyn}$ \hspace{0.1cm} PT    \includegraphics[scale=0.08]{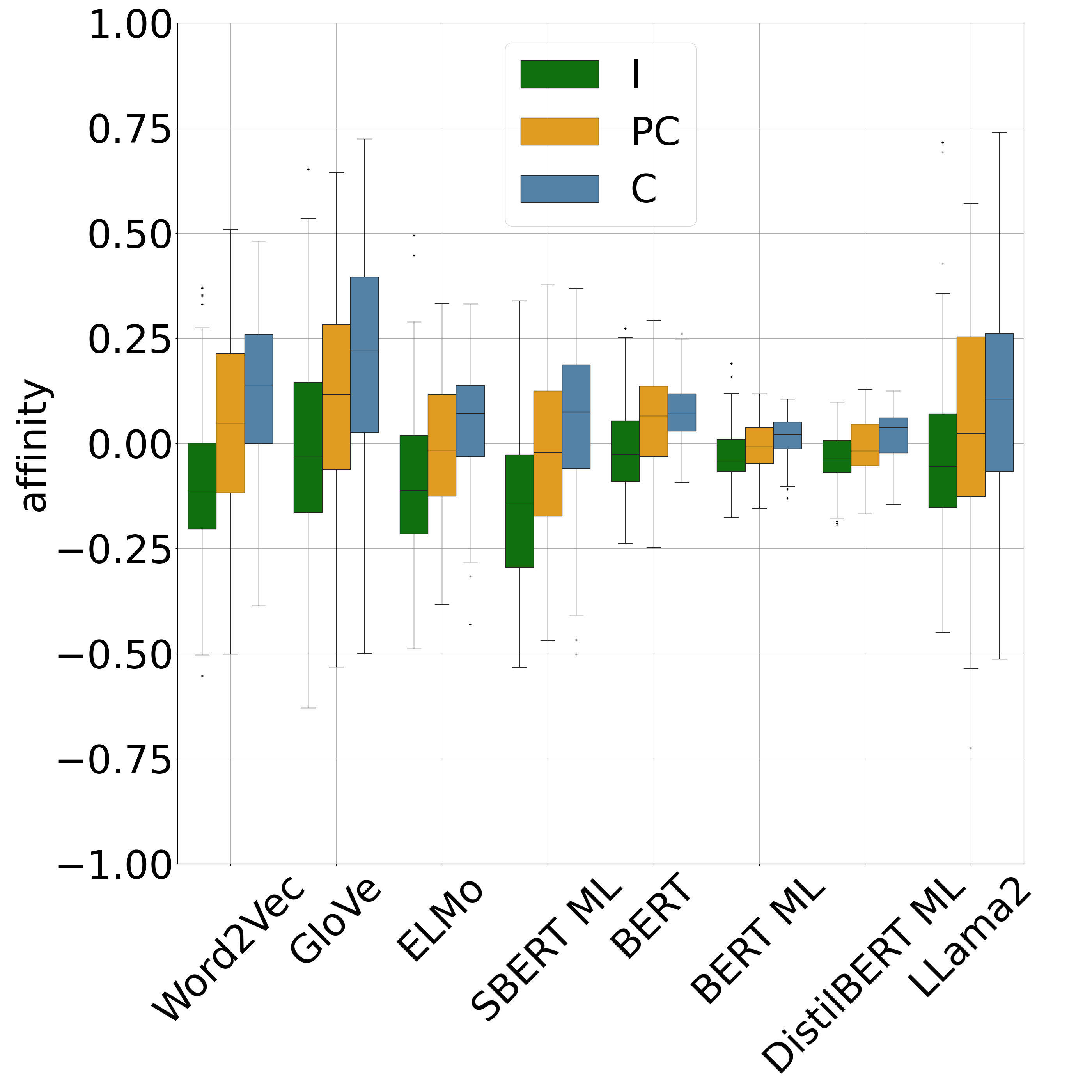}
\end{subfigure}
\begin{subfigure}{0.5\textwidth}
    \centering
          \hspace{7pt} A$_{Syn|Rand}$ \hspace{0.1cm} PT
\includegraphics[scale=0.08]{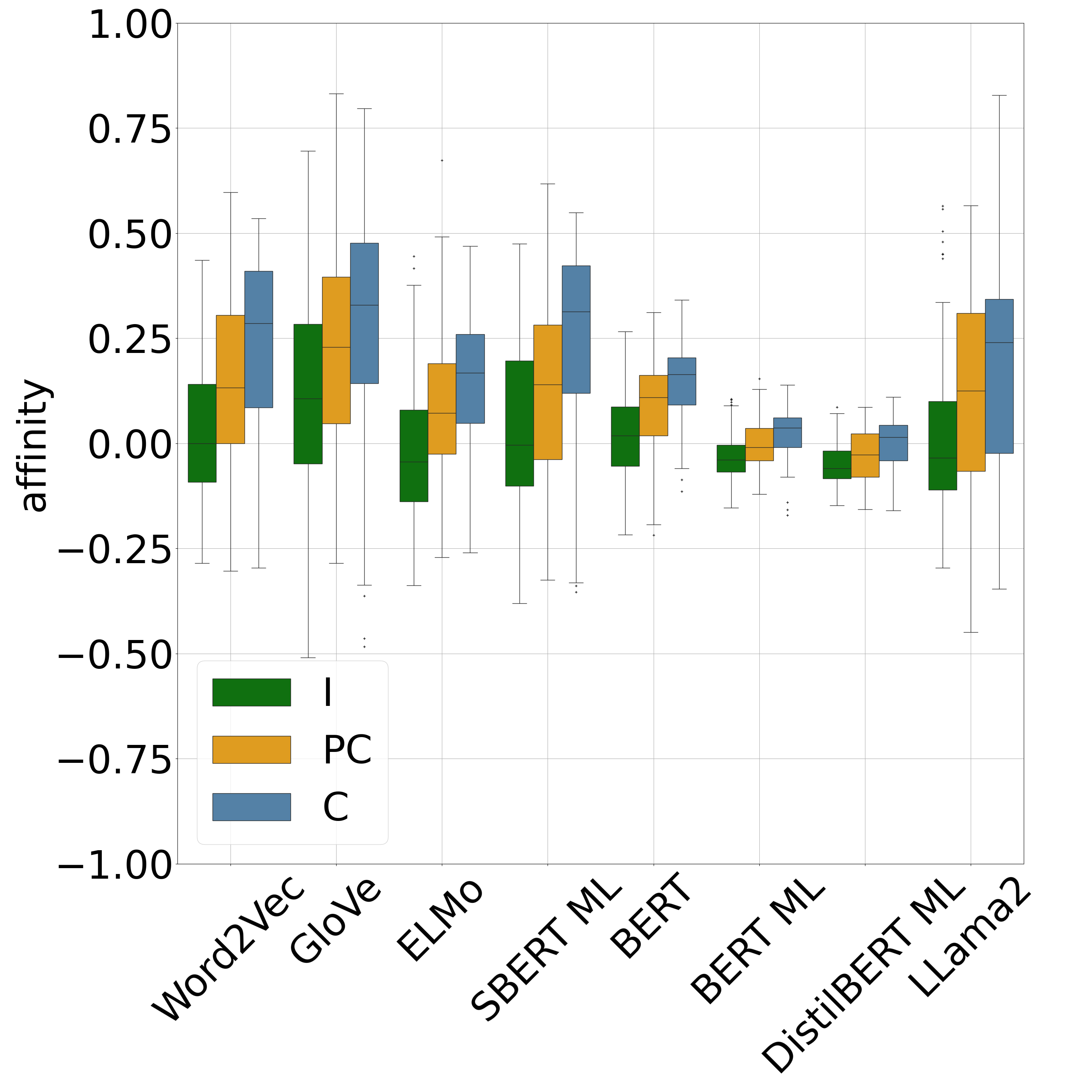}
\end{subfigure}

\caption{Affinity by idiomaticity Class at NC level for English (EN) and Portuguese (PT) naturalistic sentences. 
Idiomatic (I) in green,  partly compositional (PC) in yellow and Compositional NCs (C) in blue.}
    \protect\label{fig:affinities2}
\end{figure}

The relatively important variances in Figure \ref{fig:affinities} call for an analysis of the Affinities according to idiomaticity level.  This is displayed in Figure \ref{fig:affinities2} for English naturalistic sentences where the classification of NCs as compositional (C), partly compositional (PC) and idiomatic (I) from \citet{garcia-etal-2021-assessing} is adopted.
A striking pattern emerges in all the figures. For each model, its distribution of Affinities splits into three distinct distributions with similar variances but different averages ordered according to compositionality. Compositional NCs exhibit higher Affinities than partly compositional NCs, and the latter show higher Affinities than idiomatic NCs. This is confirmed by the correlation analysis in Table \ref{tab:results_Affinity_cor},  with most models displaying significant weak to moderate correlations between Affinities and human compositionality judgements, for all Affinities types, including neutral sentences and  Portuguese data.\footnote{We omitted the equivalent of Figure \ref{fig:affinities2} for neutral sentences and Portuguese data  due their visual similarity to the English naturalistic version.} This  contradicts what was generally expected: Affinity {A$_{Syn|WordsSyn}$} values should exhibit a negative correlation with compositionality, while Affinity {A$_{Syn|Rand}$} 
should show no correlation at all. 

These results suggest that representations of idiomatic NCs may not be accurately incorporating their meanings, since NCs are not closer to their synonyms than to other alternatives, even if they are random. Moreover, the more idiomatic NCs seem to be more similar to synonyms of their individual components, which suggests that the surface clues about their individual components may be playing a greater role in driving these similarities, even in contextualised models. This result remains valid even after removing compounds from the dataset that have lexical overlaps with the NC$_{Syn}$ produced by the annotators (see Table \ref{tab:results_Affinity_cor_remove} in the appendix).

\begin{table*}[!ht]
\centering
\begin{footnotesize}
\begin{tabular}{|l|cccccccc|}
\hline
 	&	Word2Vec &	GloVe	&	ELMo	&	SBERT  	&	BERT	&	BERT 	&	DistilBERT 	&	LLaMA2	\\ 
  &   &  &   & ML &   & ML  & ML &    \\ \hline
 {A$_{Syn|WordsSyn}$}	&		&		&	&	  	&		&	 	&	 	&		 \\  
EN-Nat	&	0.58	&	0.52	&	0.55	&	0.58	&	0.57	&	0.55	&	0.51	&	0.17	\\ 
EN-Neut 	&	0.58	&	0.52	&	0.54	&	0.58	&	0.53	&	0.49	&	0.50	&	0.23	\\ 
PT-Nat	&	0.43	&	0.37	&	0.39	&	0.38	&	0.35	&	0.34	&	0.35	&	0.27	\\ 
PT-Neut &	0.44	&	0.37	&	0.41	&	0.40	&	0.34	&	0.32	&	0.36	&	0.31	\\ \hline 

{A$_{Syn|Rand}$}	&		&		&	&	  	&		&	 	&	 	&		 \\ 
EN-Nat	&	0.60	&	0.54	&	0.63	&	0.66	&	0.69	&	0.68	&	0.61	&	0.39	\\ 
EN-Neut 	&	0.59	&	0.53	&	0.63	&	0.64	&	0.59	&	0.55	&	0.57	&	0.36	\\ 
PT-Nat	&	0.48	&	0.41	&	0.54	&	0.49	&	0.51	&	0.48	&	0.42	&	0.36	\\ 
PT-Neut	&	0.44	&	0.41	&	0.50	&	0.47	&	0.46	&	0.46	&	0.41	&	0.33	\\ \hline 
														
\end{tabular}
\end{footnotesize}
\caption{Spearman $\rho$ correlation between the \emph{Affinity} and human judgments for English and Portuguese for naturalistic (Nat) and neutral (Neut) sentences. Non-significant (p $>$ 0.05) results omitted from the table. Although these values shown are for the correlations at the NC level, the correlations at Sentence level are comparable. }
\label{tab:results_Affinity_cor}
\end{table*}

\subsection{Can a more meaningful similarity measure be found to unveil NC meaning? }
\begin{figure}[!ht]
\begin{subfigure}
{0.5\textwidth}
        \centering 
\includegraphics[scale=0.2,trim=30mm 0mm 40mm 0mm, clip=true]{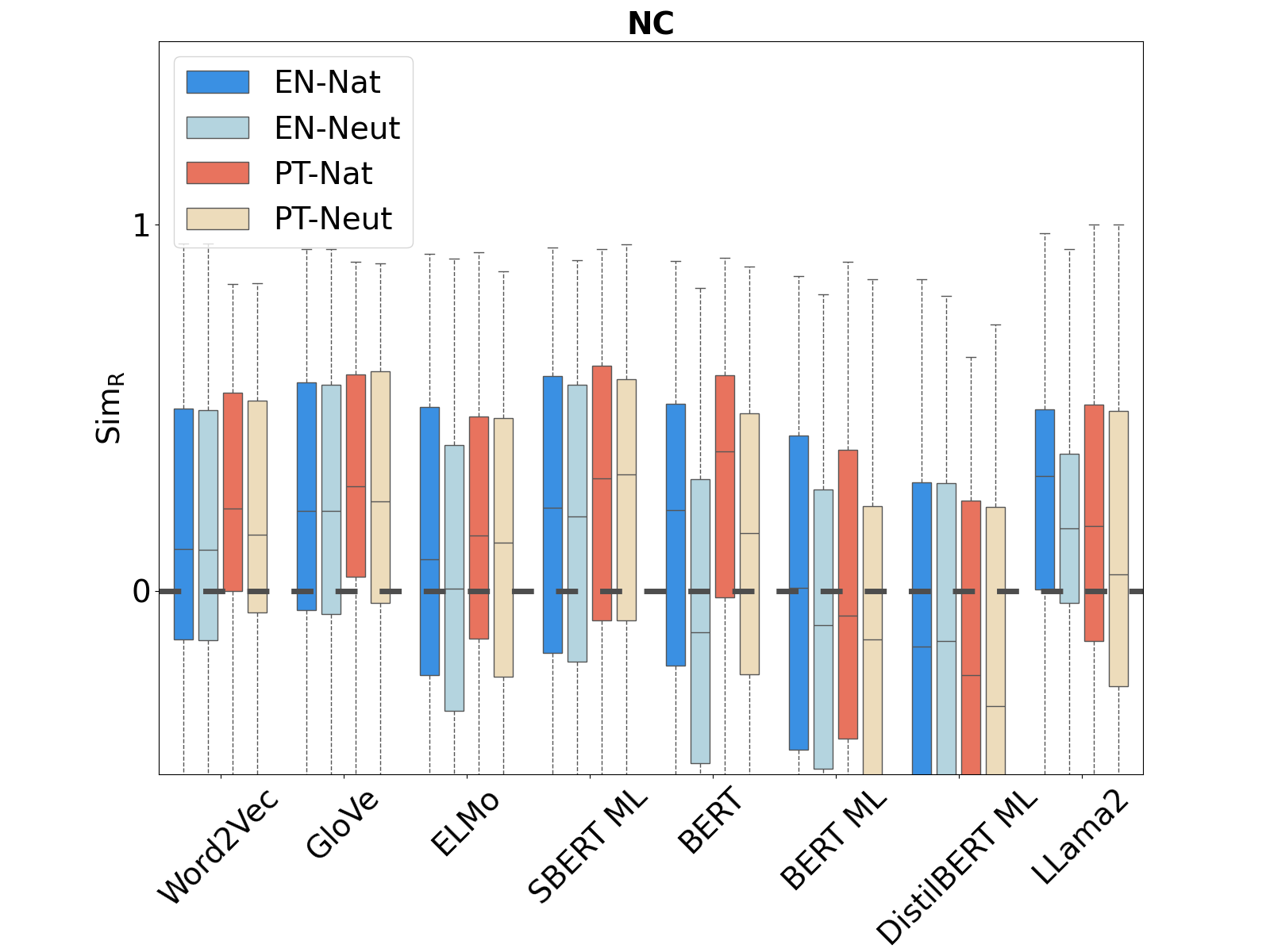} 
\hspace{8pt} 
Sim$_{R|Syn}$
\end{subfigure}
\begin{subfigure}    
{0.5\textwidth}
    \centering 
\includegraphics[scale=0.2,trim=30mm 0mm 40mm 0mm, clip=true]{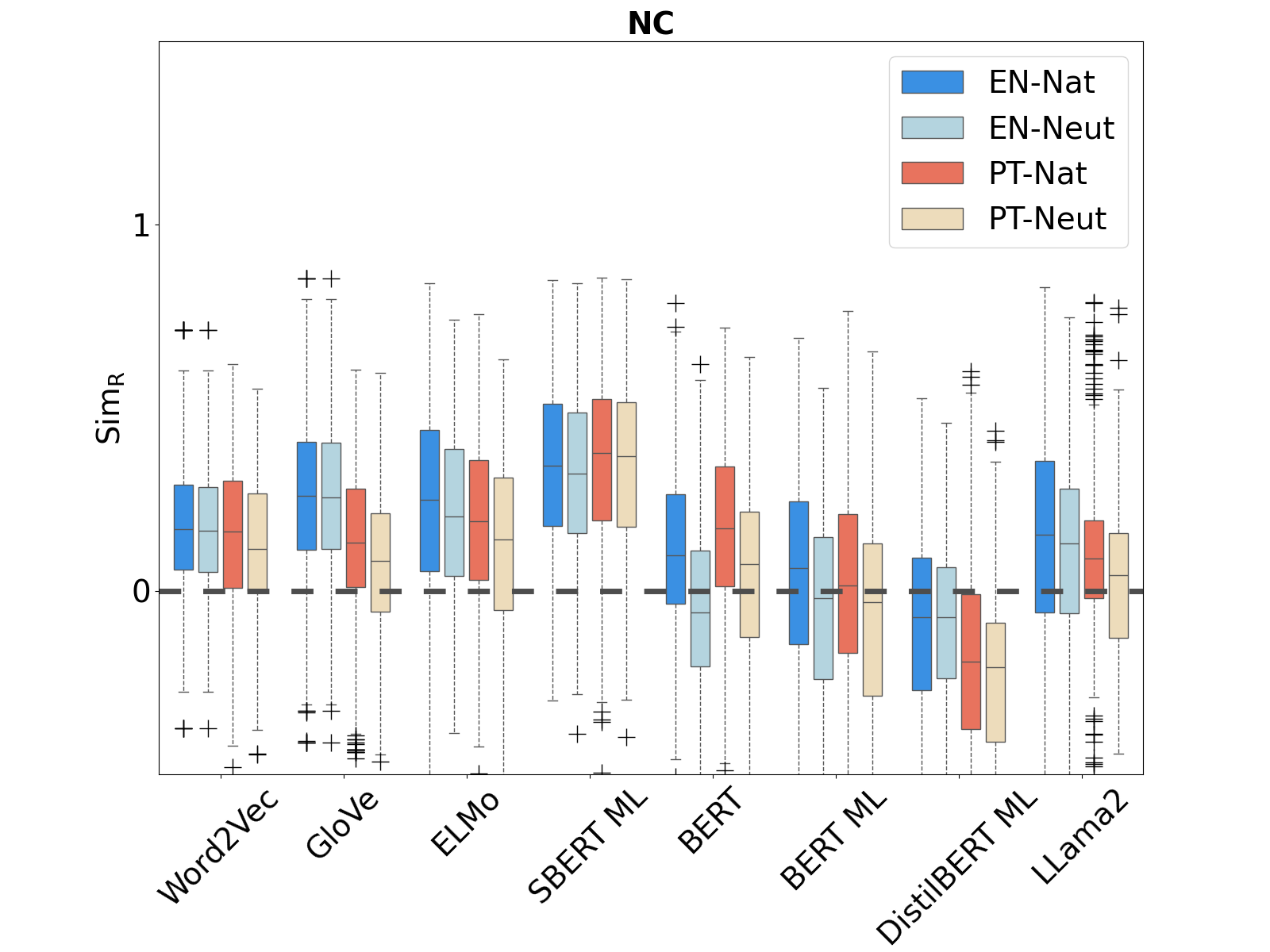}
\hspace{8pt}        {Sim$_{R|WordsSyn}$}
\end{subfigure}
\caption{Average 
Scaled Similarity when the original NCs are replaced by gold synonyms ({Sim$_{R|Syn}$}) or by the synonyms of component words (Sim$_{R|WordsSyn}$), in relation to random substitutions.  
English (blue) and Portuguese (orange), with naturalistic sentences in darker shades and for neutral in lighter. The means and standard deviations are in Table \ref{tab:num_S14_S34} in the Appendix.}
\protect \label{fig:loss-of-meaning}
\end{figure}

\begin{figure}[!ht]
\begin{subfigure}{0.5\textwidth}
    \centering
          \hspace{4pt} 
   \includegraphics[scale=0.19,trim=20mm 0mm 40mm 0mm, clip=true]{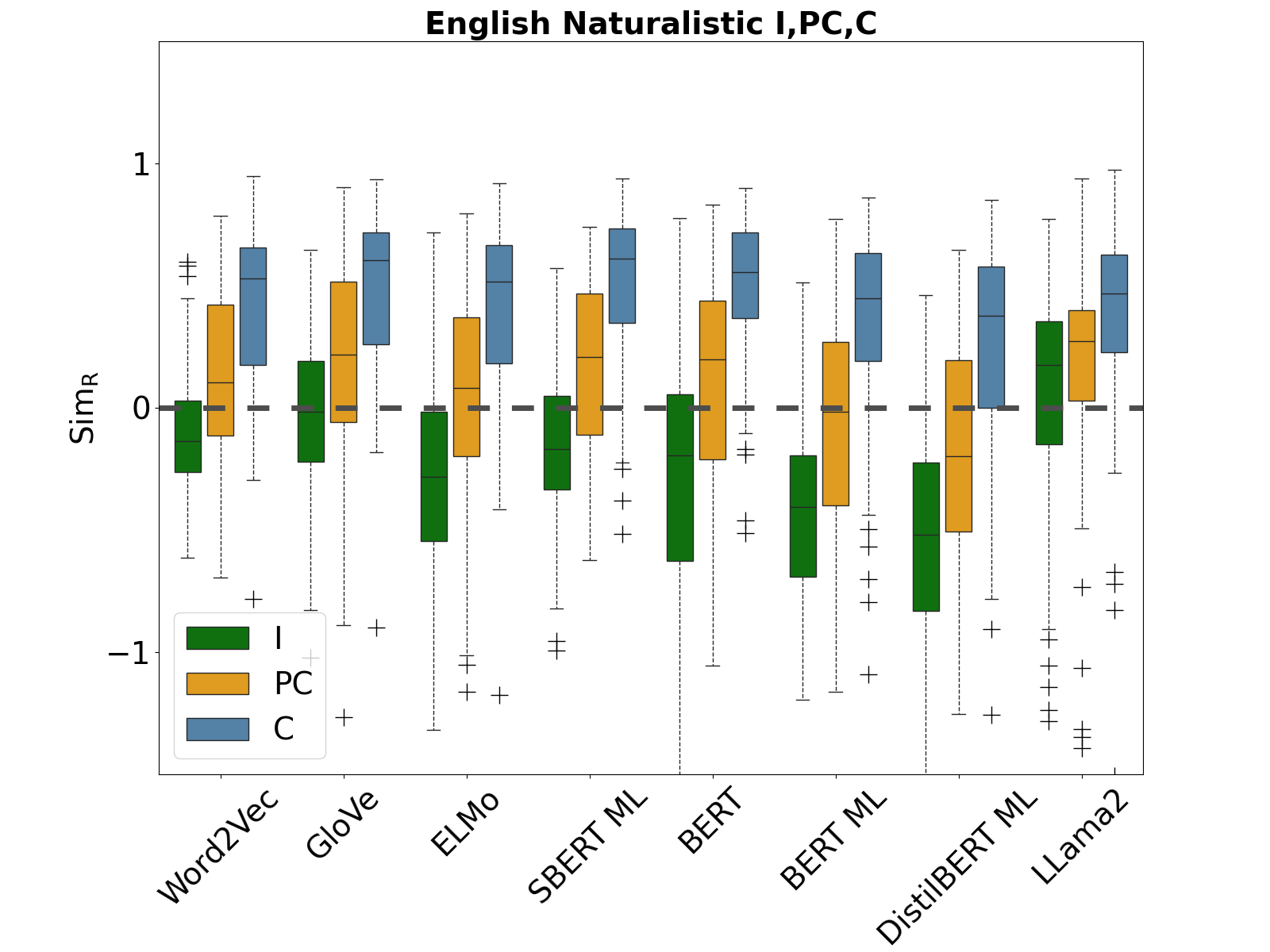}
          \hspace{8pt} 
 \end{subfigure}
 \begin{subfigure}{0.5\textwidth}
    \centering
          \hspace{4pt} 

   \includegraphics[scale=0.19,trim=20mm 0mm 40mm 0mm, clip=true]{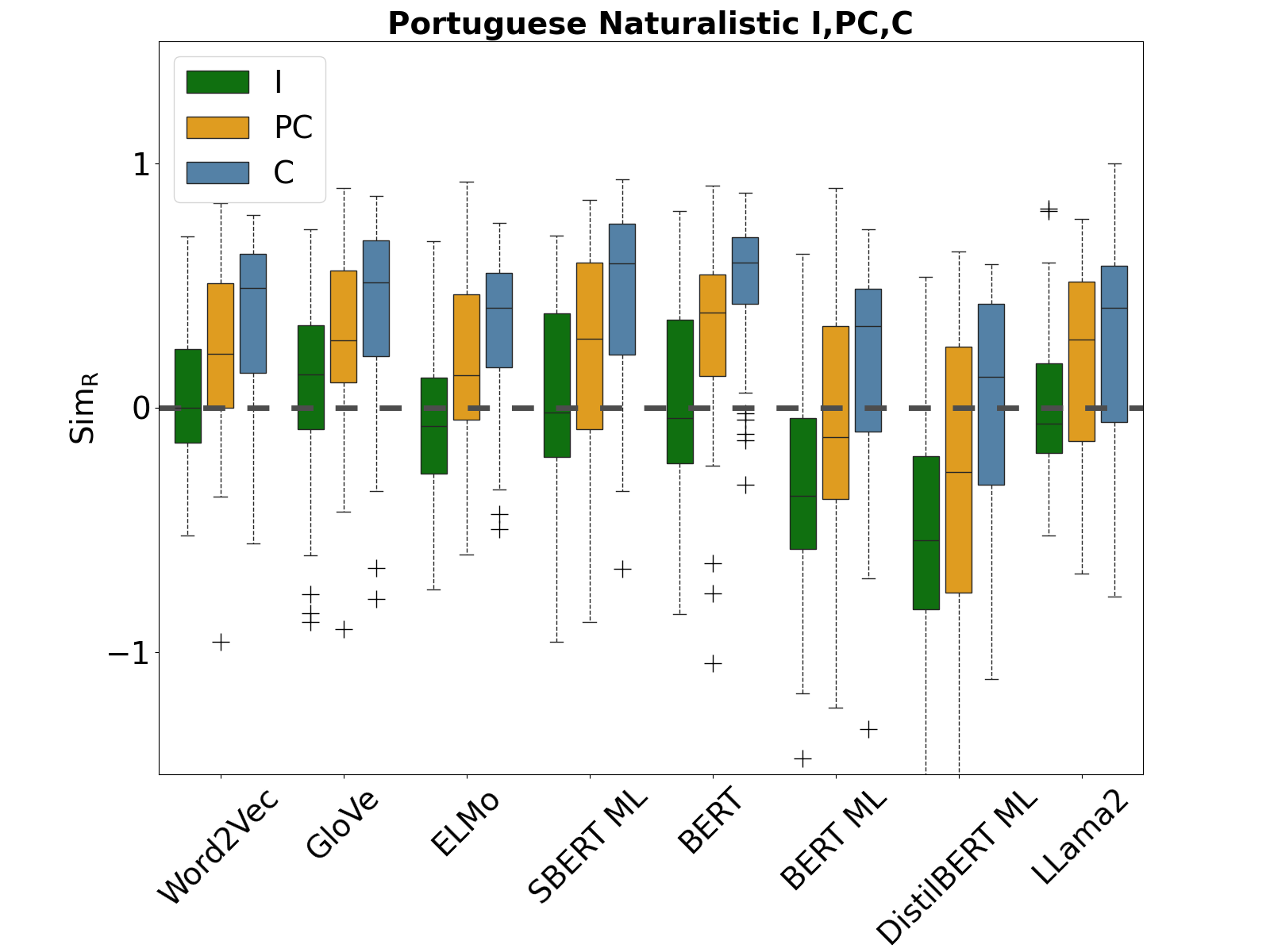}
          \hspace{8pt} 
 \end{subfigure}

  \begin{subfigure}{0.5\textwidth}
    \centering
          \hspace{4pt} 
   \includegraphics[scale=0.19,trim=20mm 0mm 40mm 0mm, clip=true]{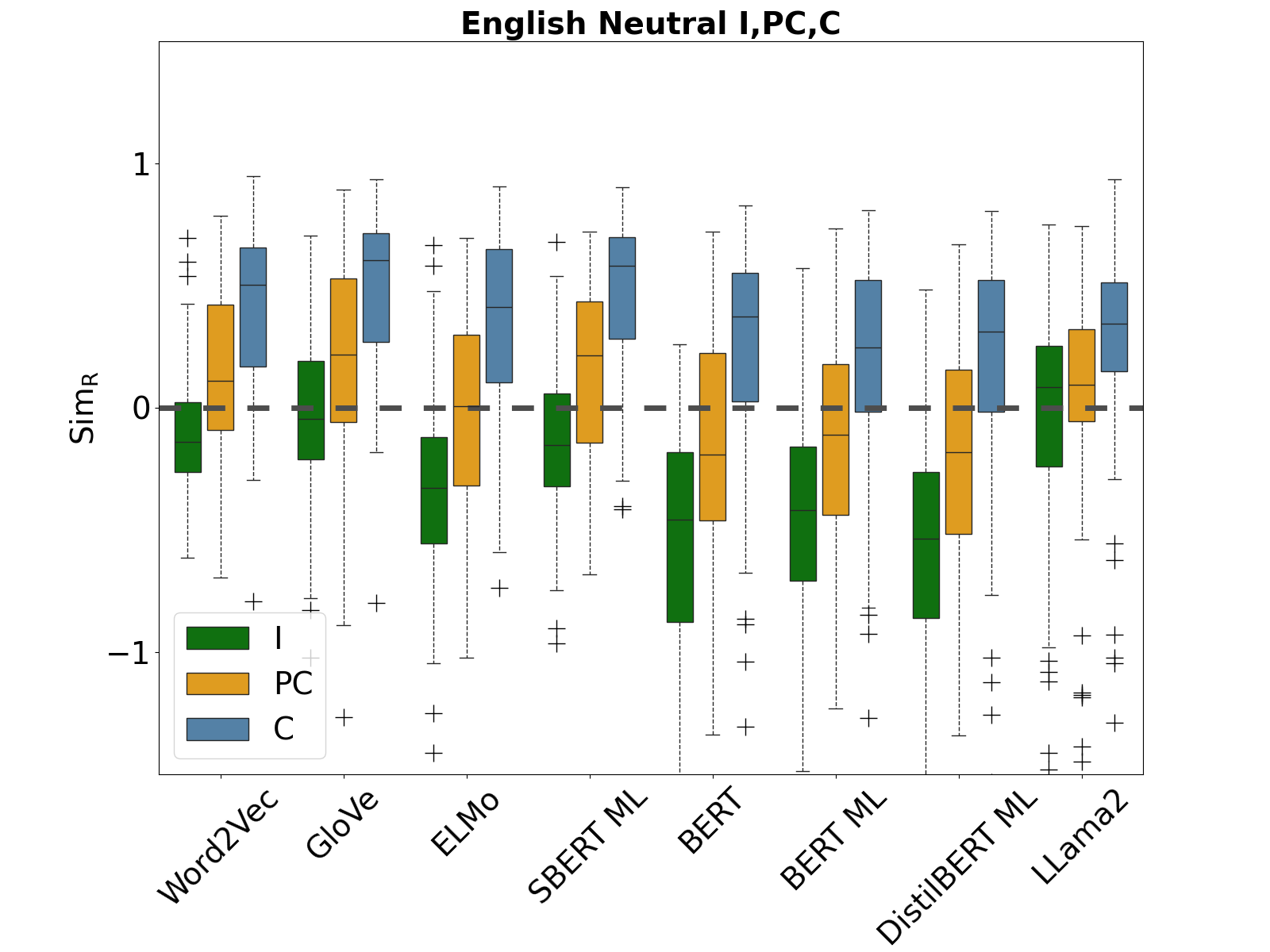}
          \hspace{8pt} 
\end{subfigure}
  \begin{subfigure}{0.5\textwidth}
    \centering
          \hspace{4pt} 

   \includegraphics[scale=0.19,trim=20mm 0mm 40mm 0mm, clip=true]{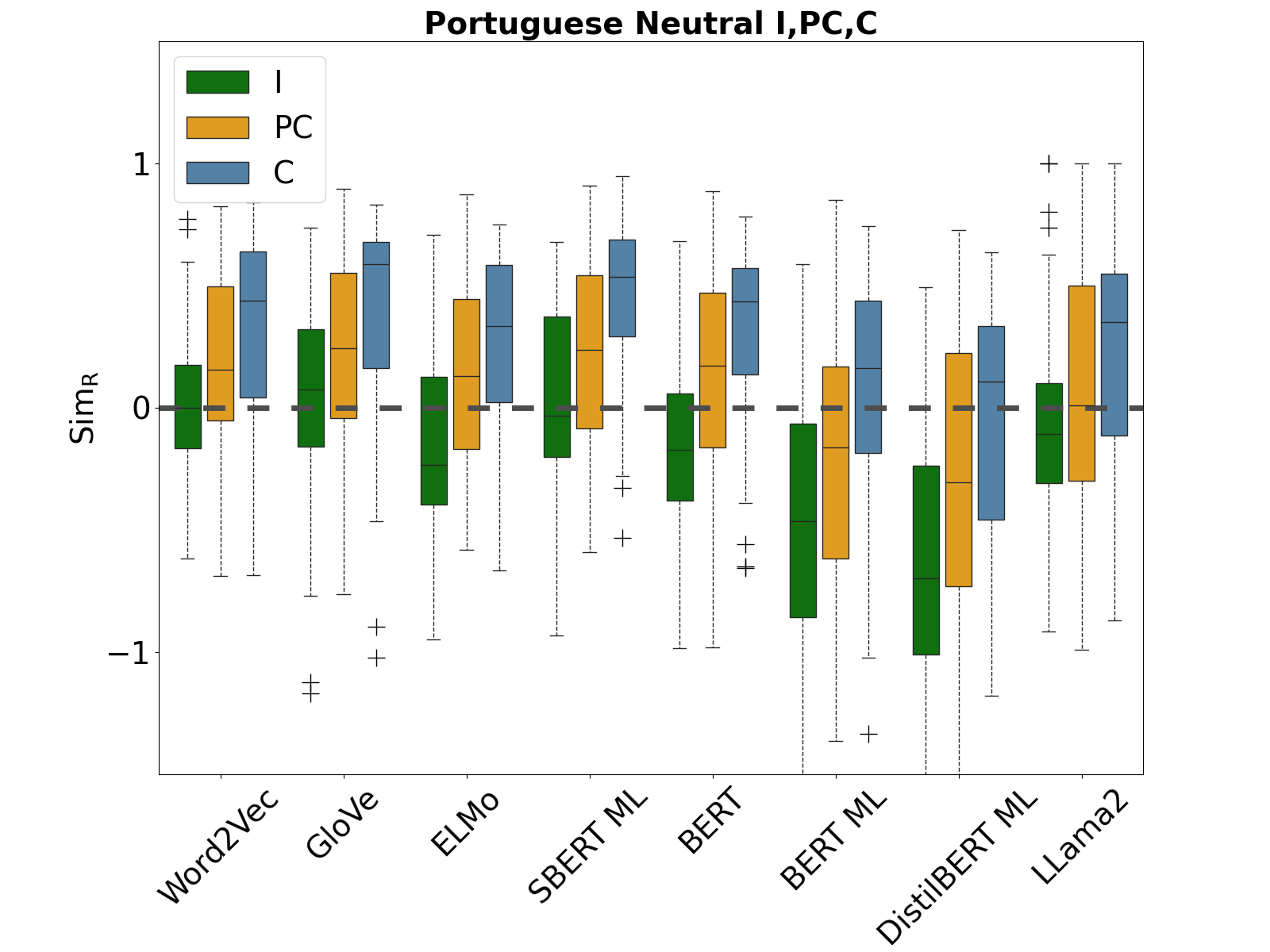}
          \hspace{8pt} 
\end{subfigure}
\caption{$\mbox{Sim}_{RP1}$ per Compositionality Class: green for Idiomatic (I), yellow for Partly Compositional (PC) and blue for Compositional (C), in English (EN) and Portuguese (PT), in Naturalistic  and Neutral sentences. } 
\protect \label{fig:loss_of_meaning-p1}
\end{figure}

\begin{figure}[!ht]
\begin{subfigure}{0.5\textwidth}
    \centering
          \hspace{4pt} 

   \includegraphics[scale=0.19,trim=20mm 0mm 40mm 0mm, clip=true]{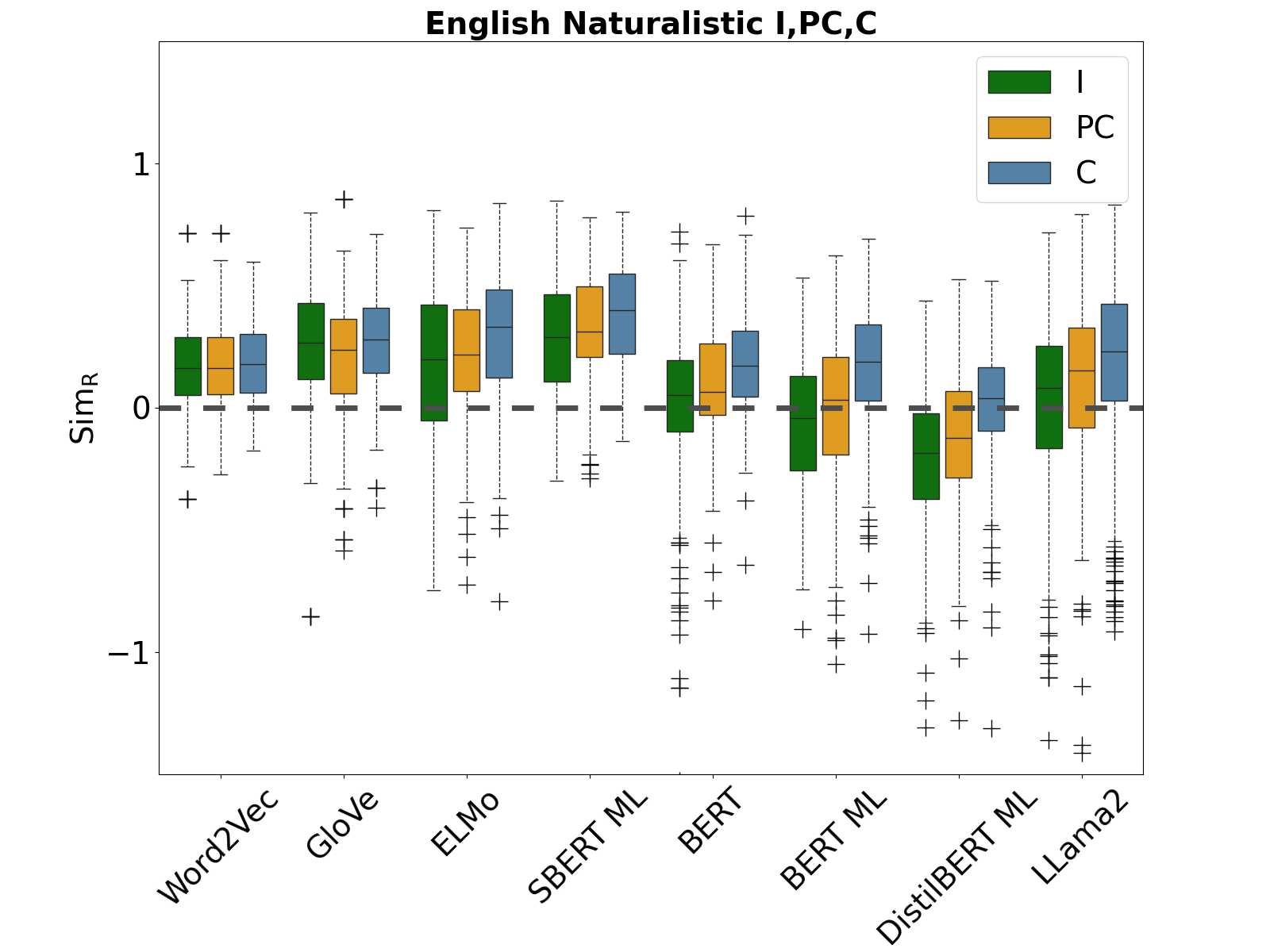}
          \hspace{12pt} 
 \end{subfigure}
 \begin{subfigure}{0.5\textwidth}
    \centering
          \hspace{4pt} 

   \includegraphics[scale=0.19,trim=20mm 0mm 40mm 0mm, clip=true]{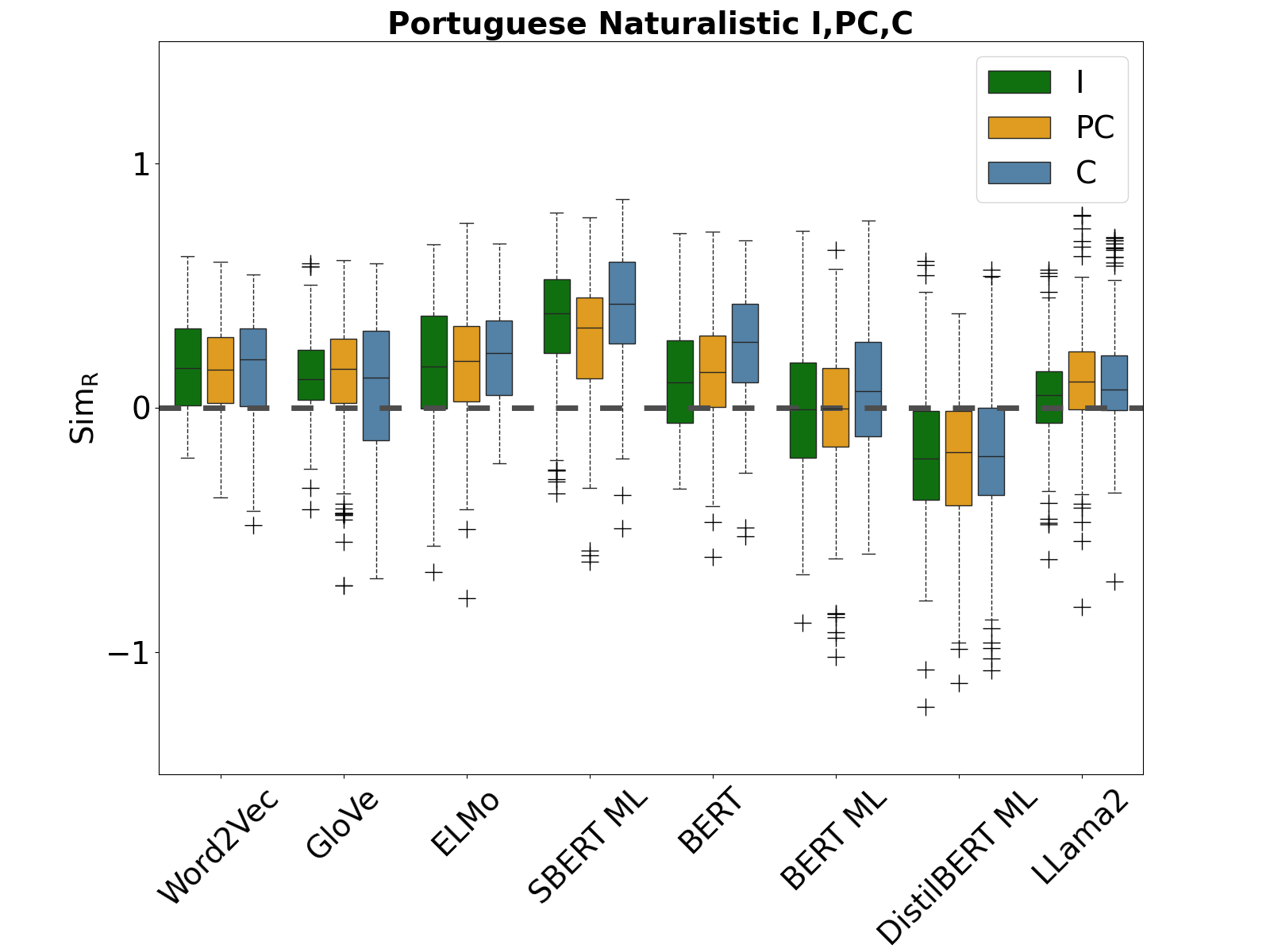}
          \hspace{12pt} 
 \end{subfigure}
  \begin{subfigure}{0.5\textwidth}
    \centering
          \hspace{4pt} 
   \includegraphics[scale=0.19,trim=20mm 0mm 40mm 0mm, clip=true]{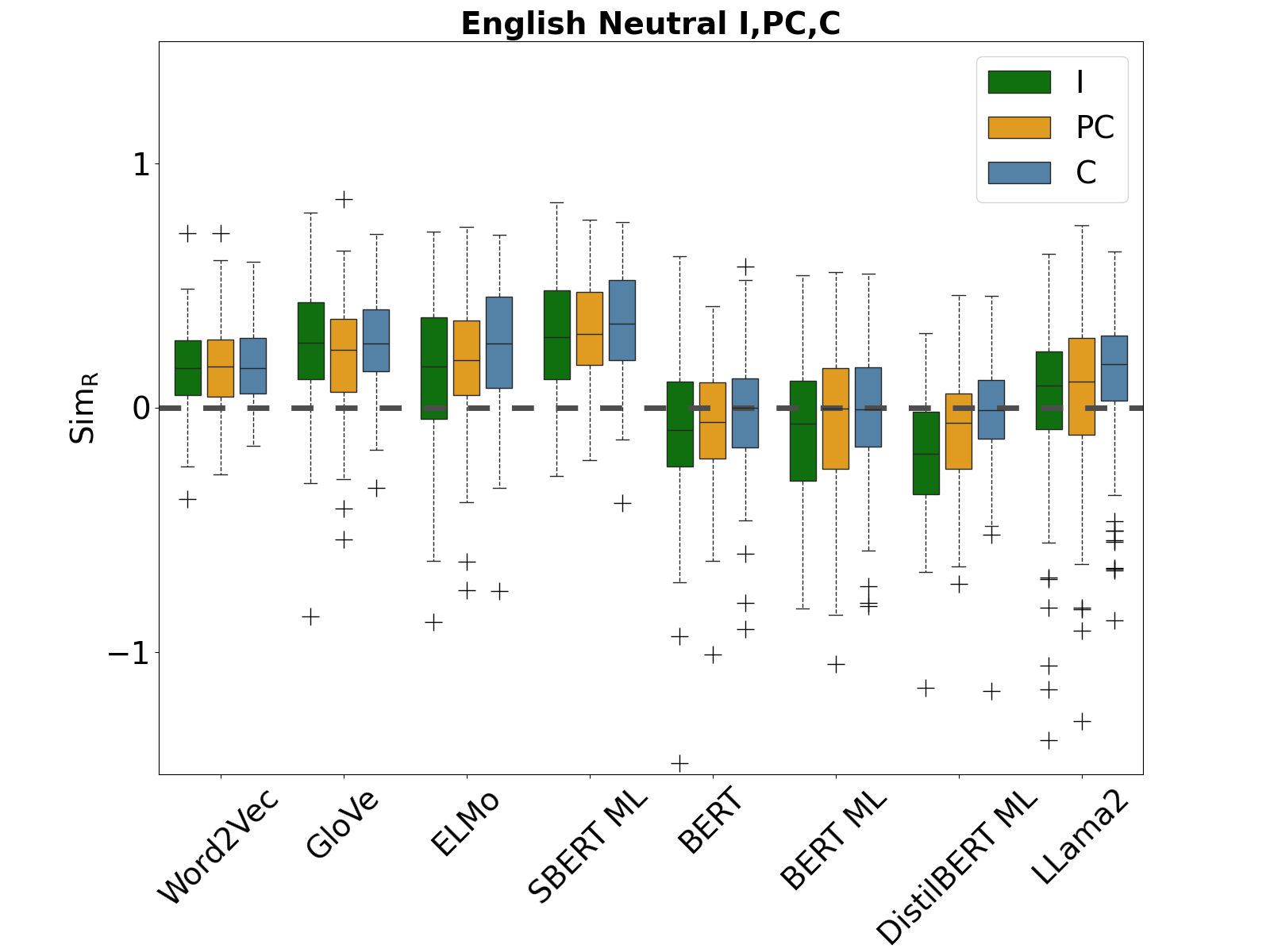}
\end{subfigure}
  \begin{subfigure}{0.5\textwidth}
    \centering
          \hspace{4pt} 

   \includegraphics[scale=0.19,trim=20mm 0mm 40mm 0mm, clip=true]{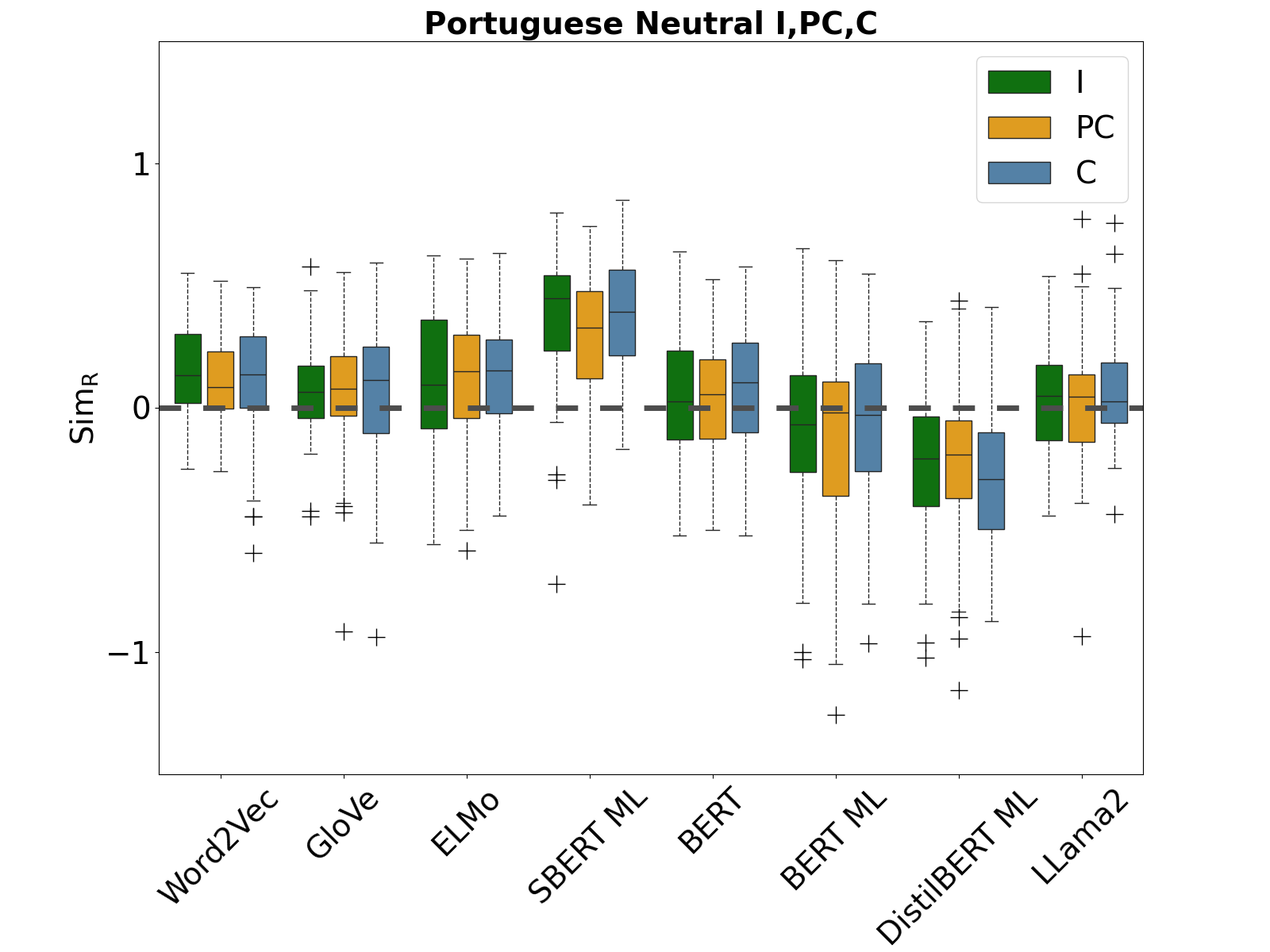}
\end{subfigure}
\caption{$\mbox{Sim}_{RP3}$ per Compositionality Class: green for Idiomatic (I), yellow for Partly Compositional (PC) and blue for Compositional (C), in English (EN) and Portuguese (PT), in Naturalistic and Neutral  sentences.} 
\protect \label{fig:p3}
\end{figure}


\begin{figure}[!ht]
\begin{center}
   \includegraphics[scale=0.2,trim=10mm 0mm 40mm 0mm, clip=true]{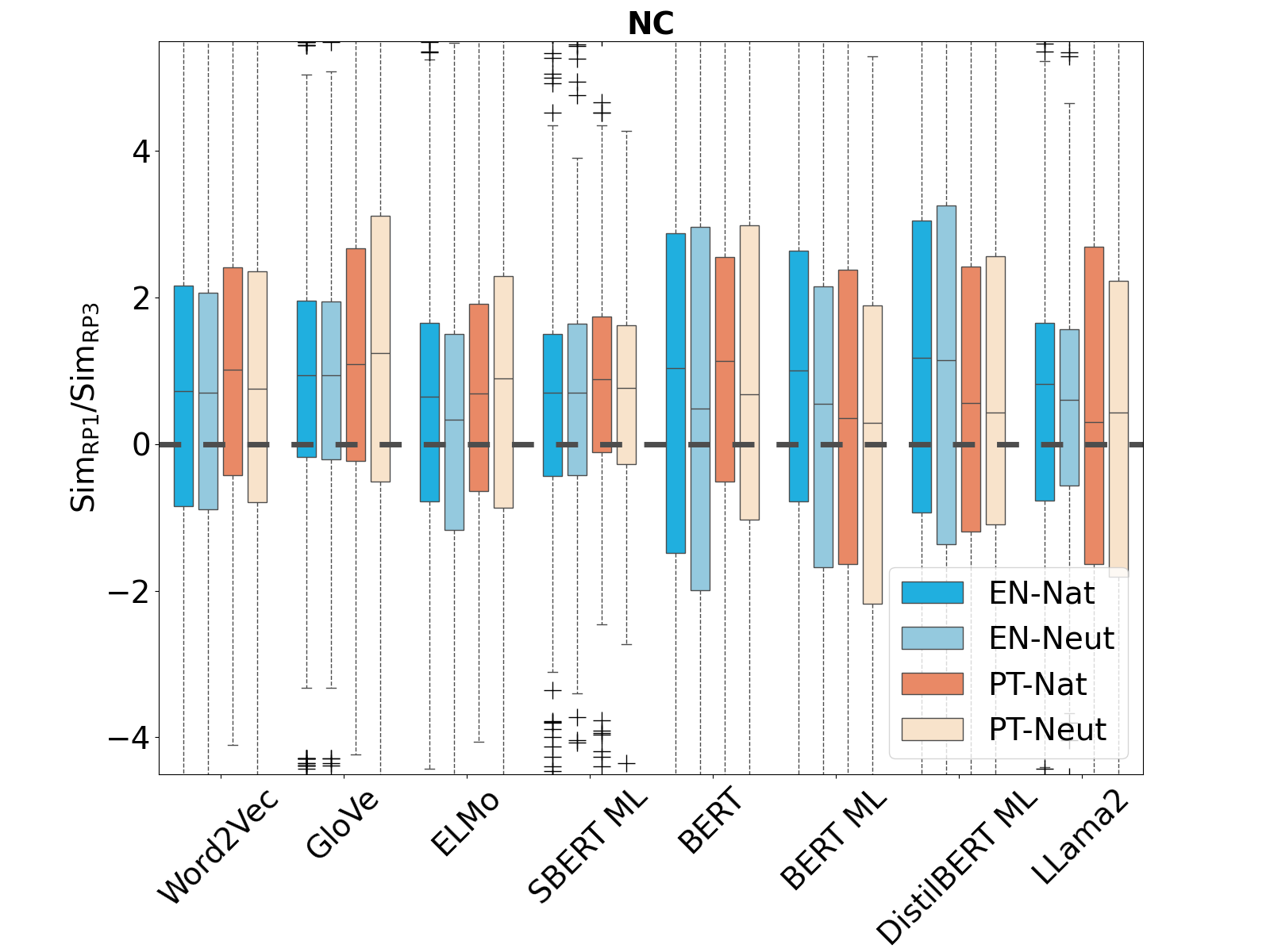} 
\caption{Ratio of Average of Scaled Similarity ({$\mbox{Sim}_{R|Syn}/\mbox{Sim}_{R|WordsSyn}$}). English data are in blue, Portuguese data in orange, values for naturalistic sentences in darker shade and for neutral in lighter. }
\protect \label{fig:p1p3}
\end{center}
\end{figure}

\begin{figure}[!ht]
\begin{subfigure}{0.5\textwidth}
    \centering
          \hspace{1pt} 
        ${\mbox{ EN-Nat} }$
\includegraphics[scale=0.19,trim=20mm 0mm 40mm 0mm, clip=true]{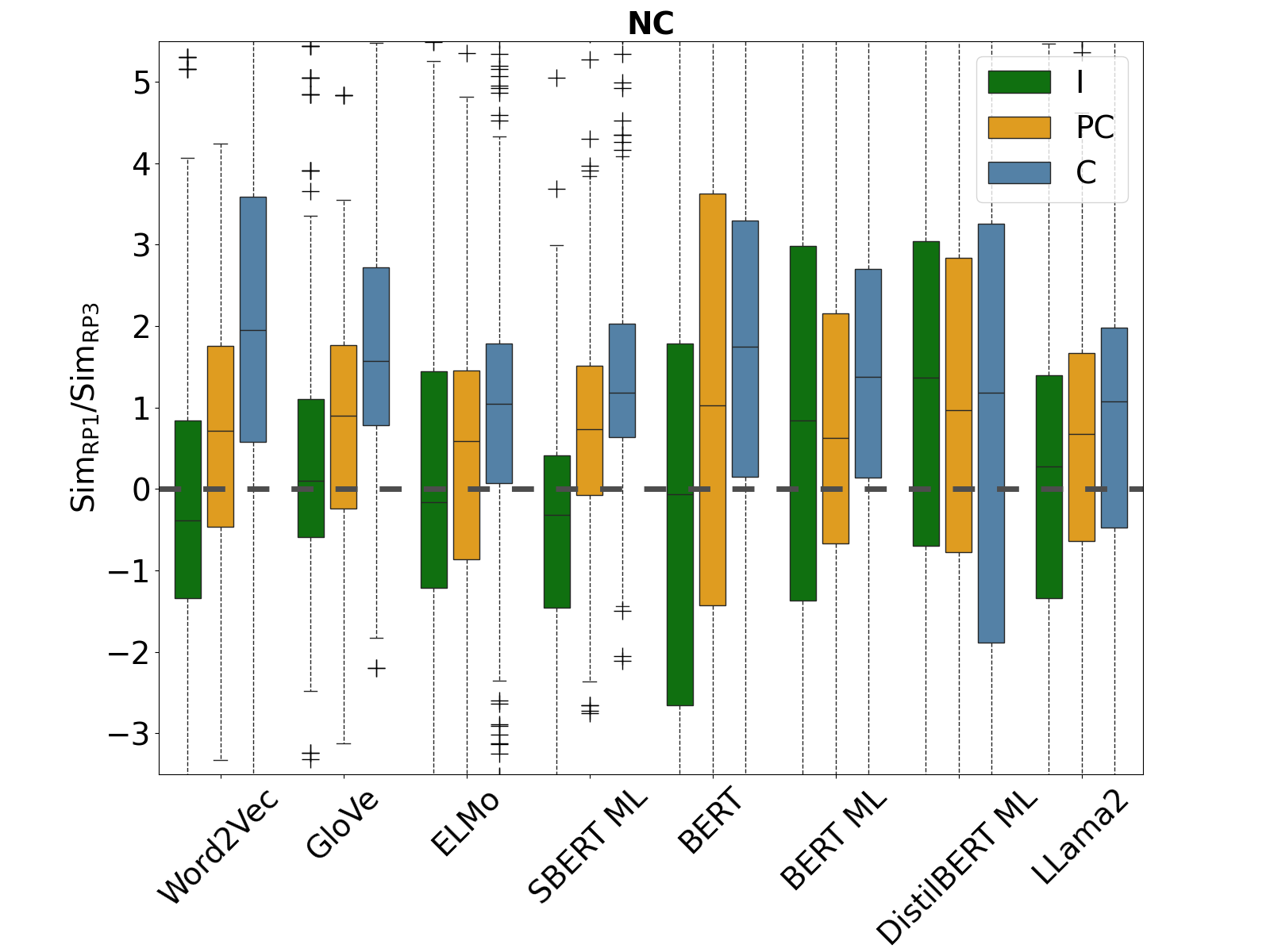} 
\end{subfigure}
\begin{subfigure}{0.5\textwidth}
    \centering
          \hspace{4pt} 
        ${\mbox{PT-Nat}}$
\includegraphics[scale=0.19,trim=20mm 0mm 40mm 0mm, clip=true]{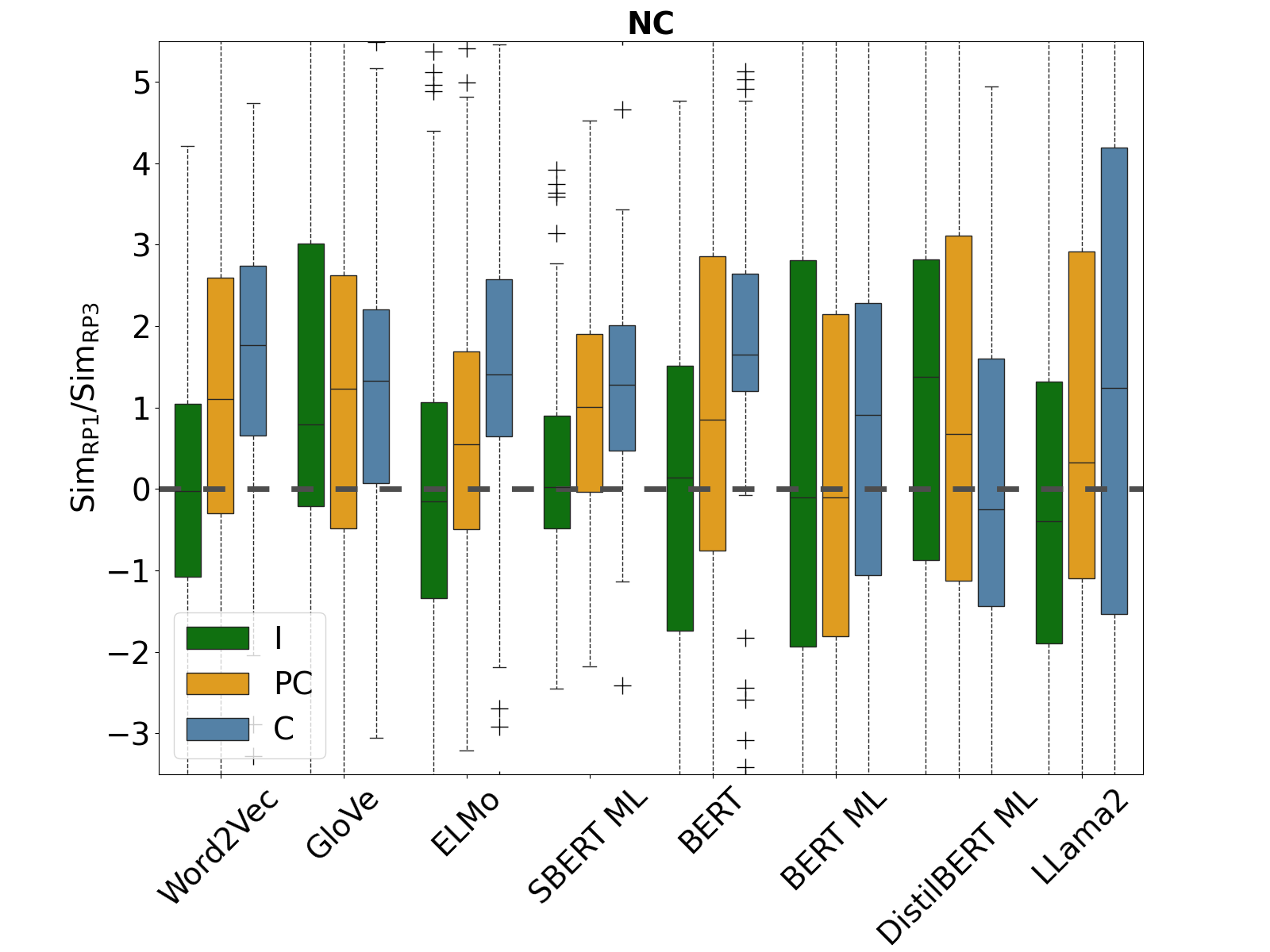} 
\end{subfigure}
\begin{subfigure}{0.5\textwidth}
    \centering
              \hspace{4pt} 
        ${\mbox{EN-Neut}}$
\includegraphics[scale=0.19,trim=20mm 0mm 40mm 0mm, clip=true]{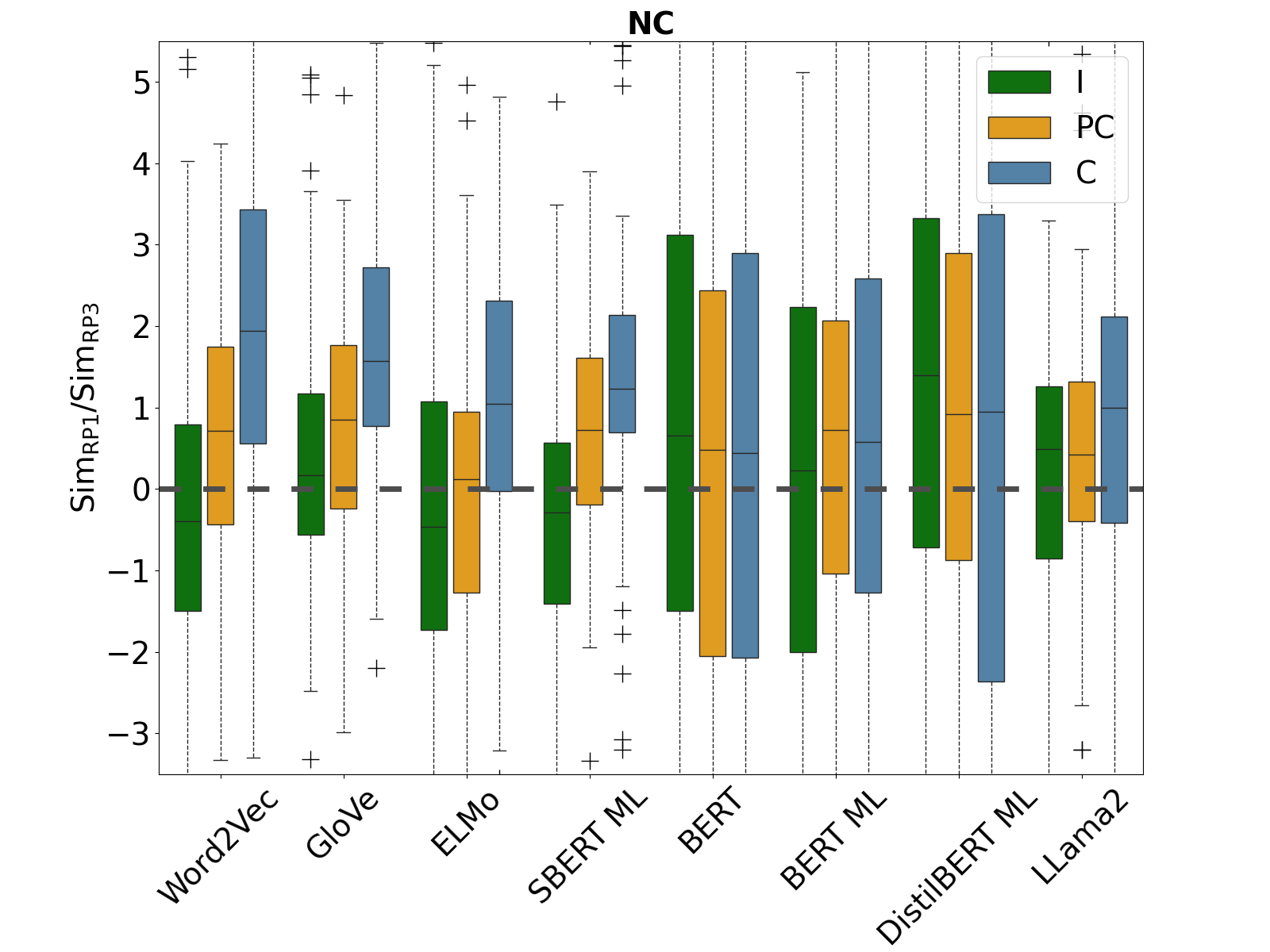} 
\end{subfigure}
\begin{subfigure}{0.5\textwidth}
    \centering
    \hspace{4pt} 
        ${\mbox{PT-Neut}}$
\includegraphics[scale=0.19,trim=20mm 0mm 40mm 0mm, clip=true]{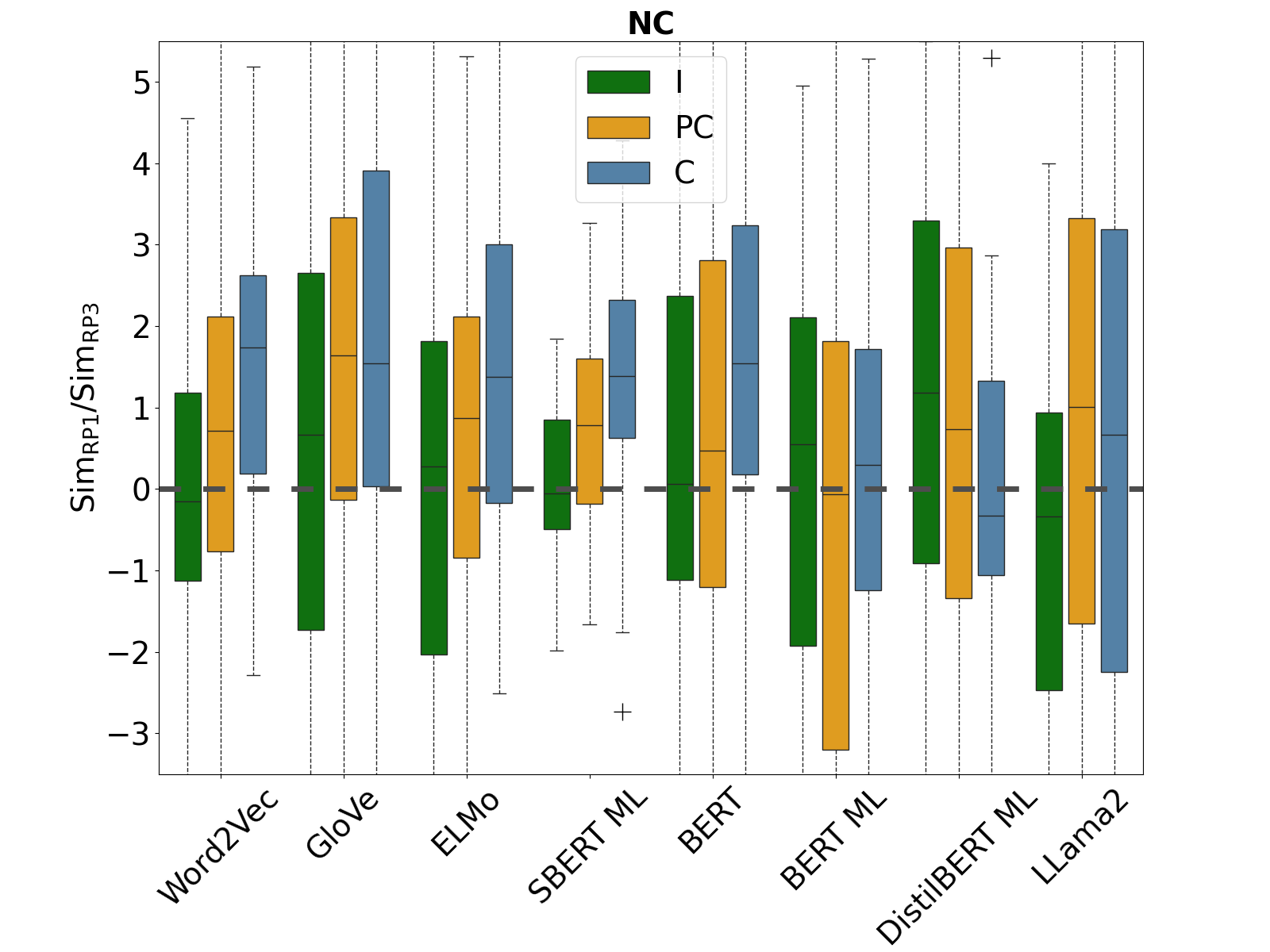} 
\end{subfigure}
\caption{Ratio of Average of Scaled Similarity ({$\mbox{Sim}_{R|Syn}$}/{$\mbox{Sim}_{R|WordsSyn}$}), per Compositionality Class: green for Idiomatic (I), orange for Partly Compositional (PC) and blue for Compositional (C), in English (EN) and Portuguese (PT), in Naturalistic (Nat) and Neutral (Neut) sentences. 
}
\protect \label{fig:p1p3-Class}
\end{figure}

If random substitutions that should result in Affinities around 1 ({A$_{Syn|Rand}$} in Figure \ref{fig:affinities} (Ideal Values)) result  instead in values mostly below 0.5, the latter may represent the de facto upperbound for Affinity for these models. In this case, a rescaling factor may need to be adopted that could magnify meaningful similarity values. To implement this, we propose the  Scaled Similarity (eq.\ref{simR}), which takes into account the threshold defined by random replacements when calculating the cosine similarities  between the target representation and a given probe. In this section we explore the behavior of {$\mbox{Sim}_{R|Syn}$ and $\mbox{Sim}_{R|WordsSyn}$} defined in section \ref{metrics}.

The Scaled Similarity values (Figure \ref{fig:loss-of-meaning}) reveal, even more than the Affinities, the equivalences displayed by the behavior of these models, with $\mbox{Sim}_R$ being able to abstract away from the particularities of the spaces defined by each of these models.    
Interestingly, comparing different levels of contextualisation  (e.g. 
static models on the left and 
contextualised on the right half of Figure \ref{fig:loss-of-meaning}) the Scaled Similarities produced by static models like Word2Vec and GloVe are comparable to those by a contextualised large language model like LLaMA2. These results seems to hold independently of how informative the context is (naturalistic vs. neutral sentences), with NC representations  from naturalistic sentences displaying no real advantage over those from neutral sentences. Overall, these results suggest that the informative contexts provided by the naturalistic sentences may not yet be adequately incorporated even by the larger contextualised models. 

Inspecting the {$\mbox{Sim}_{R|Syn}$} values according to idiomaticity level  (Figure \ref{fig:loss_of_meaning-p1}), the models display lower Scaled Similarities  for the more idiomatic than for the more compositional NCs, confirming what was already indicated by the Affinities that the models are less able to capture the idiomatic meanings and as a consequence the expected high similarities with their gold standard synonyms are not observed. This is further confirmed by analysing the values obtained for the synonyms of the individual components (Figures \ref{fig:loss-of-meaning} and \ref{fig:p3})
with the distributions of {$\mbox{Sim}_{R|WordsSyn}$} values having similar averages but considerably lower variances when compared to {$\mbox{Sim}_{R|Syn}$}, whereas the expected result would be the opposite: lower averages and variances for {$\mbox{Sim}_{R|Syn}$}. 
In fact the average and standard deviation for the ratio 
{$\mbox{Sim}_{R|Syn}/\mbox{Sim}_{R|WordsSyn}$}
 (Figure \ref{fig:p1p3}) show that the ratio oscillates around 1, which indicates that as a whole the models respond similarly to {P$_{Syn}$} and {P$_{WordsSyn}$} substitutions.
 In addition, the average values and variances for {$\mbox{Sim}_{R|WordsSyn}$} do not depend on the degree of compositionality of the target NC (Figure \ref{fig:p3} for {$\mbox{Sim}_{R|WordsSyn}$} and Figure  \ref{fig:p1p3-Class} for the average and standard deviation for the ratio {$\mbox{Sim}_{R|Syn}/\mbox{Sim}_{R|WordsSyn}$}, according to idiomaticity level). 
 The whole picture indicates that for all models (contextualised or not) replacing the NC by literal synonyms of the component words is more effective (produces higher similarities) than using their gold synonyms. In particular for idiomatic NCs   we observe that {$\mbox{Sim}_{R|Syn} < \mbox{Sim}_{R|WordsSyn}$} which indicates that the lexical similarity (as opposed to the similarity of meaning) is still a dominant factor in the representations even for the contextualised models, and provides additional confirmation for the possibility that the component words of an idiomatic NC may be represented individually and combined compositionally by these models.

Table \ref{tab:results_probes1} summarises these results in terms of the Spearman correlations between $\mbox{Sim}_{R}$ values and the human judgments for compositionality. It shows that, considering the different models, {$\mbox{Sim}_{R|Syn}$} is almost always moderately \marco{correlated} with the compositionality score: the higher the compositionality score, the higher the value  {$\mbox{Sim}_{R|Syn}$}, and consequently the more the meaning is preserved with a {P$_{Syn}$} substitution. {$\mbox{Sim}_{R|WordsSyn}$}, in contrast, rarely displays significant correlation with compositionality score. As discussed above this is a demonstration that the idiomatic meaning is not captured by these models, not even by those that are contextualised. As with Affinities, this discrepancy in the behavior of Scaled Similarities persists even after removing compounds from the dataset that have lexical overlaps with the NC$_{Syn}$ produced by the annotators (see Table \ref{tab:results_scale_remove} in the appendix).

\begin{table*}[!ht]
\centering
\begin{footnotesize}
\begin{tabular}{|l|cccccccc|}
\hline
{$\mbox{Sim}_{R|Syn}$}	&	Word2Vec	&	GloVe	&	ELMo	&	SBERT  	&	BERT	&	BERT 	&	DistilBERT 	&	LLaMA2	 \\ 
    	&		&		&		&	  ML	&	 	&	 ML	&	 ML	&		 \\ \hline
EN-Nat	&	0.61	&	0.57	&	0.61	&	0.66	&	0.63	&	0.67	&	0.61	&	0.38	\\ 

EN-Neut 	&	0.60	&	0.56	&	0.62	&	0.64	&	0.60	&	0.54	&	0.57	&	0.37	\\ 

PT-Nat	&	0.46	&	0.39	&	0.47	&	0.48	&	0.51	&	0.45	&	0.37	&	0.40	\\ 

PT-Neut	&	0.44	&	0.41	&	0.45	&	0.48	&	0.46	&	0.41	&	0.38	&	0.32	\\ \hline \hline
												
{$\mbox{Sim}_{R|Syn}$}	&	Word2Vec	&	GloVe	&	ELMo	&	SBERT 	&	BERT	&	BERT 	&	DistilBERT	&	LLaMA2	\\ 
     	&		&		&		&	  ML	&	 	&	 ML	&	 ML	&		 \\ \hline
EN-Nat 	&	-	&	-	&	0.17	&	0.18	&	0.19	&	0.36	&	0.33	&	0.25	\\ 
EN-Neut	&	-	&	-	&	0.14	&	0.12	&	-	&	-	&	0.29	&	0.17	\\ 
PT-Nat &	-	&	-	&	0.10	&	0.14	&	0.25	&	0.16	&	-	&	-	\\ 
PT-Neut	&	-	&	-	&	-	&	-	&	0.12	&	-	&	-	&	-	\\ \hline
\end{tabular}
\end{footnotesize}
\caption{Spearman $\rho$ correlation between the Scale Similarities and human judgments, for {$\mbox{Sim}_{R|Syn}$} and {$\mbox{Sim}_{R|WordsSyn}$} in both English and Portuguese. Non-significant (p $>$ 0.05) results were omitted from the table. Although these values shown are for the correlations at the NC level, the correlations at the Sentence level are comparable.}
\label{tab:results_probes1}
\end{table*}

\subsection{How are the results across models and languages?}

We have evaluated several vector models from different architectures in two languages, ranging from static to contextual representations as well as monolingual and multilingual models.
Although the results are generally far from being satisfactory, in this section we highlight some differences and similarities between models and languages.

Across models, the similarities are in general higher for Transformer-based models than for static representations. In this respect, it is worth noting that the results of ELMo and mSBERT are as similar to those of Word2Vec and GloVe than to the other BERT variants (for instance in Figure \ref{fig:p1-3-NC}). Although further research would be needed to determine the precise factors, for ELMo this behaviour  could be due either to the different vector space constructed by LSTMs or due to the smaller number of hidden layers when compared to the other models (2 vs. 6 and 12 layers), which may imply lower contextualisation effects across the network \cite{ethayarajh-2019-contextual}.

For the Transformer-based models, there are clear differences between the similarities produced by the BERT-based models and those of the autoregressive models, which are lower and with a wider range, especially for neutral sentences. When comparing monolingual and multilingual models, namely BERT and BERT ML, similar tendencies are found both in similarities and in correlations with the human judgments. In general, multilingual models seem to place the vector representations in a more restricted space, implying higher degrees of similarity and lower ranges of variation. Similar tendencies are found for DistilBERT-ML.

The proposed measures also suggest that the representations of the large autoregressive models are more similar to those of the static embeddings than to the other Transformer-based encoder models.

Although the results of the different models across languages  follow very similar trends, they also display two main differences. The first one is that when comparing the minimal pairs of the naturalistic data, the representations in English seem to be closer and occupy less space than those in Portuguese, in both monolingual and multilingual models of all types. The second is that for neutral sentences, there are larger differences than for naturalistic sentences, especially at the sentence level in both languages and similar results at the NC level, except for ELMo and BERT embeddings in {P$_{WordsSyn}$} and {P$_{Rand}$} (Figure \ref{fig:p1-3-NC}). The trends are even more aligned when considering  Affinities and Scaled Similarities for most models in both languages. 

\begin{figure}[!h!]
    \centering
   \includegraphics[scale=0.45]{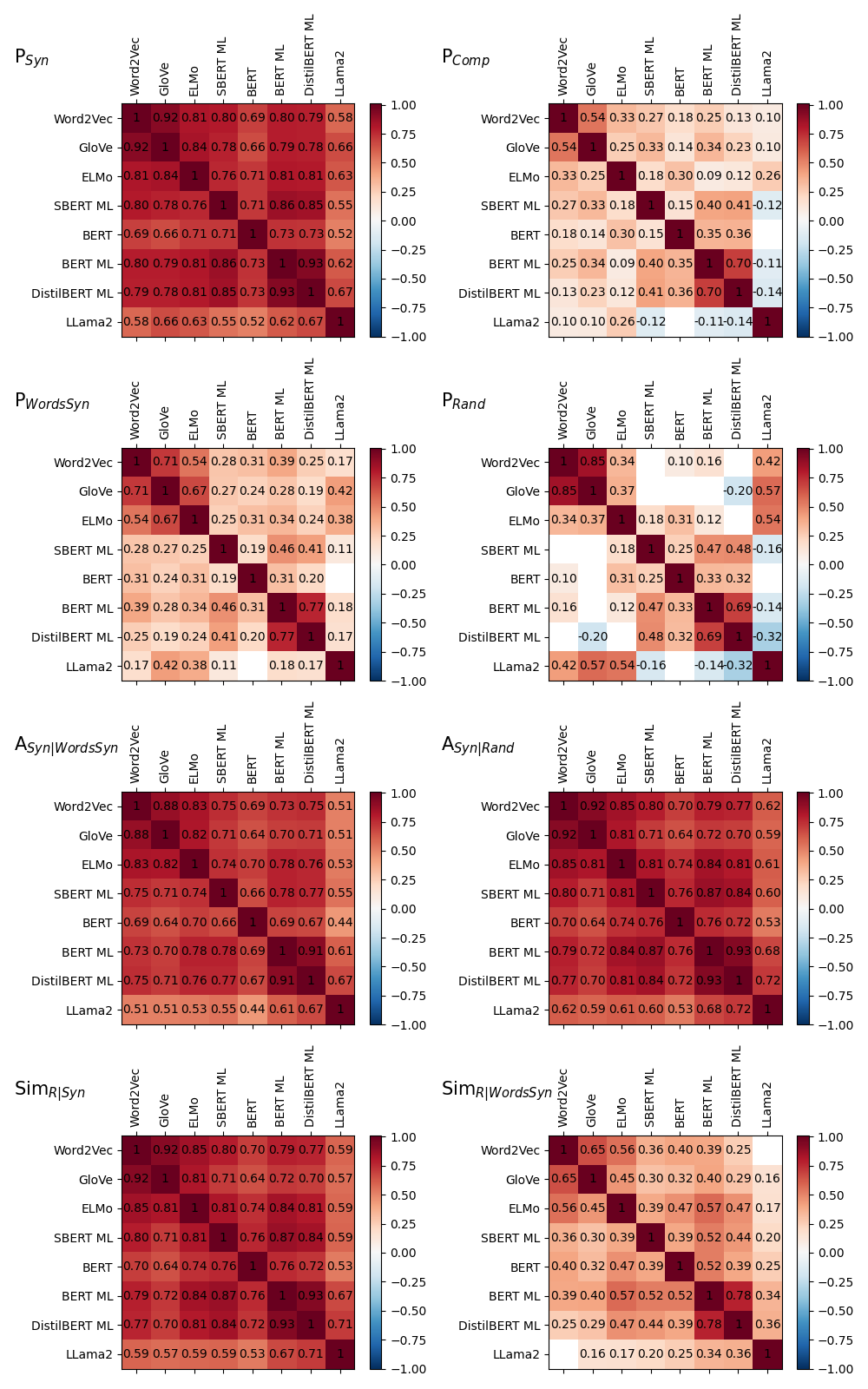}

\caption{Correlograms for all models for all sentences and languages, with only significant values displayed (p < 0.05). Red indicates positive correlation and blue negative; darker shades are for higher values, lighter for lower values.} 
\protect \label{fig:corr}
\end{figure}

Indeed, high correlations were found among all models, reflected by the correlogram in Figure \ref{fig:corr}. Correlations are particularly high for the expected congruent variants involving NC$_{Syn}$, as reflected by the darker red shades:  {P$_{Syn}$, A$_{Syn|WordsSyn}$ 
and Sim$_{R|Syn}$. } They are also higher for Affinities and Scaled Similarities, indicating that taking into account the relative preferences and random similarities within each model reveals how comparable they are in their ability to represent idiomaticity. That is, regardless of any superiority of specific models for other tasks, and in spite of their seemingly different individual performances in terms of cosines similarities in terms of idiomaticity representation this sample of models has not revealed one that is clearly better than the others. Moreover, high correlations with the static models also suggests that the relevant contextual clues for idiomaticity representation are not yet adequately incorporated by the contextualised models. 
 

In sum, our results indicate that the different models evaluated in general follow the same tendencies when representing idiomaticity in context, suggesting that they are not yet able to adequately capture the semantics of the MWEs.  More investigation is needed to determine how to effectively achieve this with these architectures and training regimes, or whether a change in paradigm is required. 
We will now discuss some representative cases, to give a flavour of how these models handle a spectrum of idiomaticity.   


\subsection{Analysing example cases}
\label{sec:examples}

\begin{table*}[!ht]
\centering
\scalebox{0.92}{
  \begin{tabular}{|c|l|p{1.8cm}p{1.8cm}p{7cm}|}
\hline
& NC & $NC_{Syn}$ & $NC_{WordsSyn}$ & Examples \\ \hline
\multirow{15}{*}{\rotatebox[origin=c]{90}{Idiomatic}} & grey matter & brain & silvery material & S1: \textit{Give your \textbf{grey matter} the workout that it needs to stay sharp and focused.} \newline S2: \textit{More ideas will follow when I get the \textbf{grey matter} functioning.} \newline S3: \textit{These youngsters can be encouraged to use their \textbf{grey matter} if the presentation is right.} \\
& eager beaver & hard worker & restless rodent & S1: \textit{Eric was being an \textbf{eager beaver} and left work late.}  \newline S2: \textit{\textbf{Eager beavers} willing to learn your job for less pay are almost always waiting in the wings.}  \newline S3: \textit{If you are a really \textbf{eager beaver} you can pre-order the DVD now from either of the below retailers.}\\
\hline 
%
\hline
\multirow{15}{*}{\rotatebox[origin=c]{90}{Partly Compositional}}& Dutch courage & alcoholic courage & Hollander bravery & S1: \textit{We had to go down to the pub to get some \textbf{Dutch courage}!} \newline S2: \textit{We suggest you try the following cocktail to work up a bit of \textbf{Dutch courage} to get you through the match!} \newline S2: \textit{After some \textbf{Dutch courage}(a few vodkas) in the nightclub, and a nerve-racking conversation, we kissed!}\\
& eternal rest & death  & permanent break & S1: \textit{They have been called home to their \textbf{eternal rest} and we are left behind.} \newline S2: \textit{These tolls announce the death of a nun and call for prayers for her \textbf{eternal rest}.} \newline S3: \textit{The passengers, with early morning porridge complexions, don't look far from \textbf{eternal rest}.}\\
\hline 
%
%
\hline
\multirow{15}{*}{\rotatebox[origin=c]{90}{Compositional}} & economic aid & financial assistance & budgetary assistance & S1: \textit{We have already extended to Greece certain types of relief and \textbf{economic aid} but these are inadequate.} \newline S2: \textit{The USSR was soon giving Cuba \textbf{economic aid}, technical support and military 'advisers' from the USSR.} \newline S3: \textit{A government's success in reducing population movement should be a key factor in allocating \textbf{economic aid}.}\\
& research lab & research facility & investigation workplace & S1: \textit{The fourth year is spent doing a research project in a 'real' \textbf{research lab}.} \newline S2: \textit{Being part of a \textbf{research lab} provides at times very exciting fieldwork experiences for individual students.} \newline S3: \textit{Bath operates several undergraduate degree programmes that include a professional placement year in industry or a \textbf{research lab}.} \\ 
\hline
\end{tabular}
}
\caption{Compositional NCs, $NC_{Syn}$, $NC_{WordSyn}$ and three sentences for qualitative analyses.}
\label{tab:examples3}
\end{table*}

\begin{table*}[hhhhhht!]
\centering
\begin{footnotesize}
\begin{tabular}{|l|cccccccc|}
\hline
 $gray$	&	Word2Vec	&	GloVe	&	ELMo	&	SBERT  	&	BERT	&	BERT 	&	DistilBERT 	&	LLama2	 \\ 
 $matter$   	&		&		&		&	  ML	&	 	&	 ML	&	 ML	&		 \\ \hline
P$_{Syn}$	&	0.27	&	0.37	&	0.45(0.03)	&	0.27(0.01)	&	0.68(0.14)	&	0.78(0.05)	&	0.77(0.01)	& 0.81(0.04)	\\ 
P$_{Comp}$	&	0.86	&	0.84	&	0.80(0.01)	&	0.77(0.00)	&	0.89(0.01)	&	0.90(0.02)	&	0.92(0.01)	&	0.89(0.01)	\\ 
P$_{WordsSyn}$	&	0.58	&	0.59	&	0.66(0.01)	&	0.59(0.02)	&	0.91(0.02)	&	0.88(0.03)	&	0.85(0.02)	&	0.69(0.00)	\\ 
P$_{Rand}$	&	0.47	&	0.52	&	0.61(0.02)	&	0.49(0.00)	&	0.92(0.02)	&	0.87(0.02)	&	0.87(0.02)	&	0.68(0.02)	\\ 
A$_{Syn|WordsSyn}$	&	-0.31	&	-0.22	&	-0.21(0.02)	 &  -0.32(0.02)	&	-0.23(0.15) 	&	-0.10(0.02)	&	-0.08(0.01)	&	0.12(0.04)	\\ 
A$_{Syn|Rand}$	&	-0.20	&	-0.15	&	-0.16(0.02)	&	-0.22(0.01)	&	-0.25(0.16) 	&	-0.09(0.04)	&	-0.09(0.01)&	0.14(0.04)	\\  
Sim$_{R|Syn}$	&	0.11		&	-0.13		&	-0.37		&	-0.39		&	-1.54		&	-1.64		&	-1.72		&	0.42		\\
Sim$_{R|WordsSyn}$	&	-0.02		&	0.25		&	0.15		&	0.15		&	0.04		&	0.02		&	-0.08		&	0.04		\\ \hline \hline
															
$eager$	&	Word2Vec	&	GloVe	&	ELMo	&	SBERT 	&	BERT	&	BERT 	&	DistilBERT	&	LLama2	\\ 
$beaver$     	&		&		&		&	  ML	&	 	&	 ML	&	 ML	&		 \\ \hline
P$_{Syn}$	&	0.34(0.02)	&	0.41(0.04)	&	0.68(0.05)	&	0.40(0.02)	&	0.78(0.06)	&	0.82(0.02)	&	0.83(0.01)	&	0.66(0.05)	\\ 
P$_{Comp}$   &       0.87	(0.01)	&	0.85(0.01)	&	0.79(0.05)	 &	0.82(0.02)	&	0.89(0.00)	&	0.94(0.01)	&	0.93(0.01)	&	0.78(0.16) \\
P$_{WordsSyn}$ &	0.45(0.03)	&	0.49(0.00)	&	0.84(0.03)	&	0.58(0.04)	&	0.86(0.04)	&	0.87(0.03)	&	0.85(0.01)	&	0.53(0.03) \\
P$_{Rand}$ &	0.41(0.01)	&	0.33(0.04)	&	0.72(0.14)	&	0.51(0.02)	&	0.92(0.03)	&	0.86(0.00)	&	0.89(0.01)	&	0.60(0.10). \\
A$_{Syn|WordsSyn}$	&	-0.10	(0.00)	&	-0.08(0.04)	&	-0.16(0.04)	&	-0.18(0.02) 	&	-0.07(0.04)	&	-0.05(0.00)& -0.02(0.00)	& 0.13(0.04) \\
A$_{Syn|Rand}$	&	-0.06	(0.03)	&	0.08	(0.00)	&	-0.04	(0.15)	&	-0.11(0.01)	&	-0.13(0.08)	&	-0.04(0.02)	&	-0.07(0.00)	&	0.06(0.05) \\
Sim$_{R|Syn}$	&	0.09		&	0.08		&	-0.10		&	-0.14		&	-0.79		&	-0.78		&	-0.90		&	0.15	  \\
Sim$_{R|WordsSyn}$	&	-0.13		&	0.15		&	0.23		&	0.26		&	-0.14		&	-0.24		&	-0.42		&	-0.20	  \\ \hline \hline

 $Dutch$	&	Word2Vec	&	GloVe	&	ELMo	&	SBERT  	&	BERT	&	BERT 	&	DistilBERT 	&	LLama2	 \\ 
 $courage$   	&		&		&		&	  ML	&	 	&	 ML	&	 ML	&		 \\ \hline
P$_{Syn}$	&	0.63	&	0.64	&	0.88(0.02)	&	0.75(0.01)	&	0.94(0.03)	&	0.93(0.01)	&	0.90(0.00)	& 0.77(0.00)	\\ 
P$_{Comp}$	&	0.77	&	0.76	&	0.91(0.01)	&	0.82(0.00)	&	0.88(0.04)	&	0.90(0.01)	&	0.90(0.01)	&	0.67(0.14)	\\ 
P$_{WordsSyn}$	&	0.57	&	0.53	&	0.82(0.01)	&	0.69(0.02)	&	0.88(0.03)	&	0.87(0.01)	&	0.89(0.01)	&	0.54(0.03)	\\ 
P$_{Rand}$	&	0.41	&	0.35	&	0.79(0.01)	&	0.51(0.00)	&	0.90(0.04)	&	0.92(0.01)	&	0.90(0.00)	&	0.59(0.00)	\\ 
A$_{Syn|WordsSyn}$	&	0.07	&	0.11	&	0.06(0.02)	 &  0.06(0.02)	&	0.06(0.04) 	&	0.05(0.02)	&	0.00(0.00)	&	0.23(0.03)	\\ 
A$_{Syn|Rand}$	&	0.22	&	0.28	&	0.09(0.02)	&	0.24(0.01)	&	0.04(0.06) 	&	0.01(0.01)	&	-0.01(0.00)&	0.18(0.00)	\\  
Sim$_{R|Syn}$ 	&      0.32 &	0.37	&	0.42	&	0.46	&	0.38	&	0.28	&	0.09	&	0.44		\\ 
Sim$_{R|WordsSyn}$	&	0.21	&	0.32	&	0.22	&	0.26	&	0.06	&	-0.17	&	-0.33	&	-0.12	\\  \hline \hline
																	
$eternal$	&	Word2Vec	&	GloVe	&	ELMo	&	SBERT 	&	BERT	&	BERT 	&	DistilBERT	&	LLama2	\\ 
$rest$     	&		&		&		&	  ML	&	 	&	 ML	&	 ML	&		 \\ \hline
P$_{Syn}$	&	0.43&	0.53	&	0.48(0.01)	&	0.53(0.02)	&	0.74(0.04)	&	0.86(0.01)	&	0.76(0.02)	& 0.71(0.04)	\\ 
P$_{Comp}$	&	0.92 &	0.89	&	0.80(0.02)	&	0.82(0.01)	&	0.87(0.03) &	0.93(0.01)	&	0.09(0.00)  &	0.75(0.06)	\\ 
P$_{WordsSyn}$	&	0.43	&	0.53	&	0.60(0.02)	&	0.71(0.02)	&	0.83(0.05)	&	0.83(0.05)	&	0.82(0.01)	&	0.66(0.03)	\\ 
P$_{Rand}$	&	0.44	&	0.38	&	0.64(0.01)	&	0.43(0.01)	&	0.80(0.08)	&	0.87(0.03)	&	0.89(0.00)	&	0.62(0.04)	\\ 
A$_{Syn|WordsSyn}$	&	-0.00	&	-0.01	&	-0.12(0.02)	&	-0.18(0.01)	&	-0.09(0.08) 	&	0.03(0.06)	&	-0.06(0.01)&	0.05(0.01)	\\  
A$_{Syn|Rand}$	&	-0.00	&	0.14	&	-0.15(0.00)	 &  0.10(0.02)	&-0.06(0.08) 	&	-0.01(0.03)	&	-0.13(0.02)	&	0.09(0.02)	\\ 
Sim$_{R|Syn}$	&	0.15	 	&	0.14		&	-0.06		&	-0.00		&	-0.25		&	-0.15		&	-0.62		&	0.23		\\
Sim$_{R|WordsSyn}$	&	-0.34	 	&	0.16		&	0.05		&	0.22		&	0.17		&	0.10		&	-0.29		&	0.11		\\ \hline \hline

 $economic$	&	Word2Vec	&	GloVe	&	ELMo	&	SBERT  	&	BERT	&	BERT 	&	DistilBERT 	&	LLama2	 \\ 
 $aid$   	&		&		&		&	  ML	&	 	&	 ML	&	 ML	&		 \\ \hline
P$_{Syn}$	&	0.65	&	0.80	&	0.90(0.01)	&	0.77(0.01)	&	0.89(0.02)	&	0.96(0.01)	&	0.94(0.01)	& 0.95(0.02)	\\ 
P$_{Comp}$	&	0.80	&	0.88	&	0.89(0.00)	&	0.78(0.00)	&	0.90(0.04)	&	0.93(0.03)	&	0.92(0.00)	&	0.90(0.02)	\\ 
P$_{WordsSyn}$	&	0.64	&	0.74	&	0.92(0.00)	&	0.78(0.01)	&	0.89(0.00)	&	0.95(0.01)	&	0.92(0.01)	&	0.90(0.02)	\\ 
P$_{Rand}$	&	0.65	&	0.73	&	0.70(0.07)	&	0.58(0.01)	&	0.76(0.05)	&	0.90(0.02)	&	0.90(0.01)	&	0.70(0.03)	\\ 
A$_{Syn|WordsSyn}$	&	0.02	&	0.06	&	-0.01(0.01)	 &  -0.00(0.00)	&	-0.00(0.02) 	&	0.01(0.01)	&	0.02(0.00)	&	0.05(0.01)	\\ 
A$_{Syn|Rand}$	&	0.01	&	0.07	&	0.20(0.06)		& 0.20(0.01) 	&  0.14(0.05)	&	0.06(0.01) 	&	0.04(0.01)	&	0.25(0.02)	\\  
Sim$_{R|Syn}$	&	0.27		&	0.22		&	0.32		&	0.47		&	0.56		&	0.53		&	0.52		&	0.84		\\
Sim$_{R|WordsSyn}$	&	0.42		&	0.28		&	0.24		&	0.41		&	0.58		&	0.49		&	0.42		&		0.66	\\  \hline \hline

$research$	&	Word2Vec	&	GloVe	&	ELMo	&	SBERT 	&	BERT	&	BERT 	&	DistilBERT	&	LLama2	\\ 
$lab$     	&		&		&		&	  ML	&	 	&	 ML	&	 ML	&		 \\ \hline
P$_{Syn}$	&	0.71&	0.82	&	0.91(0.03)	&	0.78(0.02)	&	0.93(0.01)	&	0.97(0.00)	&	0.95(0.00)	 &    0.96(0.01)	\\ 
P$_{Comp}$	&	0.86 &	0.88	&	0.90(0.00)	&	0.88(0.01)	&	0.89(0.04) &	0.94(0.01)	&	0.94(0.01)  &	0.87(0.05)	\\ 
P$_{WordsSyn}$	&	0.47	&	0.51	&	0.68(0.02)	&	0.72(0.02)	&	0.67(0.09)	&	0.90(0.01)	&	0.90(0.01)	&	0.78(0.11)	\\

P$_{Rand}$	&	0.39	&	0.40	&	0.72(0.04)	&	0.40(0.01)	&	0.73(0.07)	&	0.88(0.01)	&	0.90(0.01)	&	0.68(0.08)	\\ 
A$_{Syn|WordsSyn}$	&	0.23	&	0.30	&	0.23(0.01)	&	0.06(0.02)	&	0.26(0.08) 	&	0.07(0.01)	&	0.05(0.01)&	0.18(0.11)	\\

A$_{Syn|Rand}$	&	0.32	&	0.41	&	0.19(0.01)	 &  0.38(0.03)	&0.20(0.06) 	&	0.09(0.01)	& 0.06(0.01)	&	0.28(0.07)	\\ 
Sim$_{R|Syn}$	&	0.52		&	0.69		&	0.68		&	0.63		&	0.73		&	0.72		&	0.55		&	0.87		\\
Sim$_{R|WordsSyn}$	&	0.14		&	0.18		&	-0.17		&	0.53		&	-0.20		&	0.17		&	0.06		&	0.35	\\  \hline

\end{tabular}
\end{footnotesize}
\caption{Similarity, Affinity and Scaled Similarity values for the NCs selected in Table \ref{tab:examples3}. Values in parentheses represent the standard deviations among the three sentences. The static models are independent of context, and for them, the variance is omitted, except in the case of \emph{eager beaver} where there is a sentence where the compound appears in plural form. }\label{tab:values7}
\end{table*}


For a more concrete qualitative overview of the ability of models in representing different levels of idiomaticity, we now look at some representative English NCs evenly distributed among the three levels of compositionality (compositional,  partly compositional and idiomatic) in three naturalistic sentences (Table \ref{tab:examples3}). We start with the probes for 6 English NCs  and then look at the highest and lowest values for the A$_{Syn|WordsSyn}$ affinity focusing on the relation between a given NC and its $NC_{Syn}$ and $NC_{WordSyn}$ variants. 

\paragraph{Probes}
Considering the probing measures in terms of the average scores of all sentences for each of the 6 NCs (Table \ref{tab:values7}), we focus on the cosine similarities for the probes and whether they differ from the expected behavior compatible with capturing idiomatic meaning. 

First of all, for {P$_{Syn}$} the similarities should be close to 1. Indeed, at sentence level all similarities for all models are above 0.9, and tend to be higher for compositional NCs (0.98) than for partly compositional (0.95) and than for idiomatic NCs (0.90).  However, at NC level, they display considerable variation, and while the similarities are high for all models for compositional NCs, for idiomatic NCs, in particular, the similarities are the lowest and vary considerably per model   
(from 0.27 for SBERT ML and Word2Vec to 0.81 for LLama2   
for \textit{grey matter}).  For partly compositional NCs, 
although some of the models assign the expected high similarities for some NCs (0.94 for BERT for \emph{Dutch courage}),  
other NCs have lower similarities (0.43 for Word2Vec for \textit{eternal rest}). 

For {P$_{Comp}$}, lower similarities are expected for idiomatic NCs, as the idiomatic meaning may be lost when one of the component words is missing (e.g. \emph{grey matter} vs. \emph{grey} or vs. \emph{matter}). However, at sentence level they are higher than 0.93 for all models. At NC level, although these idiomatic NCs have lower similarities they are still high (from 0.77 for SBERT ML for \textit{grey matter} to 0.94 for BERT ML for \emph{eager beaver}). For partly compositional and compositional NCs they are mostly high for all models, except for LLama 2 for \emph{Dutch courage} (0.67). 

Although lower {P$_{WordsSyn}$} were also expected for more idiomatic NCs, at sentence level all idiomatic, partly compositional and compositional NCs, display similarities above 0.95, even though the $NC{WordSyn}$ in {P$_{WordsSyn}$} does not preserve the idiomatic meaning (e.g. \emph{grey matter} vs. \emph{silvery material}).
At NC level, even if lower values were found for idiomatic NCs with static models (Word2Vec and GloVe), high similarities were still found (e.g. 0.91 for BERT for \emph{gray matter}). 

Finally, for {P$_{Rand}$}, there should be low similarities for all NCs and randomly generated substitutions. However, most of the similarities are still high, regardless of the level of idiomaticity (e.g. 0.92 for BERT for \emph{grey matter} and for BERT ML for \emph{Dutch courage}). 

Overall, the expected high similarities for {P$_{Syn}$} are not displayed by these models at NC level, and for the other probes the perturbations to the idiomatic meaning are not reflected by lower similarities.

\paragraph{Affinities}

For the affinity measures, considering the examples with the highest and lowest values for {A$_{Syn|WordsSyn}$} as a proxy for how a particular model represents an NC compared to its synonym and to a word-by-word replacement (NC$_{Syn|WordsSyn}$), we focus on the results for BERT in the naturalistic sentences in English. As discussed in section~\ref{sec:results_aff}, we expect higher {A$_{Syn|WordsSyn}$} values for idiomatic NCs, since the model should display a stronger preference for a semantically related synonym than to a potentially unrelated substitution, representing the former as closely as possible from the NC in a vector space. 
In contrast, for more compositional cases, both substitutions may be possible and close to one another (reflected by {A$_{Syn|WordsSyn}$} values around 0). 
However, the NCs with the highest {A$_{Syn|WordsSyn}$} values were mostly compositional (starting with \emph{video game} with {A$_{Syn|WordsSyn}$}$=0.44$, and \emph{parking lot} with {A$_{Syn|WordsSyn}$}$=0.40$), with the first partly compositional NC appearing at the position 16 (\emph{sparkling water} with {A$_{Syn|WordsSyn}$}$=0.32$). The idiomatic NC with the highest A$_{Syn|WordsSyn}$ value is at position 53  (\emph{box office}, referring to the popularity of a movie with {A$_{Syn|WordsSyn}$}$=0.24$).

At the other end of the ranking, we find mostly idiomatic cases. Among the top 10 examples with the lowest values we find 7 idiomatic (e.g., \emph{agony aunt} with {A$_{Syn|WordsSyn}$}$=-0.29$ and the NC with the lowest value, \emph{grey matter}, {A$_{Syn|WordsSyn}$}$=-0.40$), with 2 partly compositional and only one compositional NC in position 10 (\emph{cooking stove} with {A$_{Syn|WordsSyn}$}$=-0.24$).

In sum, these confirm that the models do not display the expected preference for representing NCs closer to their synonyms than to distractors, even when these involve idiomatic NCs and/or random items. 


\section{Conclusions}

This paper presented an evaluation of the ability of widely available word representation models to capture  idiomatic meaning, focusing on noun compounds in two languages, English and Portuguese. For evaluation we introduced the NCIMP dataset, containing NCS in English and Portuguese in naturalistic and neutral sentences forming minimal pairs with idiomatic probes using their component words, synonyms and other variant replacements, resulting in a dataset containing 29,900 items, extending the datasets by \citet{garcia-etal-2021-assessing} and \citet{garcia-etal-2021-probing}. 
These pairs can be used to measure the ability of models to detect the loss of the idiomatic meaning in the presence of lexical substitutions and different contexts. We also propose two types of measure for quantifying this ability: Affinities and Scaled Similarities. Affinity is a relative measure of the proximity of the NC to two alternative probes, determining which of them is the closest to the NC. Focusing on idiomaticity, we analysed if the models were able to generate a representation for a given NC that was more similar to a semantically related paraphrase given by the gold standard synonym than to an alternative possibly semantically unrelated representation. The proposed measures of scaled similarities, $\mbox{Sim}_R$, take sample random similarities into account for rescaling the space of a given model, to magnify high similarities and distinguish them from those that are artifacts of the characteristics of the landscape of that model. As a consequence, $\mbox{Sim}_R$ also seems to abstract away from the particularities of the semantic space of each model and provides a more direct way of comparing idiomaticity representation across models. The results obtained indicate that models are not able to accurately capture idiomaticity, as they fail to reflect actual similarities between NCs and their gold synonyms, especially for idiomatic cases, while at the same time not displaying enough awareness of perturbations that lead to changes in meaning, such as those involving the synonyms of the component words, and even random words. 
It seems that the lexical clues provided by the component words are prioritised when representing an NC over a more holistic combination of the relevant semantic clues needed for representing its idiomatic meaning.
Moreover, although the contexts could provide relevant information about the idiomatic meanings, they do not seem to be adequately incorporated in these widely adopted models, regardless of their degree of contextualisation. They also seem to fail to incorporate the relevant context for idiomaticity, seeing as static and contextualised models show comparable performances.

In this paper we evaluated the proposed measures focusing on idiomaticity, but they may be applied to other tasks, and serve as basis to detect unwanted biases towards non-target meanings more generally. Moreover, they may be informative when fine-tuning models to assess if the changes are going towards the intended target representations.  

\subsection{Future Work}

In this paper, we inspected the similarities produced by a number of models to determine how accurately they represent idiomatic expressions. The results obtained are that not even large models like LLama2  seem to display the expected patterns that would confirm idiomatic understanding. 

It is important to note that some of the difficulties in extracting information from cosine similarity measures may be attributed to the presence of rogue dimensions \cite{timkey2021all} rather than anisotropy in semantic space. Measures like Affinity and Scaled Similarity may not fully address this issue. We conducted a preliminary analysis using \citet{timkey2021all} method to identify and standardise the top three rogue dimensions per model/layer. After standardisation, we conducted an analysis focusing on P$_{Syn}$ measures and found correlations mostly above 0.85, except for BERT-PT-Neut (0.79) and LLama2-EN-Neut (0.65) (see Table \ref{tab:results_P1_rogue_original_cor} in the Appendix). Further investigation is needed to assess the impact of standardising these dimensions and different approaches for standardisation, but given the high correlations with our original results, we will leave this for future work.

Although our proposed assessment protocol and measures are model-independent, they rely on access to the models and to their representations for subwords, words and multiwords. Therefore, probing large generative AI chatbots for their understanding of idiomaticity, especially closed-source models,  presents additional challenges potentially requiring adaptation in the application of the protocol, due to the restricted access to their base models and of the potential variation in their answers. These warrant further investigations that are outside the scope of this paper. However, one possible alternative would be to perform probing using question-answering, following \citet{zeng2022getting}.
We illustrate this question-answering approach with recent AI Chatbots: GPT-3,  Gemini pro\footnote{\url{https://gemini.google.com/app}} \cite{team2023gemini} and ChatGPT4\footnote{\url{https://chat.openai.com/}}. For testing these models, simple questions containing idiomatic expressions are used, after having instructed each model to provide the shortest answer to each question before asking them. The assumption is that the questions could only be answered correctly if the model understood the meaning of the idioms in context. The questions and answers are included in Table \ref{tab:QA}. 

\begin{table}[!ht]
\centering
\small
\begin{tabular}{|>{\centering\arraybackslash}m{1.8cm} | >{\centering\arraybackslash}m{5cm} | >{\centering\arraybackslash}m{1.5cm} | >{\centering\arraybackslash}m{1.5cm} | >{\centering\arraybackslash}m{1.5cm} |}
\hline
\textbf{Idiom} & \textbf{Question} & \textbf{GPT-3} & \textbf{Gemini Pro} & \textbf{ChatGPT 4}\\
\hline
Never say die & If I have a never say die attitude, would I run the marathon injured or forfeit? & I would never run a marathon injured. & Run. & Run the marathon. \\
\hline
All at sea & If I am all at sea with my math assignment, am I making progress or am I lost? & You are making progress. & Lost. & Lost. \\
\hline
Every trick in the book & If I use every trick in the book to guarantee my grade, am I a magician or a determined student? & A magician. & Cheater. & Determined student. \\
\hline
Kill two birds in one stone & If I wanted to kill two birds in one stone, what kind of a workplace should I work in? & A slaughterhouse. & Multiskilled. & A multitasking environment. \\
\hline
Ahead of the game & If I want to be ahead of the game, would I study early or procrastinate? & Procrastinate. & Early. & Study early. \\
\hline
\end{tabular}
\caption{Questions used to probe the understanding of idioms and the answers provided by recent generative models.
The leftmost column lists idiomatic expressions, the second column presents hypothetical questions using these expressions, and the following columns show the responses from different models, including GPT-3, Gemini Pro, and ChatGPT 4. Answers by GPT-3 are from \cite{zeng2022getting}. }
\label{tab:QA}
\end{table}

The responses from different systems vary in terms of correctly interpreting the idioms. 
The responses from GPT-3 often miss the mark, while the responses from Gemini Pro and ChatGPT 4 are mixed, with some answers suggesting correct interpretations and some incorrect.
For instance, for ``Every trick in the book'', GPT-3 responds with ``A magician'', which is a literal interpretation, while ChatGPT 4 correctly identifies the figurative meaning with ``Determined student''. 
Gemini Pro response to the question is ``Cheater''. The idiom ``every trick in the book'' generally means to use all available means or strategies to achieve one's goal, often implying ingenuity or resourcefulness rather than dishonesty. The response from Gemini Pro could either be due to ``trick'' or it could be seen as a misinterpretation.
This table could also be seen as indicative of the evolution of AI language models over time, with newer models potentially being trained to better handle idiomatic expressions and context, as seen in the generally more accurate responses from ChatGPT 4 compared to GPT 3. 
Although the questions in the table are indeed useful for exemplifying the comprehension of idiomatic expressions by these models they only cover a very limited and focused sample.  In this paper, we propose the use of minimal pairs containing synonyms and other distractors for a more in-depth assessment of idiomatic understanding. Although their adaptation for a question-answering setting is left for future work, our results for open models is in line with comparative analyses of the ability of some of these models for idiomatic and figurative language \cite{Phelps:2024}.  

Moreover, as idiomatic expressions can be extremely diverse and nuanced,  a comprehensive evaluation of the ability of a model to understand them requires a controlled but extensive set of idiomatic expressions and their variations. Therefore, we  plan to extend the test items to contain additional types of multiword expressions, including verb-noun combinations and phrasal verbs. In addition for a larger crosslingual examination of idiomaticity, and in particular of whether multilingual models capture language-specific realisations of idiomatic expressions, we plan to extend the dataset with additional languages. These would also allow the investigation of factors relevant to specific tasks, such as machine translation, for which the translatability of MWEs from source into  target languages for may also affect performance when processing MWEs \cite{dankers-etal-2022-transformer}. 

Possible next steps also include extending the probing strategy with  additional measures that go beyond similarities and correlations. Moreover, for ambiguous NCs in particular, we intend to add sense-specific probes which could be used to measure and address training biases towards particular senses.  
Finally, this paper has focused the evaluation on off-the-shelf pre-trained models to provide an analysis of their ability to capture idiomaticity, and left the investigation of fine-tuned models for future work. 
In particular, although fine-tuning can improve model performance \cite{tayyar-madabushi-etal-2022-semeval} it is unclear to what extent the models are able to generalise beyond the specific items seen to other unseen idiomatic expressions, or if each new expression would have to be individually learned by the model. But these points are left for future investigation.     


\begin{acknowledgments}

This work was partly supported by UKRI EPSRC EP/T02450X/1 and NAF/R2/202209 (UK), by CNPq 311497/2021-7 and CAPES/PRINT 88887.583995/2020-00 (Brazil), by MCIN/AEI/10.13039/501100011033 (grants PID2021-128811OA-I00 and TED2021-130295B-C33, the latter also funded by ``European Union Next Generation EU/PRTR''), by the Galician Government (ERDF 2014-2020: Call ED431G 2019/04, ED431F 2021/01, and ED431F 2021/01), and by a Ramón y Cajal grant (RYC2019-028473-I), and by COST-Action UniDive.

\end{acknowledgments}

\starttwocolumn
\bibliography{compling_style,bib/anthology,bib/acl2020,bib/emnlp2020,bib/coling2020,bib/idioms,bib/bib2022}
\onecolumn

\appendix
\section{Appendix A: Measures for English and Portuguese}
\label{sec:appendix}

In this section we present the mean and standard deviation for the NCs in English and Portuguese in naturalistic and neutral sentences, for the different probes at Sentence level (Table \ref{tab:num_sent}) and at NC level (Table \ref{tab:num_NC}),  for Affinities (Table \ref{tab:num_Affinity}) and for Scaled Similarities (Table \ref{tab:num_S14_S34}).

\begin{table}[!h]
\small
\begin{center}
\begin{tabular}{|l|cc|cc|cc|cc|}
\hline
\multicolumn{9}{|c|}{P$_{Syn}$} \\ \hline
\multirow{2}{*}{Model Name} & \multicolumn{2}{c|}{EN-Nat} & \multicolumn{2}{c|}{EN-Neut} & \multicolumn{2}{c|}{PT-Nat} & \multicolumn{2}{c|}{PT-Neut} \\
                            & mean & std & mean & std & mean & std & mean & std \\
\hline
Word2Vec     & 0.985 & 0.012 & 0.811 & 0.083 & 0.968 & 0.025 & 0.883 & 0.062 \\
GloVe        & 0.990 & 0.008 & 0.868 & 0.063 & 0.980 & 0.018 & 0.931 & 0.054 \\
ELMo         & 0.974 & 0.022 & 0.841 & 0.070 & 0.938 & 0.045 & 0.782 & 0.116 \\
SBERT ML     & 0.974 & 0.022 & 0.810 & 0.101 & 0.955 & 0.035 & 0.833 & 0.096 \\
BERT         & 0.988 & 0.011 & 0.927 & 0.035 & 0.980 & 0.017 & 0.915 & 0.041 \\
BERT ML      & 0.992 & 0.007 & 0.924 & 0.040 & 0.984 & 0.012 & 0.929 & 0.044 \\
DistilBERT ML& 0.996 & 0.003 & 0.952 & 0.023 & 0.991 & 0.007 & 0.966 & 0.018 \\
LLama2       & 0.992 & 0.010 & 0.955 & 0.020 & 0.981 & 0.018 & 0.903 & 0.065 \\
\hline \hline
\multicolumn{9}{|c|}{P$_{Comp}$} \\ \hline
Word2Vec     & 0.996 & 0.004 & 0.941 & 0.018 & 0.987 & 0.011 & 0.957 & 0.026 \\
GloVe        & 0.996 & 0.003 & 0.955 & 0.011 & 0.993 & 0.006 & 0.982 & 0.012 \\
ELMo         & 0.989 & 0.009 & 0.914 & 0.019 & 0.966 & 0.020 & 0.890 & 0.035 \\
SBERT ML     & 0.990 & 0.007 & 0.922 & 0.021 & 0.982 & 0.013 & 0.929 & 0.029 \\
BERT         & 0.992 & 0.007 & 0.951 & 0.016 & 0.986 & 0.013 & 0.933 & 0.025 \\
BERT ML      & 0.996 & 0.003 & 0.957 & 0.016 & 0.993 & 0.005 & 0.962 & 0.016 \\
DistilBERT ML& 0.998 & 0.001 & 0.977 & 0.006 & 0.996 & 0.002 & 0.987 & 0.005 \\
LLama2       & 0.995 & 0.008 & 0.986 & 0.007 & 0.991 & 0.008 & 0.964 & 0.020 \\
\hline \hline
\multicolumn{9}{|c|}{P$_{WordsSyn}$} \\ \hline
Word2Vec     & 0.983 & 0.013 & 0.797 & 0.049 & 0.958 & 0.031 & 0.845 & 0.060 \\
GloVe        & 0.989 & 0.009 & 0.863 & 0.041 & 0.974 & 0.025 & 0.904 & 0.062 \\
ELMo         & 0.975 & 0.020 & 0.861 & 0.048 & 0.930 & 0.042 & 0.760 & 0.088 \\
SBERT ML     & 0.977 & 0.017 & 0.844 & 0.057 & 0.956 & 0.033 & 0.855 & 0.060 \\
BERT         & 0.983 & 0.014 & 0.919 & 0.032 & 0.967 & 0.025 & 0.891 & 0.038 \\
BERT ML      & 0.991 & 0.006 & 0.925 & 0.036 & 0.983 & 0.012 & 0.934 & 0.032 \\
DistilBERT ML& 0.995 & 0.003 & 0.952 & 0.016 & 0.990 & 0.006 & 0.963 & 0.014 \\
LLama2       & 0.986 & 0.014 & 0.945 & 0.021 & 0.977 & 0.017 & 0.891 & 0.052 \\
\hline \hline
\multicolumn{9}{|c|}{P$_{Rand}$} \\ \hline
Word2Vec     & 0.984 & 0.012 & 0.799 & 0.043 & 0.960 & 0.033 & 0.851 & 0.099 \\
GloVe        & 0.988 & 0.009 & 0.849 & 0.038 & 0.974 & 0.026 & 0.911 & 0.095 \\
ELMo         & 0.966 & 0.025 & 0.829 & 0.040 & 0.912 & 0.048 & 0.725 & 0.115 \\
SBERT ML     & 0.968 & 0.023 & 0.769 & 0.053 & 0.935 & 0.043 & 0.768 & 0.063 \\
BERT         & 0.979 & 0.018 & 0.924 & 0.027 & 0.956 & 0.028 & 0.886 & 0.033 \\
BERT ML      & 0.990 & 0.008 & 0.925 & 0.024 & 0.980 & 0.013 & 0.933 & 0.030 \\
DistilBERT ML& 0.995 & 0.004 & 0.951 & 0.012 & 0.990 & 0.007 & 0.967 & 0.016 \\
LLama2       & 0.980 & 0.019 & 0.937 & 0.015 & 0.962 & 0.026 & 0.879 & 0.058 \\
\hline
\end{tabular}
\caption{Mean and standard deviation at Sentence level for {P$_{Syn}$}, {P$_{Comp}$}, P$_{WordsSyn}$ and {P$_{Rand}$}, for English (EN) and Portuguese (PT) for naturalistic (Nat) and neutral (Neut) sentences. }\label{tab:num_sent}
\end{center}
\end{table}

\begin{table}[!h]
\small
\begin{center}
\begin{tabular}{|l|cc|cc|cc|cc|}
\hline
\multicolumn{9}{|c|}{P$_{Syn}$} \\ \hline
\multirow{2}{*}{Model Name} & \multicolumn{2}{c|}{EN-Nat} & \multicolumn{2}{c|}{EN-Neut} & \multicolumn{2}{c|}{PT-Nat} & \multicolumn{2}{c|}{PT-Neut} \\
 & mean & std & mean & std & mean & std & mean & std \\
\hline
Word2Vec     & 0.517 & 0.209 & 0.517 & 0.207 & 0.498 & 0.251 & 0.488 & 0.258 \\
GloVe        & 0.551 & 0.227 & 0.555 & 0.222 & 0.465 & 0.278 & 0.473 & 0.275 \\
ELMo         & 0.714 & 0.147 & 0.646 & 0.155 & 0.629 & 0.166 & 0.551 & 0.192 \\
SBERT ML     & 0.591 & 0.208 & 0.577 & 0.203 & 0.632 & 0.199 & 0.612 & 0.198 \\
BERT         & 0.816 & 0.086 & 0.854 & 0.060 & 0.824 & 0.090 & 0.831 & 0.079 \\
BERT ML      & 0.876 & 0.061 & 0.861 & 0.059 & 0.880 & 0.056 & 0.866 & 0.063 \\
DistilBERT ML& 0.867 & 0.058 & 0.864 & 0.057 & 0.868 & 0.059 & 0.870 & 0.056 \\
LLama2       & 0.702 & 0.189 & 0.612 & 0.200 & 0.533 & 0.216 & 0.589 & 0.205 \\
\hline \hline
\multicolumn{9}{|c|}{P$_{Comp}$} \\ \hline
Word2Vec     & 0.840 & 0.039 & 0.838 & 0.039 & 0.714 & 0.269 & 0.703 & 0.280 \\
GloVe        & 0.835 & 0.041 & 0.837 & 0.040 & 0.715 & 0.276 & 0.710 & 0.282 \\
ELMo         & 0.859 & 0.042 & 0.823 & 0.040 & 0.781 & 0.080 & 0.733 & 0.093 \\
SBERT ML     & 0.815 & 0.042 & 0.805 & 0.038 & 0.823 & 0.050 & 0.808 & 0.052 \\
BERT         & 0.849 & 0.060 & 0.886 & 0.037 & 0.855 & 0.066 & 0.864 & 0.041 \\
BERT ML      & 0.923 & 0.022 & 0.913 & 0.020 & 0.930 & 0.021 & 0.921 & 0.023 \\
DistilBERT ML& 0.922 & 0.015 & 0.922 & 0.013 & 0.929 & 0.018 & 0.931 & 0.014 \\
LLama2       & 0.828 & 0.102 & 0.844 & 0.086 & 0.741 & 0.174 & 0.749 & 0.174 \\
\hline \hline
\multicolumn{9}{|c|}{P$_{WordsSyn}$} \\ \hline
Word2Vec     & 0.524 & 0.098 & 0.524 & 0.097 & 0.459 & 0.185 & 0.450 & 0.189 \\
GloVe        & 0.569 & 0.119 & 0.572 & 0.116 & 0.356 & 0.196 & 0.357 & 0.198 \\
ELMo         & 0.759 & 0.083 & 0.707 & 0.091 & 0.644 & 0.100 & 0.557 & 0.110 \\
SBERT ML     & 0.659 & 0.112 & 0.645 & 0.112 & 0.670 & 0.119 & 0.662 & 0.122 \\
BERT         & 0.780 & 0.105 & 0.850 & 0.064 & 0.783 & 0.077 & 0.820 & 0.054 \\
BERT ML      & 0.881 & 0.035 & 0.867 & 0.040 & 0.887 & 0.035 & 0.877 & 0.039 \\
DistilBERT ML& 0.870 & 0.029 & 0.868 & 0.027 & 0.875 & 0.027 & 0.877 & 0.026 \\
LLama2       & 0.668 & 0.148 & 0.601 & 0.151 & 0.490 & 0.137 & 0.560 & 0.118 \\
\hline \hline
\multicolumn{9}{|c|}{P$_{Rand}$} \\ \hline
Word2Vec     & 0.419 & 0.064 & 0.423 & 0.065 & 0.460 & 0.185 & 0.371 & 0.151 \\
GloVe        & 0.413 & 0.108 & 0.419 & 0.108 & 0.356 & 0.196 & 0.293 & 0.219 \\
ELMo         & 0.674 & 0.082 & 0.628 & 0.069 & 0.644 & 0.100 & 0.482 & 0.097 \\
SBERT ML     & 0.479 & 0.067 & 0.473 & 0.067 & 0.670 & 0.119 & 0.479 & 0.072 \\
BERT         & 0.746 & 0.117 & 0.855 & 0.061 & 0.783 & 0.077 & 0.808 & 0.037 \\
BERT ML      & 0.872 & 0.031 & 0.872 & 0.028 & 0.887 & 0.035 & 0.883 & 0.032 \\
DistilBERT ML& 0.879 & 0.024 & 0.879 & 0.021 & 0.875 & 0.027 & 0.898 & 0.021 \\
LLama2       & 0.631 & 0.100 & 0.568 & 0.105 & 0.490 & 0.137 & 0.544 & 0.102 \\
\hline
\end{tabular}
\caption{Mean and standard deviation at NC level for {P$_{Syn}$}, {P$_{Comp}$}, P$_{WordsSyn}$ and {P$_{Rand}$}, for English (EN) and Portuguese (PT) for naturalistic (Nat) and neutral (Neut) sentences. } \label{tab:num_NC}
\end{center}
\end{table}

\begin{table}[!h]
\small
\begin{center}
\begin{tabular}{|l|cc|cc|cc|cc|}
\hline
\multicolumn{9}{|c|}{A$_{Syn|WordsSyn}$} \\ \hline
\multirow{2}{*}{Model Name} & \multicolumn{2}{c|}{EN-Nat} & \multicolumn{2}{c|}{EN-Neut} & \multicolumn{2}{c|}{PT-Nat} & \multicolumn{2}{c|}{PT-Neut} \\
 & mean & std & mean & std & mean & std & mean & std \\
\hline
Word2Vec     & -0.002 & 0.149 & 0.004 & 0.156 & 0.025 & 0.152 & 0.038 & 0.160 \\
GloVe        & -0.009 & 0.166 & -0.006 & 0.170 & 0.058 & 0.193 & 0.072 & 0.193 \\
ELMo         & -0.023 & 0.108 & -0.040 & 0.134 & -0.003 & 0.124 & 0.008 & 0.182 \\
SBERT ML     & -0.036 & 0.160 & -0.051 & 0.178 & -0.019 & 0.154 & -0.036 & 0.176 \\
BERT         & 0.021 & 0.090 & 0.006 & 0.067 & 0.027 & 0.077 & 0.017 & 0.073 \\
BERT ML      & -0.002 & 0.044 & -0.003 & 0.056 & -0.003 & 0.044 & -0.008 & 0.059 \\
DistilBERT ML& -0.001 & 0.041 & -0.003 & 0.045 & -0.003 & 0.047 & -0.002 & 0.049 \\
LLama2       & 0.020 & 0.137 & 0.011 & 0.154 & 0.024 & 0.169 & 0.021 & 0.166 \\
\hline \hline
\multicolumn{9}{|c|}{A$_{Syn|Rand}$} \\ \hline
Word2Vec     & 0.049 & 0.156 & 0.054 & 0.162 & 0.076 & 0.165 & 0.074 & 0.182 \\
GloVe        & 0.070 & 0.177 & 0.077 & 0.179 & 0.110 & 0.213 & 0.100 & 0.219 \\
ELMo         & 0.024 & 0.116 & 0.015 & 0.135 & 0.051 & 0.127 & 0.062 & 0.198 \\
SBERT ML     & 0.059 & 0.173 & 0.072 & 0.186 & 0.081 & 0.165 & 0.099 & 0.171 \\
BERT         & 0.040 & 0.103 & 0.001 & 0.065 & 0.057 & 0.085 & 0.026 & 0.070 \\
BERT ML      & 0.003 & 0.048 & -0.006 & 0.052 & 0.000 & 0.042 & -0.011 & 0.054 \\
DistilBERT ML& -0.005 & 0.047 & -0.007 & 0.049 & -0.012 & 0.044 & -0.015 & 0.046 \\
LLama2       & 0.042 & 0.124 & 0.031 & 0.133 & 0.064 & 0.165 & 0.034 & 0.162 \\
\hline 
\end{tabular}
\caption{Mean and standard deviation at NC level for {A$_{Syn|WordsSyn}$} and {A$_{Syn|Rand}$}, for English (EN) and Portuguese (PT) for naturalistic (Nat) and neutral (Neut) sentences.} \label{tab:num_Affinity}
\end{center}
\end{table}

\begin{table}[!h]
\small
\begin{center}
\begin{tabular}{|l|cc|cc|cc|cc|}
\hline
\multicolumn{9}{|c|}{Sim$_{R|Syn}$} \\ \hline
\multirow{2}{*}{Model Name} & \multicolumn{2}{c|}{EN-Nat} & \multicolumn{2}{c|}{EN-Neut} & \multicolumn{2}{c|}{PT-Nat} & \multicolumn{2}{c|}{PT-Neut} \\
 & mean & std & mean & std & mean & std & mean & std \\
\hline
Word2Vec     & 0.164 & 0.365 & 0.159 & 0.362 & 0.221 & 0.356 & 0.183 & 0.373 \\
GloVe        & 0.221 & 0.407 & 0.220 & 0.406 & 0.264 & 0.391 & 0.225 & 0.424 \\
ELMo         & 0.076 & 0.512 & 0.012 & 0.470 & 0.154 & 0.384 & 0.104 & 0.412 \\
SBERT ML     & 0.190 & 0.441 & 0.172 & 0.429 & 0.259 & 0.419 & 0.244 & 0.395 \\
BERT         & 0.075 & 0.735 & -0.166 & 0.659 & 0.289 & 0.486 & 0.098 & 0.437 \\
BERT ML      & -0.024 & 0.533 & -0.128 & 0.525 & -0.057 & 0.510 & -0.194 & 0.566 \\
DistilBERT ML& -0.147 & 0.566 & -0.166 & 0.544 & -0.257 & 0.589 & -0.320 & 0.618 \\
LLama2       & 0.194 & 0.506 & 0.095 & 0.466 & 0.129 & 0.389 & 0.056 & 0.448 \\
\hline \hline
\multicolumn{9}{|c|}{Sim$_{R|WordsSyn}$} \\ \hline
Word2Vec     & 0.173 & 0.182 & 0.167 & 0.181 & 0.165 & 0.187 & 0.124 & 0.204 \\
GloVe        & 0.245 & 0.236 & 0.243 & 0.233 & 0.113 & 0.237 & 0.061 & 0.246 \\
ELMo         & 0.231 & 0.276 & 0.193 & 0.272 & 0.185 & 0.234 & 0.118 & 0.262 \\
SBERT ML     & 0.336 & 0.231 & 0.315 & 0.230 & 0.339 & 0.258 & 0.340 & 0.256 \\
BERT         & 0.092 & 0.307 & -0.058 & 0.272 & 0.169 & 0.275 & 0.057 & 0.238 \\
BERT ML      & 0.034 & 0.294 & -0.068 & 0.357 & 0.007 & 0.296 & -0.098 & 0.366 \\
DistilBERT ML& -0.105 & 0.284 & -0.104 & 0.246 & -0.196 & 0.300 & -0.244 & 0.304 \\
LLama2       & 0.094 & 0.380 & 0.058 & 0.368 & 0.099 & 0.240 & 0.039 & 0.255 \\
\hline 
\end{tabular}
\caption{
Mean and standard deviation at NC level for {Sim$_{R|Syn}$} and {Sim$_{R|WordsSyn}$}, for English (EN) and Portuguese (PT) for naturalistic (Nat) and neutral (Neut) sentences.} \label{tab:num_S14_S34}
\end{center}
\end{table}

\section{Appendix B: Results After Removing Examples with Synonym Lexical Overlaps}

\begin{table}[!ht]
\centering
\begin{footnotesize}
\begin{tabular}{|l|cccccccc|}
\hline
	&	Word2Vec	&	GloVe	&	ELMo	&	SBERT  	&	BERT	&	BERT 	&	DistilBERT 	&	LLaMA2	 \\ 
    	&		&		&		&	  ML	&	 	&	 ML	&	 ML	&		 \\  \hline
A$_{Syn|WordsSyn}$	&		&		&		&		&		&		&		&		 \\  
EN-Nat	&	0.44	&	0.35	&	0.44	&	0.48	&	0.42	&	0.44	&	0.32	&	-	\\ 
EN-Neut	&	0.44	&	0.35	&	0.42	&	0.48	&	0.34	&	0.36	&	0.29	&	-	\\ 
PT-Nat	&	0.26	&	0.19	&	0.23	&	0.16	&	0.32	&	0.12	&	-	&	-	\\ 
PT-Neut	&	0.28	&	-	&	0.26	&	-	&	0.21	&	-	&	-	&	-	\\ \hline 

A$_{Syn|Rand}$	&		&		&		&		&		&		&		&		 \\ 
EN-Nat	&	0.49	&	0.37	&	0.57	&	0.53	&	0.57	&	0.56	&	0.46	&	0.16	\\ 
EN-Neut	&	0.47	&	0.36	&	0.54	&	0.51	&	0.41	&	0.39	&	0.37	&	-	\\ 
PT-Nat	&	0.25	&	0.16	&	0.25	&	0.25	&	0.40	&	0.20	&	-	&	-	\\ 
PT-Neut	&	-	&	-	&	0.28	&	0.22	&	0.31	&	-	&	-	&	-	\\ \hline 
\end{tabular}
\end{footnotesize}
\caption{Spearman $\rho$ correlation between the \emph{Affinity} and human judgments for English and Portuguese for naturalistic (Nat) and neutral (Neut) sentences after removing NCs with lexical overlap between NC and NC$_{Syn}$. Non-significant (p $>$ 0.05) results omitted from the table. }
\label{tab:results_Affinity_cor_remove}
\end{table}

\begin{table*}[!ht]
\centering
\begin{footnotesize}
\begin{tabular}{|l|cccccccc|}
\hline
	&	Word2Vec	&	GloVe	&	ELMo	&	SBERT  	&	BERT	&	BERT 	&	DistilBERT 	&	LLaMA2	 \\ 
    	&		&		&		&	  ML	&	 	&	 ML	&	 ML	&		 \\ \hline
$\mbox{Sim}_{R|Syn}$	&		&		&		&		&		&		&		&		 \\       
EN-Nat	&	0.49	&	0.37	&	0.57	&	0.53	&	0.57	&	0.56	&	0.46	&	0.15	\\ 
EN-Neut 	&	0.47	&	0.36	&	0.54	&	0.51	&	0.41	&	0.39	&	0.37	&	-	\\ 
PT-Nat	&	0.29	&	0.21	&	0.37	&	0.24	&	0.32	&	0.24	&	-	&	-	\\ 
PT-Neut	&	-	&	-	&	0.40	&	-	&	0.29	&	-	&	-	&	-	\\ \hline 
												
$\mbox{Sim}_{R|WordsSyn}$	&		&		&		&		&		&		&		&		 \\       
EN-Nat 	&	-	&	-	&	0.14	&	0.11	&	0.38	&	0.26	&	0.28	&	0.20	\\ 
EN-Neut	&	-	&	-	&	-	&	-	&	-	&	-	&	0.22	&	-	\\ 
PT-Nat &	-	&	-	&	-	&	-	&	-	&	-	&	-	&	0.17	\\ 
PT-Neut	&	-	&	-	&	-	&	-	&	-	&	-	&	-	&	-	\\ \hline
\end{tabular}
\end{footnotesize}
\caption{Spearman $\rho$ correlation between the Scaled Similarities and human judgments, for $\mbox{Sim}_{R|Syn}$ and $\mbox{Sim}_{R|WordsSyn}$ in both English and Portuguese after removing NCs with lexical overlap between NC and NC$_{Syn}$. Non-significant (p $>$ 0.05) results were omitted from the table. }
\label{tab:results_scale_remove}
\end{table*}

As the NC$_{Syn}$ were selected from the synonyms proposed by the human annotators, and chosen according to frequency, this led to cases of lexical overlap. Removing the NCs with lexical overlap with their NC$_{Syn}$ and analysing the correlations for Affinities and Scaled Similarities, the results are as shown in Tables \ref{tab:results_Affinity_cor_remove} and  \ref{tab:results_scale_remove}. The results are compatible with those of Tables \ref{tab:results_Affinity_cor} and \ref{tab:results_probes1} for the complete set of NCs. As expected the correlations are smaller and less significant than those obtained for the full set, as with the removal of the NCs with lexical overlap a smaller set was used to calculate correlations. The ultimate test will be to redo the analysis with the full list of NCs but only using NC$_{Syn}$ without lexical overlap, but this requires additional human annotation and is left for future work.

\section{Appendix C: The impact of Rogue Dimensions}
\subsection{Standardisation Process}

To mitigate the impact of rogue dimensions, a standardisation process using z-scores\footnote{$z=(x-\mu)/\sigma$.} was applied as proposed by \citet{timkey2021all}.
The mean vector $\mu$  was calculated across the NC sentences and subtracted from each embedding vector to center the data. Each dimension of the embedding was divided by its standard deviation $\sigma$.

\subsection{Spearman Correlation Analysis}
To assess the impact of standardisation, Spearman correlation was calculated between the P$_{Syn}$ cosine similarities before and after standardisation:

\begin{itemize}
    \item \textbf{Pre-standardisation:} Cosine similarities calculated using the original representations.
    \item \textbf{Post-standardisation:} Cosine similarities recalculated after standardisation.
\end{itemize}
The results are reported on Table \ref{tab:results_P1_rogue_original_cor}.

\begin{table*}[!ht]
\centering
\begin{footnotesize}
\begin{tabular}{|l|cccccccc|}
\hline
 Sent	&	Word2Vec	&	GloVe	&	ELMo	&	SBERT  	&	BERT	&	BERT 	&	DistilBERT 	&	LLaMA2	 \\ 
    	&		&		&		&	  ML	&	 	&	 ML	&	 ML	&		 \\ \hline
EN-Nat &      &       &      & 0.974    & 0.964 & 0.964   & 0.954          & 0.960  \\ 
EN-Neut &     &       &      & 0.965    & 0.888 & 0.876   & 0.908          & 0.650  \\ 
PT-Nat &      &       &      & 0.976    & 0.860 & 0.955   & 0.955          & 0.960  \\ 
PT-Neut &     &       &      & 0.952    & 0.874 & 0.874   & 0.911          & 0.927  \\ \hline \hline
NC	&	Word2Vec	&	GloVe	&	ELMo	&	SBERT  	&	BERT	&	BERT 	&	DistilBERT 	&	LLaMA2	 \\ 
    	&		&		&		&	  ML	&	 	&	 ML	&	 ML	&		 \\ \hline
EN-Nat &      &       &      & 0.991    & 0.967 & 0.951   & 0.953          & 0.937  \\ 
EN-Neut &     &       &      & 0.984    & 0.940 & 0.916   & 0.947          & 0.875  \\ 
PT-Nat &      &       &      & 0.987    & 0.852 & 0.939   & 0.939          & 0.939  \\ 
PT-Neut &     &       &      & 0.975    & 0.795 & 0.903   & 0.910          & 0.925  \\ \hline 
\end{tabular}
\end{footnotesize}
\caption{Spearman $\rho$ correlation for P$_{Syn}$ cosine similarities before and after standardisation (results significant for $p < 0.05$.) }
\label{tab:results_P1_rogue_original_cor}
\end{table*}

\end{document}